\RequirePackage{etex}
\documentclass[10pt, letter, onecolumn]{arxiv}
\usepackage{helvet}
\usepackage{cite}
\usepackage{kantlipsum, lipsum}
\usepackage[pagebackref=false,breaklinks=false,%
            colorlinks=true,bookmarks=true,citecolor=ourdarkblue,%
            urlcolor=ourdarkblue,linkcolor=ourdarkblue]{hyperref}
\usepackage{multibib}

\usepackage{dm-colors}
\usepackage{amsmath}
\usepackage{pstricks, pst-node}
\usepackage{verbatim}
\usepackage{multirow}
\usepackage{array}
\usepackage{longtable}
\usepackage{pdflscape}
\usepackage{scalerel}
\usepackage{booktabs}
\usepackage{enumitem}
\usepackage{xspace}
\usepackage{bm}
\usepackage{bbm}
\usepackage{mathtools}
\usepackage{soul}
\usepackage{epsfig}
\usepackage{graphicx}
\usepackage{amssymb}
\usepackage{colortbl}
\usepackage{csquotes}
\usepackage{setspace}
\usepackage{bbding}
\usepackage{siunitx} 
\usepackage{threeparttable}
\usepackage{tabularx,ragged2e}
\usepackage{placeins}
\usepackage{bbding}
\definecolor{darkpastelgreen}{rgb}{0.13, 0.55, 0.13}
\definecolor{darkpastelred}{rgb}{0.55, 0.13, 0.13}
\definecolor{mygray}{rgb}{1, 1, 1}
\usepackage{lmodern}
\usepackage[hang,flushmargin]{footmisc}
\usepackage{nameref}
\usepackage{varioref}
\usepackage{amssymb}
\usepackage{pifont}
\usepackage{rotating}
\usepackage{graphicx}

\usepackage[noabbrev,capitalize]{cleveref}
\usepackage{etoc}
\usepackage{tikz}

\usepackage{wasysym}
\usepackage{algorithm}
\usepackage{makecell} 
\usepackage{algpseudocode}
\usepackage{xcolor}
\usepackage{wasysym}  
\usepackage[most]{tcolorbox}

\newtcolorbox{casestudy}[2][]{%
  colback=gray!5,colframe=gray!50,
  fonttitle=\bfseries,
  title={Case Study: #2},
  breakable,enhanced,
  left=1em,right=1em,top=0.8em,bottom=0.8em
}

\newcommand{\comments}[1]{%
  \par\medskip
  {\color{black!15}\rule{\linewidth}{0.4pt}}\par
  \noindent\textbf{Comments}\quad #1\par
}

\usepackage{amsmath}
\usepackage{thmtools}
\usepackage{wasysym}
\declaretheoremstyle[
    spaceabove=6pt, spacebelow=6pt,
    headfont=\bfseries, headpunct={.}, headformat={\NAME\ \NUMBER},
    bodyfont=\normalfont,
    postheadspace=0.5em
]{promptstyle}

\declaretheorem[name=Prompt, style=promptstyle]{prompt}

\tcolorboxenvironment{prompt}{
    colback=gray!10!,
    colframe=gray!75!,
    fonttitle=\bfseries,
    title=Prompt,
    boxrule=0.5pt,
    sharp corners
}

\usepackage{lineno}

\graphicspath{{figures/}}
\definecolor{mygray}{rgb}{0.85, 0.85, 0.85}

\newcommand{\change}[1]{{\textcolor{black}{#1}}}
\newcommand{\changee}[1]{{\textcolor{black}{#1}}}

\definecolor{lightpink}{RGB}{255,235,238}
\definecolor{lightgreen}{RGB}{232,245,233}

\newcommand{\blockthree}[2]{\multicolumn{3}{>{\columncolor{#1}}c}{#2}}
\newcommand{\colhead}[2]{\multicolumn{1}{>{\columncolor{#1}}c}{#2}}

\usepackage{listings}

\definecolor{codegreen}{rgb}{0,0.6,0}
\definecolor{codegray}{rgb}{0.5,0.5,0.5}
\definecolor{codepurple}{rgb}{0.58,0,0.82}
\definecolor{backcolour}{rgb}{0.95,0.95,0.92}
\definecolor{framecolor}{rgb}{0.8,0.8,0.8}

\lstdefinestyle{prettyjson}{
    backgroundcolor=\color{backcolour},   
    commentstyle=\color{codegreen},
    keywordstyle=\color{blue}\bfseries,
    numberstyle=\tiny\color{codegray},
    stringstyle=\color{codepurple},
    basicstyle=\ttfamily\small,
    breakatwhitespace=false,         
    breaklines=true,                 
    captionpos=b,                    
    keepspaces=true,                 
    numbers=left,                    
    numbersep=8pt,                  
    showspaces=false,                
    showstringspaces=false,
    showtabs=false,                  
    tabsize=2,
    frame=single,
    frameround=tttt,
    framerule=0.5pt,
    rulecolor=\color{framecolor},
    xleftmargin=15pt,
    xrightmargin=15pt,
    aboveskip=15pt,
    belowskip=15pt,
    columns=flexible,
    escapeinside={(*@}{@*)}
}

\title{\Large{An Agentic System for Rare Disease Diagnosis with \\ Traceable Reasoning}}

\author[1,2,3,$\ast$]{Weike Zhao} 
\author[1,$\ast$]{Chaoyi Wu} 
\author[4,5,$\ast$]{Yanjie Fan}
\author[1,2,3]{Pengcheng Qiu}
\author[6]{Xiaoman Zhang}
\author[1]{\\ \vspace{0.1cm} Yuze Sun}
\author[3]{Xiao Zhou}
\author[7]{Shuju Zhang}
\author[7]{Yu Peng}
\author[1]{Yanfeng Wang} 
\author[4]{\\ \vspace{0.1cm} Xin Sun} 
\author[1,3,8,$\dagger$]{Ya Zhang}
\author[4,$\dagger$]{Yongguo Yu}
\author[4,9,$\dagger$]{Kun Sun}
\author[1,3,$\dagger$]{Weidi Xie}

\affil[1]{\normalsize School of Artificial Intelligence, Shanghai Jiao Tong University, Shanghai, China  \authorcr \vspace{0.1cm}
}

\affil[2]{\normalsize School of Integrated Circuits, School of Information Science and Electronic Engineering, Shanghai Jiao Tong University, Shanghai, China \authorcr \vspace{0.1cm}}

\affil[3]{\normalsize Shanghai Artificial Intelligence Laboratory, Shanghai, China  \authorcr \vspace{0.1cm}
}

\affil[4]{\normalsize Xinhua Hospital affiliated to Shanghai Jiao Tong University School of Medicine, Shanghai, China \authorcr \vspace{0.1cm}
}

\affil[5]{\normalsize Shanghai Institute for Pediatric Research, Shanghai, China \authorcr \vspace{0.1cm}
}

\affil[6]{\normalsize Department of Biomedical Informatics, Harvard Medical School, Boston, MA, USA \authorcr \vspace{0.1cm}
}

\affil[7]{\normalsize The Affiliated Children's Hospital of Xiangya School of Medicine, Hunan, China \authorcr \vspace{0.1cm}
}
\affil[8]{\normalsize Institute of Artificial Intelligence for Medicine, School of Medicine, Shanghai Jiao Tong University, Shanghai, China  \authorcr \vspace{0.1cm}
}
\affil[9]{\normalsize Engineering Research Center of Techniques and Instruments for Diagnosis and Treatment of Congenital Heart Disease, Ministry of Education, China  \authorcr \vspace{0.1cm}
}

\affil[$\ast$]{\normalsize Equal contributions\hspace{1cm}}
\affil[$\dag$]{\normalsize Corresponding author\authorcr Ya Zhang: ya\_zhang@sjtu.edu.cn; 
Yongguo Yu: yuyongguo@shsmu.edu.cn; \authorcr Kun Sun: sunkun@xinhuamed.com.cn; Weidi Xie: weidi@sjtu.edu.cn}

\begin{document}

\begin{abstract}
Rare diseases affect over 300 million individuals worldwide~\cite{valdez2016public, nguengang2020estimating, lancet2024landscape}, yet timely and accurate diagnosis remains an urgent challenge~\cite{valdez2016public, lancet2024landscape, schieppati2008rare, genomics_diagnostic_odyssey}. Patients often endure a prolonged ``diagnostic odyssey'' exceeding five years, marked by repeated referrals, misdiagnoses, and unnecessary interventions, leading to delayed treatment and substantial emotional and economic burdens~\cite{schieppati2008rare, genomics_diagnostic_odyssey}. 
Here we present \textbf{DeepRare}, a multi-agent system for rare‑disease differential diagnosis decision support~\cite{chen2024rareagents, zou2025rise, anthropic_context_protocol} powered by large language models, integrating over 40 specialized tools and up-to-date knowledge sources. DeepRare processes heterogeneous clinical inputs, including free-text descriptions, structured Human Phenotype Ontology terms, and genetic testing results, to generate ranked diagnostic hypotheses with transparent reasoning linked to verifiable medical evidence.  
Evaluated across nine datasets from literature, case reports and clinical centres across Asia, North America and Europe spanning 14 medical specialties, DeepRare demonstrates exceptional performance on 3,134 diseases. In human-phenotype-ontology-based tasks, it achieves an average Recall@1
of 57.18\%, outperforming the next best method by 23.79\%; in multi-modal tests, it
reaches 69.1\% compared with Exomiser’s 55.9\% on 168 cases. Expert review achieved 95.4\% agreement on its reasoning chains, confirming their validity and traceability. Our work not only advances rare disease diagnosis but also demonstrates how the
latest powerful large-language-model-driven agentic systems can reshape current
clinical workflows.

\end{abstract}

\maketitle

\section{Main}

Rare diseases—defined as conditions affecting fewer than 1 in 2,000 individuals—collectively impact over 300 million people worldwide, with more than 7,000 distinct disorders identified to date, approximately 80\% of which are genetic in origin~\cite{valdez2016public, nguengang2020estimating, lancet2024hope, lancet2024landscape}. Despite their cumulative burden, rare diseases remain notoriously difficult to diagnose due to their clinical heterogeneity, low individual prevalence, and limited clinician familiarity~\cite{valdez2016public, nguengang2020estimating, schieppati2008rare, chen2024rareagents, liu2025generalist, chen2024rarebench, mcduff2025towards, lunke2023integrated, kernohan2024expanding, health2025exploring}. Patients often experience a prolonged “diagnostic odyssey” averaging over five years, marked by repeated referrals, misdiagnoses, and unnecessary interventions, all of which contribute to delayed treatment and adverse outcomes~\cite{schieppati2008rare, genomics_diagnostic_odyssey}. These challenges highlight the urgent need for scalable, accurate, and interpretable diagnostic tools, an area where recent advances in multi-agent systems offer transformative potential.

Developing artificial intelligence (AI) systems for rare disease diagnosis presents several inherent challenges, (i) \textbf{multidisciplinary}: rare diseases often manifest with complex, heterogeneous, and multisystem symptoms, requiring diagnostic models to possess multidisciplinary medical knowledge and the ability to interpret diverse patient phenotypes~\cite{macken2022specialist, khalife2023multidisciplinary}; 
(ii) \textbf{limited cases}: the scarcity of cases for individual rare diseases limits the availability of training data, making it difficult to develop robust models and increasing the risk of overfitting and catastrophic forgetting; 
(iii) \textbf{dynamic knowledge updates}: the rare disease knowledge landscape is rapidly evolving, with approximately 260 to 280 rare genetic diseases discovered per year, 
according to the International Rare Diseases Research Consortium (IRDiRC)~\cite{dawkins2017progress}. This dynamic nature demands AI systems that are not only updatable but also capable of integrating new knowledge efficiently; (iv) \textbf{transparency and traceability}: clinical deployment demands interpretability: diagnostic suggestions must be accompanied by transparent, traceable reasoning to support clinician trust and accountability.

\change{Recent advances in agentic large language model (LLM) systems have opened new avenues for rare disease diagnosis~\cite{chen2024rareagents, qiu2024llm, zou2025rise, lee2024ai, talebirad2023multi, zheng2024can, langchain, anthropic_agents, zheng2025end, qiu2025evolving}. 
These systems orchestrate multiple specialized tools and sub-agents~\cite{langchain, anthropic_agents}}, enabling seamless integration of external knowledge bases, case repositories, and multimodal analytical components~\cite{anthropic_context_protocol, talebirad2023multi}. Unlike conventional supervised learning approaches, these systems are typically training-free and excel in few-shot and zero-shot scenarios—an essential capability for rare disease applications where annotated data are scarce. Their modular and interpretable architectures further facilitate transparent, auditable, and clinically actionable diagnostic workflows.

\changee{Here, we present \textbf{DeepRare}, an agentic LLM-based system designed specifically for rare disease differential diagnosis decision support.} DeepRare is capable of processing heterogeneous patient inputs, including free-text clinical descriptions, structured Human Phenotype Ontology (HPO) terms, and genomic testing results. Based on the input, the system generates a ranked list of candidate diagnoses, each supported by a transparent chain of reasoning that directly references verifiable medical evidence, enhancing interpretability and supporting clinician trust in AI-assisted decisions. \change{Inspired by the Model Context Protocol (MCP)~\cite{anthropic_context_protocol}, DeepRare employs a three-tier architecture: a central LLM-powered host with memory coordinates the process, specialized agent servers handle phenotype and genotype analysis, normalization, and knowledge retrieval, and the outer tier integrates curated and web-scale medical resources. To improve robustness, DeepRare further employs a self-reflective loop that iteratively reassesses hypotheses, reducing over-diagnosis and mitigating LLM hallucinations.}

We evaluate \textbf{DeepRare} 
on 6,563 clinical cases collected from seven public datasets and two in-house dataset, sourced from diverse populations across Asia, North America, and Europe. Significantly, the two in-house datasets, from Xinhua Hospital~(Shanghai) and The Affiliated Children’s Hospital~(Hunan), contain 330 cases with not only phenotypes but also whole exome sequencing data. To the best of our knowledge, this is the only rare disease diagnosis benchmark featuring original gene testing data. All diagnoses in this cohort are rigorously validated by genetic testing, providing a high-quality standard for assessing diagnostic performance. 
\textbf{DeepRare} consistently achieves superior diagnostic accuracy across all 8 datasets of 3,134 rare diseases spanning 14 medical specialties. 

In HPO-based evaluations, compared with other 15 methods like traditional bioinformatics tools, large language models, and agentic systems, \textbf{DeepRare} achieves an average score of 57.18\%, 65.25\% at Recall@1, and Recall@3, respectively, 
surpassing the second-best method (Reasoning LLM) by substantial margins of 23.79\%, 18.65\%.
In multi-modal input scenarios, \textbf{DeepRare} achieves a Recall@1 of 69.1\%, outperforming Exomiser’s 55.9\%  in the Xinhua whole-exome cases. Additionally, we engage 10 rare disease physicians to manually verify the traceable reasoning chains generated by the system across 180 cases. 
DeepRare demonstrates high reliability in evidence factuality, achieving 95.4\% agreement with clinical experts, thereby confirming that its intermediate reasoning steps are both medically valid and traceable to authoritative sources. 
To facilitate clinical adoption, we have deployed \textbf{DeepRare} as a user-friendly web application as a diagnostic copilot for rare disease physicians. 
Finally, we discuss the robustness of our agentic framework by evaluating different underlying LLMs and analyzing the contribution of each module, demonstrating the superiority of our system design.

\section{Results}

This section presents the results of our study, 
beginning with an overview of the proposed framework, \textbf{DeepRare}, and the evaluation settings, 
followed by a detailed analysis of the main findings.

\subsection{System Overview} 

\begin{figure}
    \centering
    \includegraphics[width=1\linewidth]{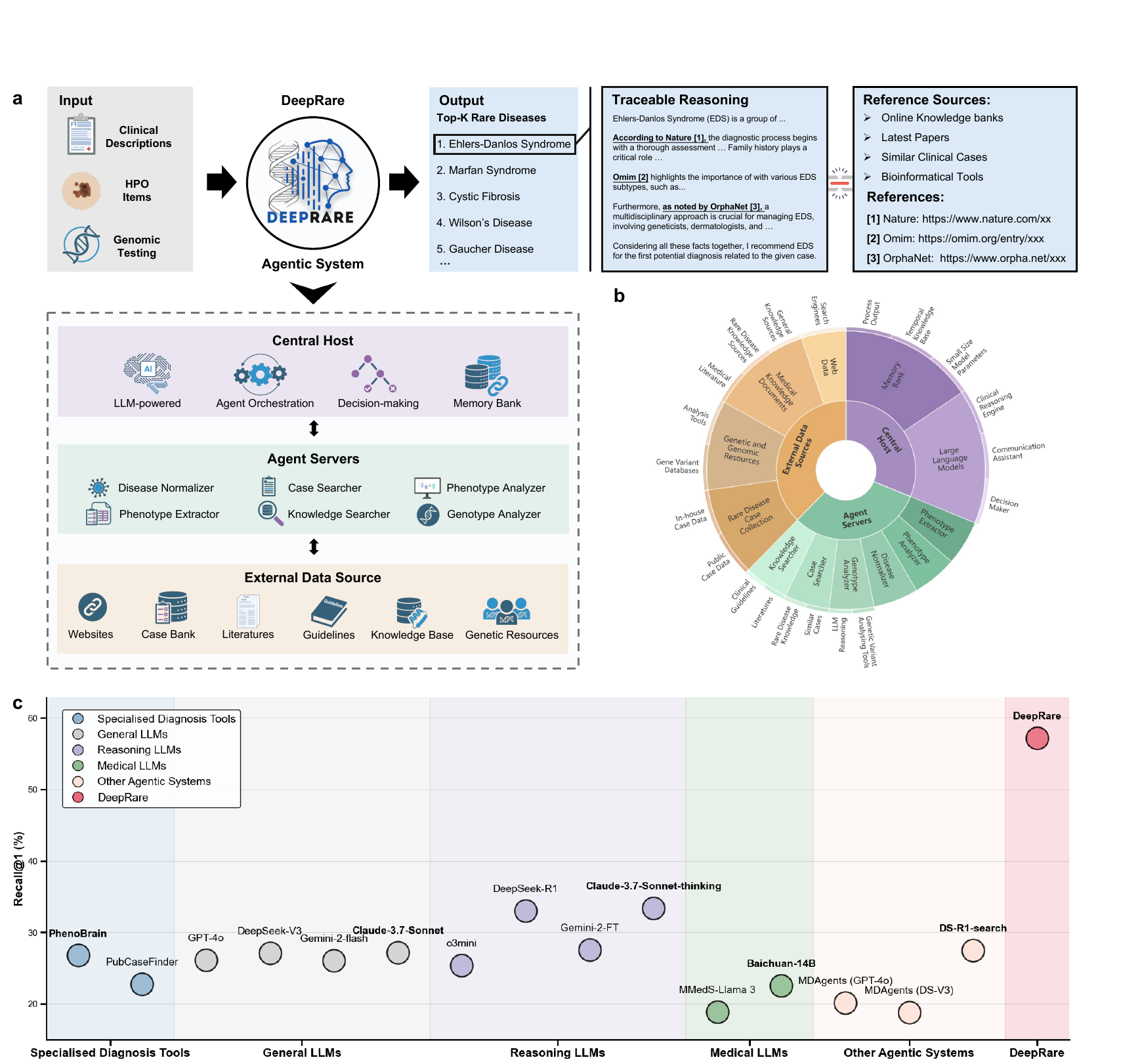}
    \vspace{0.01cm}
    \caption{\textbf{DeepRare: An agentic framework for rare disease prioritization.} 
    \textbf{a} System workflow: Multi-modal patient data (HPO terms, genomic variants) are processed through a tiered MCP-inspired architecture, generating a ranked Top-K diagnosis list with evidence-supported reasoning chains. 
    \textbf{b} Knowledge architecture: Sunburst visualization depicting hierarchical integration of diagnostic tools and biomedical knowledge sources within DeepRare. 
    \textbf{c} Performance benchmarking: Comparative evaluation across diagnostic APIs, general-purpose LLMs, reasoning-enhanced LLMs, medically-tuned LLMs, and agentic systems. Created in BioRender. \url{https://BioRender.com/ija3tl0}}
    \label{fig:teaser}
\end{figure}

\textbf{DeepRare} is an LLM-powered agentic system for rare disease diagnosis. 
It features a three-tier architecture inspired by Model Context Protocol~(MCP)~\cite{anthropic_context_protocol} that synergistically integrates reasoning-enhanced large language models with a broad range of clinical knowledge sources as shown in Figure~\ref{fig:teaser}a and~\ref{fig:teaser}b. 
The system comprises:
(i) a central host, powered by LLMs~(\change{locally implemented DeepSeek-V3 by default}) and equipped with a memory bank, 
which orchestrates the entire diagnostic workflow by synthesizing collected evidence; 
(ii) multiple specialized agent servers, each managing a local set of tools to perform various rare disease-related analytical tasks and interact with distinct resource environments; and (iii) heterogeneous web-scale medical sources, which provide essential and traceable diagnostic evidence—such as research articles, clinical guidelines, and existing patient cases, \emph{etc.}.

Upon receiving a clinical case—provided as free-text phenotypic descriptions, structured Human Phenotype Ontology (HPO) terms, raw VCF files, or any combination thereof—the central host systematically decomposes the diagnostic task. It first orchestrates the agent servers to retrieve relevant evidence and references from external data sources, tailored to the patient’s information. The host then synthesizes this evidence to generate preliminary diagnostic hypotheses, followed by a self-reflection phase in which it conducts additional searches to rigorously validate or refute these hypotheses. 
If no hypothesized diseases meet the self-reflection criteria, 
the system iteratively revisits earlier steps to acquire further patient-specific evidence, repeating this diagnostic loop until a satisfactory resolution is achieved. Ultimately, \textbf{DeepRare} outputs a ranked list of potential rare diseases, each accompanied by a transparent reasoning chain that directly links each inference step to trusted medical evidence. Further details on the system workflow are provided in the \textbf{Method} section.

\subsection{Evaluation Settings}

To evaluate the performance of \textbf{DeepRare}, 
we consider \textbf{three baseline approaches}: 

\vspace{-0.1cm}
\begin{itemize}
\setlength\itemsep{0.2cm}
    \item \textbf{Specialised rare disease diagnosis tools.} 
    We consider the bioinformatics tools that are directly designed for rare disease diagnosis. 
    Two HPO-wise analysis tools, PhenoBrain~\cite{mao2025phenotype} and PubCaseFinder~\cite{Pubcase2}, are considered here.

    \item \textbf{Latest LLMs.} 
    We compare different LLMs, including \textbf{general LLMs, reasoning-enhanced LLMs, and medical LLMs}. \textbf{General LLMs} denote the most commonly used LLMs without extra reasoning enhancement or domain alignment, including GPT-4o~\cite{openai2023gpt}, DeepSeek-V3~\cite{guo2025deepseek}, Gemini-2.0-flash~\cite{team2024gemini}, and Claude-3.7-Sonnet~\cite{claude}. \textbf{Reasoning LLMs} denote the latest generation LLMs enhanced with an explicit reasoning chain, including o3mini~\cite{openai_o3_mini}, DeepSeek-R1~\cite{guo2025deepseek}, Gemini-2.0-FT~\cite{team2024gemini}, Claude-3.7-Sonnet-thinking~\cite{claude}. \textbf{Medical LLMs} refers to LLMs specifically developed for the medical domain, with Baichuan-14B~\cite{wang2025baichuan} and MMedS-Llama 3~\cite{wu2025towards} serving as notable representatives. All these LLMs are adapted to rare disease diagnosis, leveraging well-designed Prompt~\ref{prompt1:LLM diagnosis}.

    \item \textbf{Other agentic systems.} 
    We compare to existing agentic systems that build upon different LLMs, including MDAgents~\cite{kim2024mdagents}, which employs multidisciplinary consultations via GPT-4o or DeepSeek-V3, and DS-R1-search, which enhances DeepSeek-V3 with real-time internet retrieval.
    
\end{itemize}
More detailed descriptions on these baselines can be found in the \textbf{Method} section.

To demonstrate the effectiveness of our method, we perform a thorough cross-center evaluation. As shown in Figure~\ref{fig:teaser}c, we consider \textbf{eight rare disease diagnostic evaluation datasets}, with 6,401 clinical cases collected from seven public datasets and one in-house dataset. 
These datasets can be categorized into three groups based on sources, denoting varying diagnostic difficulties:
\vspace{-0.1cm}
\begin{itemize}
\setlength\itemsep{0.15cm}
    \item \textbf{From research papers}: RareBench-MME~\cite{jiang2025mme}~(40 cases), RareBench-LIRICAL~\cite{robinson2020LIRICAL}~(370 cases), DDD~\cite{firth2009decipher}~(2283 cases). 
    These cases are extracted from papers and manually verified. 
    Considering that the cases present in the literature are often typical and well-documented, diagnosis on these datasets tends to be relatively easy.
    
    \item \textbf{From case reports}: RareBench-RAMEDIS~\cite{topel2010RAMEDIS} (624 cases), MyGene2~\cite{mygene2, alsentzer2022few} (146 cases). These rare disease cases are uploaded by scientists or patients through case reports. They offer authenticity but also have undergone extra manual filtering, thus presenting moderate diagnostic difficulty.
    
    \item \textbf{From real clinical centers}: RareBench-HMS~\cite{ronicke2019HMS}~(88 cases), MIMIC-IV-Rare~\cite{mimic-iv-note}~(1,875 cases), Xinhua Hosp.~(975 cases), Hunan Hosp.(162 cases). These datasets are directly collected from four independent clinical centers of real patients in daily diagnostic procedures. 
    Due to the complexity and diversity of real patients, 
    these benchmarks are more challenging and aligned with real clinical practice.
    Specifically, RareBench-HMS is collected from the outpatient clinic at Hannover Medical School in Germany. MIMIC-IV-Rare is collected from Beth Israel Deaconess Medical Center in Boston, MA, USA by filtering for the rare-disease related cases. 
    \change{Xinhua Hosp. and Hunan Hosp. are two newly collected in-house data from Xinhua Hospital Affiliated to Shanghai Jiao Tong University School of Medicine, China and the Hunan Children’s Hospital, respectively.
    Notably, the 168 cases within the Xinhua Hosp. dataset and the entire Hunan Hosp. dataset~(162 cases) include original gene VCF files generated from whole-exome sequencing (WES), constituting a test subset for evaluating genomic data analysis within our collection, spanning two distinct health systems that differ in patient demographics, disease profiles, and clinical practice patterns.}
    \textbf{Due to patient privacy considerations, the Xinhua hosp. dataset and Hunan hosp. dataset evaluate exclusively using local models without external API access.} These datasets cover three distinct regions, enabling cross-center evaluation of our methods on different population distributions in various countries.
\end{itemize}

To the best of our knowledge, this collection is the most comprehensive benchmark for rare disease diagnosis, covering \textbf{3,134 diseases} from different case sources, and multiple independent clinical centers.

For each diagnostic task, we generated the top five most probable diagnostic predictions. The position of the correct diagnosis within these predictions was determined using GPT-4o under Prompt~\ref{prompt2:eval}~(more detailed reliability validation can be found in Supplementary Materials), and we subsequently calculated Recall@1, Recall@3, and Recall@5 metrics across the entire dataset.

\subsection{HPO-wise Analysis across Dataset}

Figure~\ref{fig:teaser}d presents a comparison on HPO-wise diagnosis of average Recall@1 across all benchmarks~(except for Xinhua Hosp. due to privacy issues).
\textbf{Our proposed DeepRare clearly demonstrates superior performance across all method categories, achieving 57.18\% top-1 diagnosis recall, and significantly outperforming the second-best method, Claude-3.7-Sonnet-thinking (33.39\%).} 
Specifically, we can draw the following key observations: 
(i) LLM-supported approaches consistently outperform traditional rare disease diagnostic models (PhenoBrain, PubCaseFinder), demonstrating enhanced flexibility in handling diverse clinical presentations; (ii) Reasoning-enhanced LLMs systematically surpass their general-purpose counterparts without explicit reasoning, likely due to their transparent reasoning traces that improve diagnostic accuracy; 
(iii) General-purpose LLMs unexpectedly exceed medical domain-tuned LLMs in performance, potentially reflecting parameter scale advantages and broader training diversity; (iv) Our multi-agent framework significantly advances beyond existing single-model approaches, highlighting the value of orchestrated specialist agents in complex diagnostic reasoning.

\begin{figure}[!t]
    \centering
    \includegraphics[width=\linewidth]{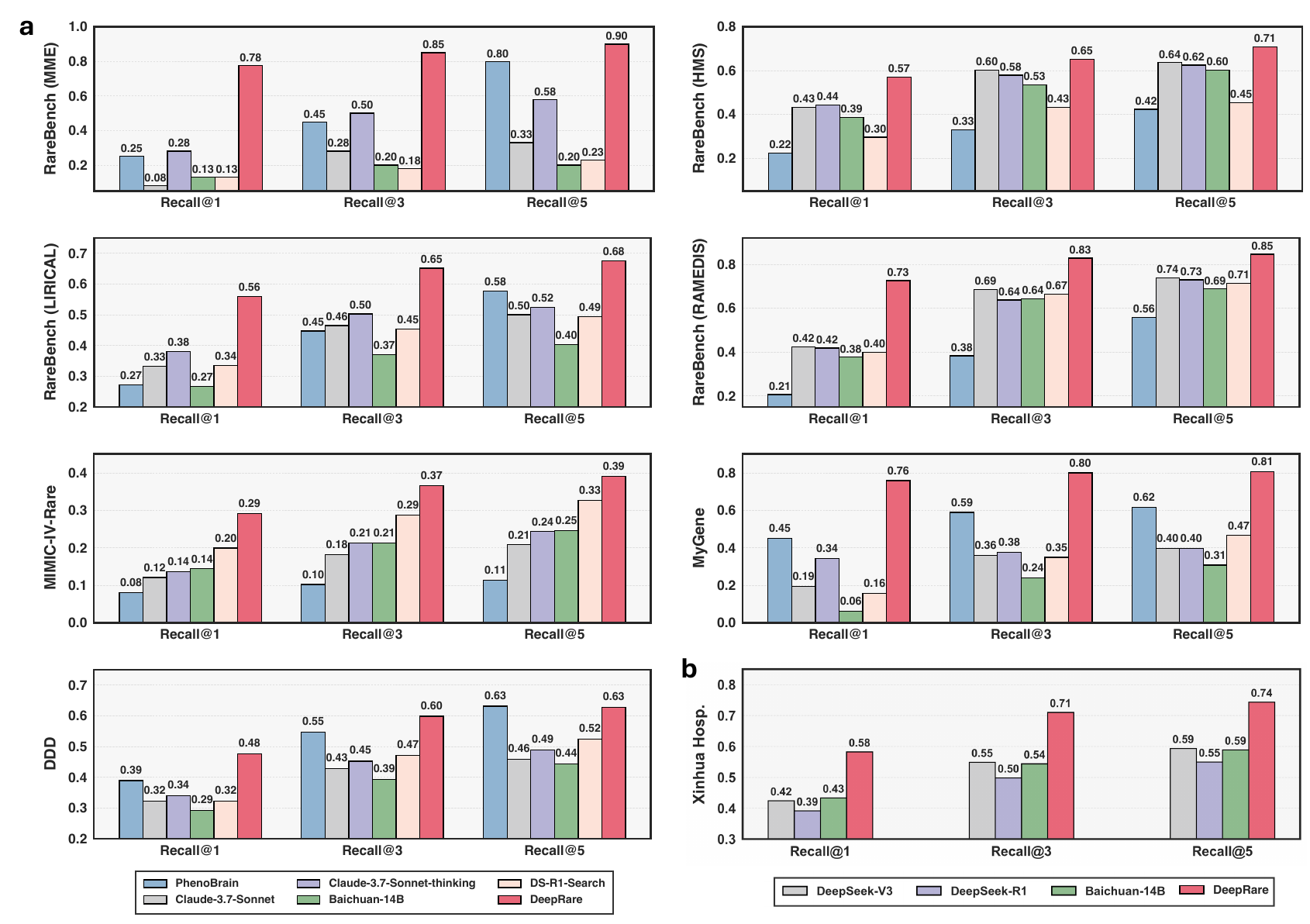}
    \vspace{0.06cm}
    \caption{\textbf{HPO-wise cross-dataset evaluation and comparative performance of DeepRare.}
\textbf{a} Diagnostic accuracy on seven public rare disease registries, demonstrating DeepRare's significant advantage over leading baselines – particularly in RareBench-MME (70.0\% top-1 accuracy) and RareBench-RAMEDIS (72.6\% top-1 accuracy). 
\textbf{b} Superior performance consistency on the Xinhua Hospital cohort (local model evaluation only due to privacy consideration). }
    \label{fig:result1}
\end{figure}

As shown in Figure~\ref{fig:result1}a and~\ref{fig:result1}b, 
we present detailed comparison on each dataset~(only the top-performing models in each baseline category are shown).
Complete results for all methods can be found in Supplementary Table~\ref{tab:s1}.
\textbf{Our proposed DeepRare system consistently outperforms all existing methods on all benchmarks.} 
Specifically, in the RareBench (MME) evaluation, \textbf{DeepRare} achieves exceptional scores of 78\%, and 85\% for Recall@1, Recall@3, respectively, surpassing the second-best baseline method~(PubCaseFinder) by margins of 30\%, 20\%. 
The system demonstrates particularly strong results on the MyGene2 evaluation with scores of 74\%, 81\% surpassing second methods by substantial margins of 35\%, 28\%.

In clinical datasets, \textbf{DeepRare} maintains its performance edge on the MIMIC-IV-Rare test (29\%, 37\%). 
In addition to the public benchmarks, we also report its performance on the in-house clinical testset, Xinhua Hosp (Figure~\ref{fig:result1}b). We mainly compare to the DeepSeek-V3, DeepSeek-R1, and MedIns, which can be implemented locally. 
\textbf{DeepRare} achieves 58\%, 71\% for Recall@1 and Recall@3, significantly surpassing the other methods.

\subsection{HPO-wise Analysis across Specialties}

In addition to analyzing HPO-wise diagnostic performance across datasets, 
we also present results across different medical specialties, highlighting the system's broad understanding of diverse medical knowledge.

Specifically, we categorized all test cases based on 14 body system specialties, following the taxonomy introduced by MedlinePlus\footnote{\url{https://medlineplus.gov/healthtopics.html}}~\cite{miller2000medlineplus}, namely, \texttt{Blood, Heart and Circulation; Bones, Joints and Muscles; Brain and Nerves; Digestive System; Ear, Nose and Throat; Endocrine System; Eyes and Vision; Immune System; Kidneys and Urinary System; Lungs and Breathing; Mouth and Teeth; Skin, Hair and Nails; Female Reproductive System; and Male Reproductive System}. 
The diseases are categorized by DeepSeek-V3 under Prompt~\ref{prompt3:classification}.

We then present the performance comparison on various specialties. 
It is important to note that each case may involve multiple specialties. 
Similarly, due to privacy concerns regarding the in-house cases,
we only evaluate methods that do not require uploading cases via online LLM APIs. Thus, DeepSeek-V3, DeepSeek-R1, and MedIns are retained to represent general LLMs, reasoning LLMs, and medical LLMs, respectively. 

\begin{figure}[!t]
    \centering
    \includegraphics[width=0.99\linewidth]{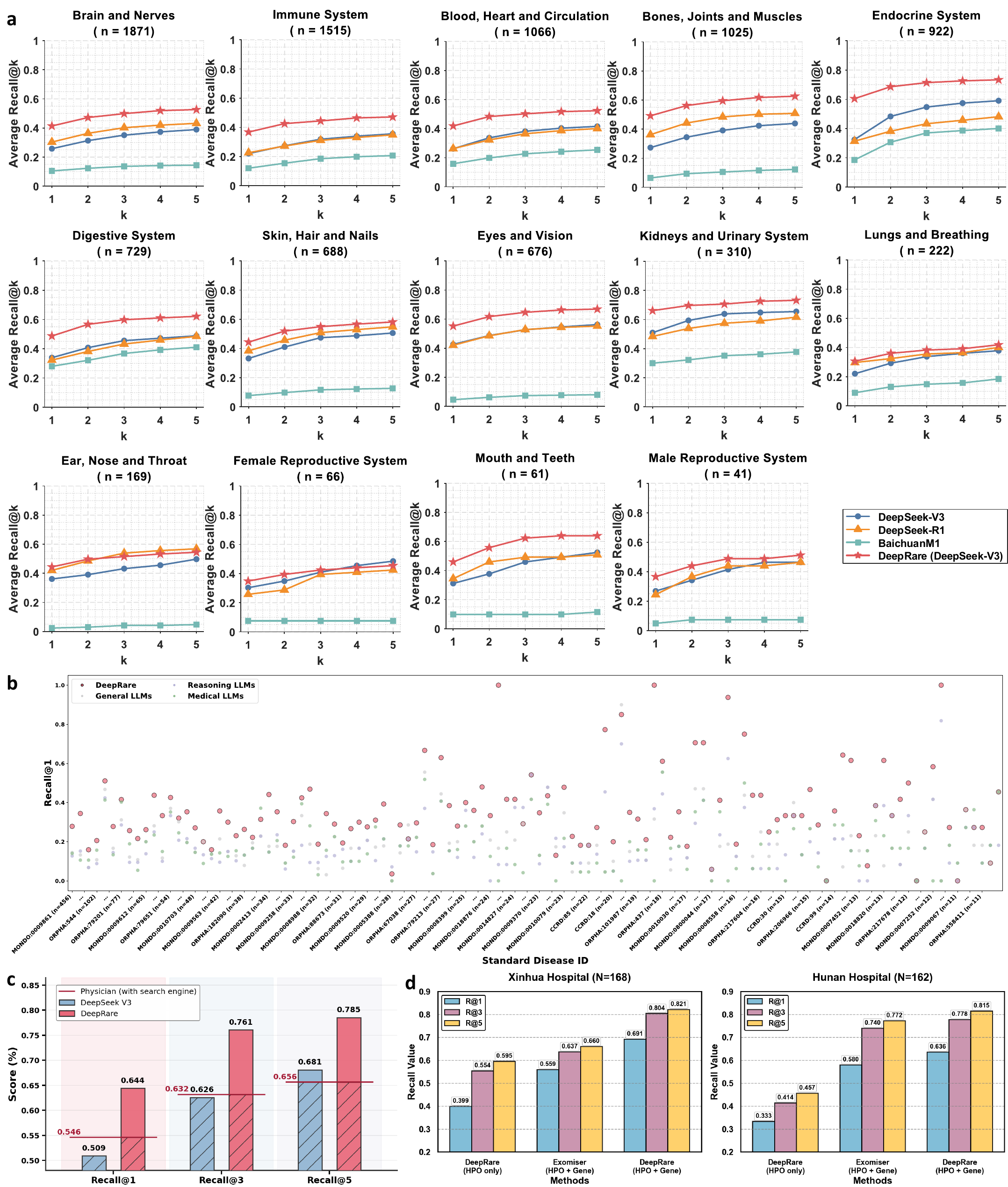}
    \caption{\textbf{DeepRare's diagnostic performance.} \textbf{a} Comparison of diagnostic accuracy across fourteen body systems: showing DeepRare's superior performance in most specialties compared to LLM (DeepSeek-V3), Reasoning LLM (DeepSeek-R1), and Medical LLM (MedIns). \change{\textbf{b} Disease-level recall performance comparison for diseases with $>10$ cases, showing DeepRare's consistent superiority. \textbf{c} Real-world clinical validation study: Diagnostic recall performance comparison of specialized rare disease physicians (10+ years experience with search engine), LLM (DeepSeek V3), and DeepRare using unprocessed outpatient clinical narratives (free-text context only).} \textbf{d} Diagnosis performance with HPO and gene data input compared with baseline method and only HPO input.}
    \label{fig:result2}
\end{figure}

The results are illustrated in Figure~\ref{fig:result2}a,
\textbf{DeepRare} demonstrates substantial performance superiority across almost all specialties. 
For example, in the Endocrine System category, \textbf{DeepRare} achieves a top-1 diagnostic accuracy of 60\%, significantly higher than the second-best method at 32\%. 
Similarly, for 729 cases in the Digestive System category, 
\textbf{DeepRare}'s top-1 diagnostic accuracy reached 49\%, substantially outperforming the second-best method at 34\%.
Notably, this analysis reveals that our \textbf{DeepRare} performs best in the Kidneys and Urinary System, achieving an accuracy of 66\%, while showing relatively lower performance in the Lungs and Breathing System with an accuracy of 31\%, reflecting its clinical application boundaries.

\subsection{\change{HPO-wise Analysis across Diseases}}

\change{To ensure a comprehensive evaluation beyond aggregate metrics, we conducted a fine-grained, disease-level performance analysis. This is crucial for contextualizing the results, as the benchmark encompasses a long-tail of rare diseases with varying case counts. We stratified the 3,134 diseases based on their representation in the test set to determine if performance was uniform or driven by specific subsets.}

\change{For diseases with substantial representation (>10 cases), DeepRare demonstrated a consistent and clear performance advantage. As shown in Figure~\ref{fig:result2}b, the recall@1 for these diseases is consistently higher for DeepRare compared to all baseline models across the spectrum of case volumes, confirming that its efficacy is not an artifact of data-sparse categories.}

\change{We further focused on the most challenging segment of the benchmark: diseases with $\leq10$ cases. Performance on these ``long-tail'' categories is a critical test of a model's generalization capability. The results show that DeepRare achieves a high level of diagnostic accuracy (Recall@1 > 0.8) for 31.8\% of these data-sparse diseases. This significantly outperforms DeepSeek-V3 (23.5\%) and DeepSeek-R1 (26.6\%). Notably, the smaller, domain-specific medical LLM (Baichuan-M1) struggled profoundly in this setting, achieving the same high recall threshold for only 2.5\% of tail-end diseases, highlighting a significant limitation in its ability to generalize. This stratified analysis confirms that DeepRare's superior performance is robust across the case distribution. It excels not only on well-represented diseases but also demonstrates a stronger capacity to diagnose challenging, rare conditions with limited available data, a key requirement for a practical clinical decision support system.}

\subsection{\change{HPO-wise Comparison against Experts}}

\change{To further validate the \textbf{real-world clinical utility} of our approach, we conduct a comparative study using 163 clinical cases from Xinhua Hospital's test dataset (selected from an initial pool of 200, after excluding cases physicians deemed unreasonable due to insufficient information in outpatient narratives). In this setting, DeepRare's diagnostic performance is benchmarked against 5 experienced physicians (with $\geq$10 years of clinical practice in rare diseases). To ensure a fair comparison, both the physicians and DeepRare are provided with the identical input: the structured HPO extracted from the free-text outpatient narratives. The physicians are allow to use search engines and reference materials but were prohibited from using AI tools. They are instructed to list up to five differential diagnoses, providing fewer if they are highly confident.}

\change{As shown in Figure~\ref{fig:result2}c, DeepRare achieve a Recall@5 of 78.5\%, significantly outperforming the clinicians' average accuracy of 65.6\%. Crucially, at Recall@1, DeepRare (64.4\%) surpasses the physicians' performance (54.6\%) for the first time. To the best of our knowledge, this represents a landmark result: DeepRare is the first computational model to surpass the diagnostic performance of human physician experts in the complex task of rare disease phenotyping and diagnosis, providing compelling evidence of its practical value in real-world clinical scenarios.}

\subsection{Analysis on HPO and Gene Data}
To comprehensively evaluate our system's diagnostic capabilities, 
we investigate the performance when incorporating both HPO and genetic data as inputs. We conducted this evaluation on \change{168 cases from the Xinhua Hosp. dataset and 162 cases from the Hunan Hosp. dataset}, specifically selecting cases with complete whole-exome sequencing data to ensure robust comparative analysis. \change{As shown in Figure~\ref{fig:result2}d, 
the integration of genetic information yielded substantial performance improvements, with Recall@1 increasing dramatically from 39.9\% to 69.1\% in the Xinhua Hospital dataset and from 33.3\% to 63.6\% in the Hunan Hospital dataset. 
In addition, we compared our approach with other bioinformatics tools that similarly process both HPO and genetic data, such as Exomiser~\cite{exomiser}, which is also utilized as a component within our system. 
When comparing systems using both HPO and genetic data, our DeepRare system achieved Recall@1 of 69.1\% vs. Exomiser's 55.9\% in the Xinhua Hospital dataset, and 63.6\% vs. 58.0\% in the Hunan Hospital dataset, demonstrating superior performance across both cohorts.} These results demonstrate that our agentic system significantly outperforms existing bioinformatics diagnostic tools in rare disease comprehensive analysis.

\subsection{Traceable Reasoning Chain Validation}

\begin{figure}[!t]
    \includegraphics[width=1\linewidth]{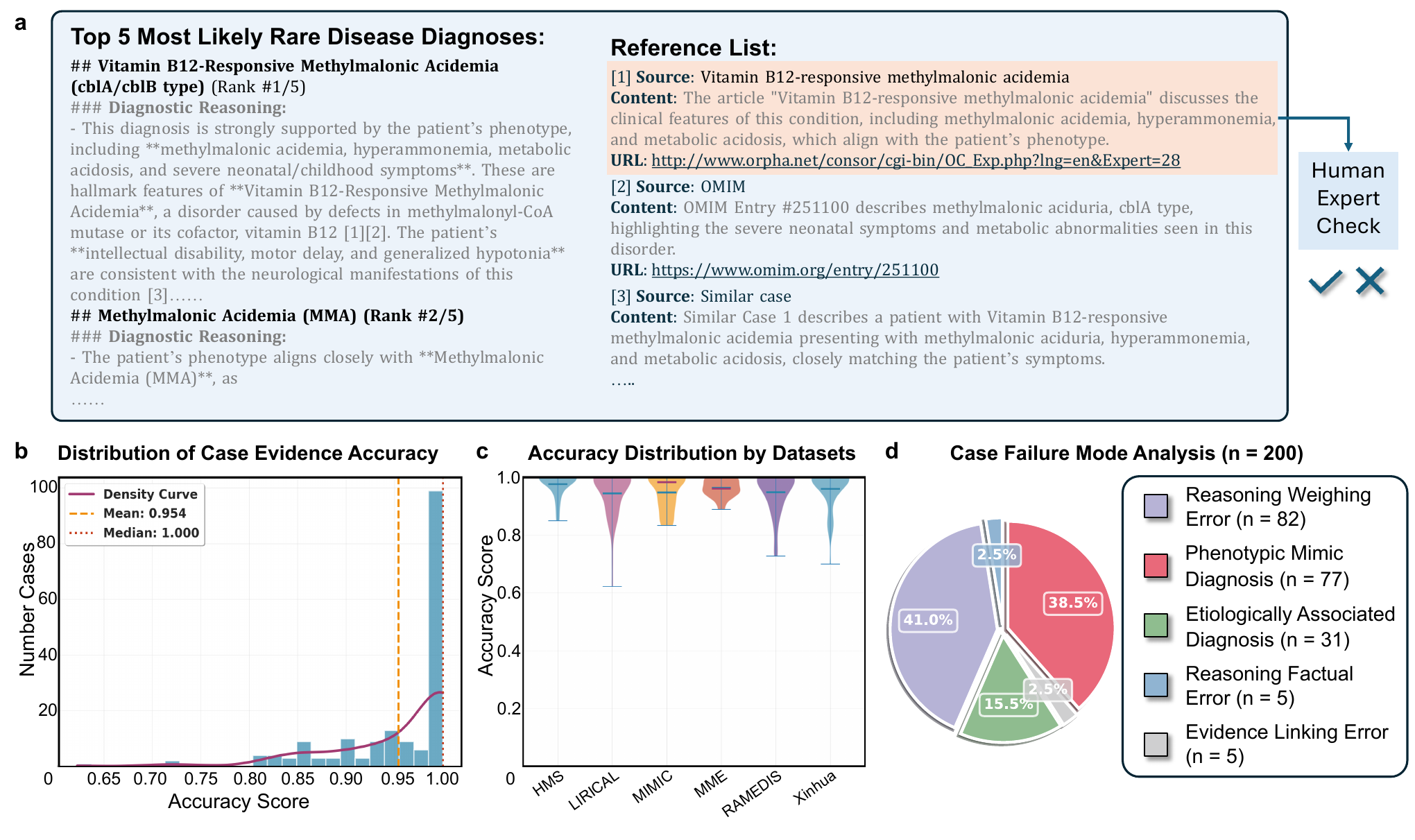}
    \centering
    \caption{\textbf{Human expert validation of traceable reasoning chain and failure mode in DeepRare diagnostic system.} \textbf{a} Representative case output demonstrating differential diagnosis with an evidence-based reference list. \textbf{b} Histogram of reference accuracy scores with density curve (mean = 0.954, median = 1.000). \textbf{c} Dataset-specific accuracy distributions showing robust performance across eight rare disease datasets. \textbf{(d)} Distribution of Failure Modes in randomly sampled 200 failed cases from the entire HPO‑wise test set.}
    \label{fig:trace}
\end{figure}

To assess the reliability and clinical relevance of the reference lists generated by \textbf{DeepRare}, we enlisted 10 associate chief physicians specializing in rare diseases to evaluate the system’s outputs on complex cases. A total of 180 cases were randomly sampled from \textbf{DeepRare}’s predictions across eight datasets. 
Each case was independently reviewed by three specialists, and the consensus was calculated as the mean score. We developed a dedicated annotation interface that presented experts with patient information, model-generated diagnostic results, 
and corresponding reference lists (see Figure~\ref{fig:trace}a). 
Physicians were asked to assess the accuracy of each reference (including literature, case reports, and websites), with accuracy defined as the reference being both reliable and directly relevant to the model’s final diagnostic decision.

Statistical results at the case level (Figure~\ref{fig:trace}b) show an average reference accuracy of 95.4\%.
At the dataset level (Figure~\ref{fig:trace}c), the system consistently demonstrates high performance across all datasets. Further analysis of references deemed incorrect by physicians revealed two main error categories: \textbf{(1) Hallucinated references}, where the system generated plausible but nonexistent URLs in the absence of actual literature links, leading to erroneous webpages; \textbf{(2) Irrelevant references}, resulting from incorrect diagnostic conclusions that caused the model to cite sources unrelated to the true disease.

Overall, physician validation confirms the robustness of \textbf{DeepRare}’s source attribution, highlighting its potential to substantially streamline the literature and case retrieval process during clinical diagnosis and to enhance diagnostic efficiency for healthcare professionals.

\subsection{\change{Failure Cases Analysis}}

\changee{A rigorous analysis of DeepRare's diagnostic failures is performed to delineate its limitations and inform the next stages of development. Detailed examples are presented in Supplementary Section~\ref{sec:failure_examples}. We also perform a quantitative failure mode analysis on 200 randomly sampled cases from the HPO-wise test set where the correct diagnosis is not ranked among the top five recommendations. Each case is independently reviewed and classified by three rare-disease specialists, each with a decade of clinical experience.}

\changee{We categorized failures based on the system's output structure: (i) reasoning processes, (ii) utilization of external evidence, and (iii) the final diagnosis. This yielded five distinct failure modes (Figure~\ref{fig:trace}d).}

\changee{The most prevalent failure mode is \textbf{Reasoning Weighing Error} (41.0\%, 82/200), where the logical structure is sound but the diagnostic weight assigned to specific phenotypes is suboptimal. For instance, DeepRare occasionally overemphasizes non-specific features (e.g., Antinuclear antibody positivity) while underweighting more pathognomonic findings (e.g., significantly low alkaline phosphatase), leading to a plausible but incorrect diagnosis. The second most common mode is \textbf{Phenotypic Mimic Diagnosis} (38.5\%, 77/200), quantitatively substantiating the challenge of high phenotypic overlap. In these cases, DeepRare identifies the correct clinical category but cannot differentiate between molecularly distinct entities based on HPO terms alone. For example, it frequently confuses CTCF-related disorder with its clinical mimic, Cornelia de Lange syndrome. A smaller proportion of errors is classified as \textbf{Etiologically Associated Diagnosis} (15.5\%, 31/200), where the predicted diagnosis is a taxonomically distinct but clinically and pathophysiologically related entity, capturing the fundamental pathology of the case. In contrast, fundamental errors in reasoning (\textbf{Reasoning Factual Error}, 2.5\%) or in using retrieved information (\textbf{Evidence Linking Error}, 2.5\%) are rare, indicating the robustness of the system's core knowledge and retrieval capabilities.}

\subsection{Ablation Study}

In this section, we conduct a comprehensive ablation study on our system design, focusing on central host selection and the effectiveness of introducing various agent designs.

\begin{figure}[!t]
    \centering
    \includegraphics[width=\linewidth]{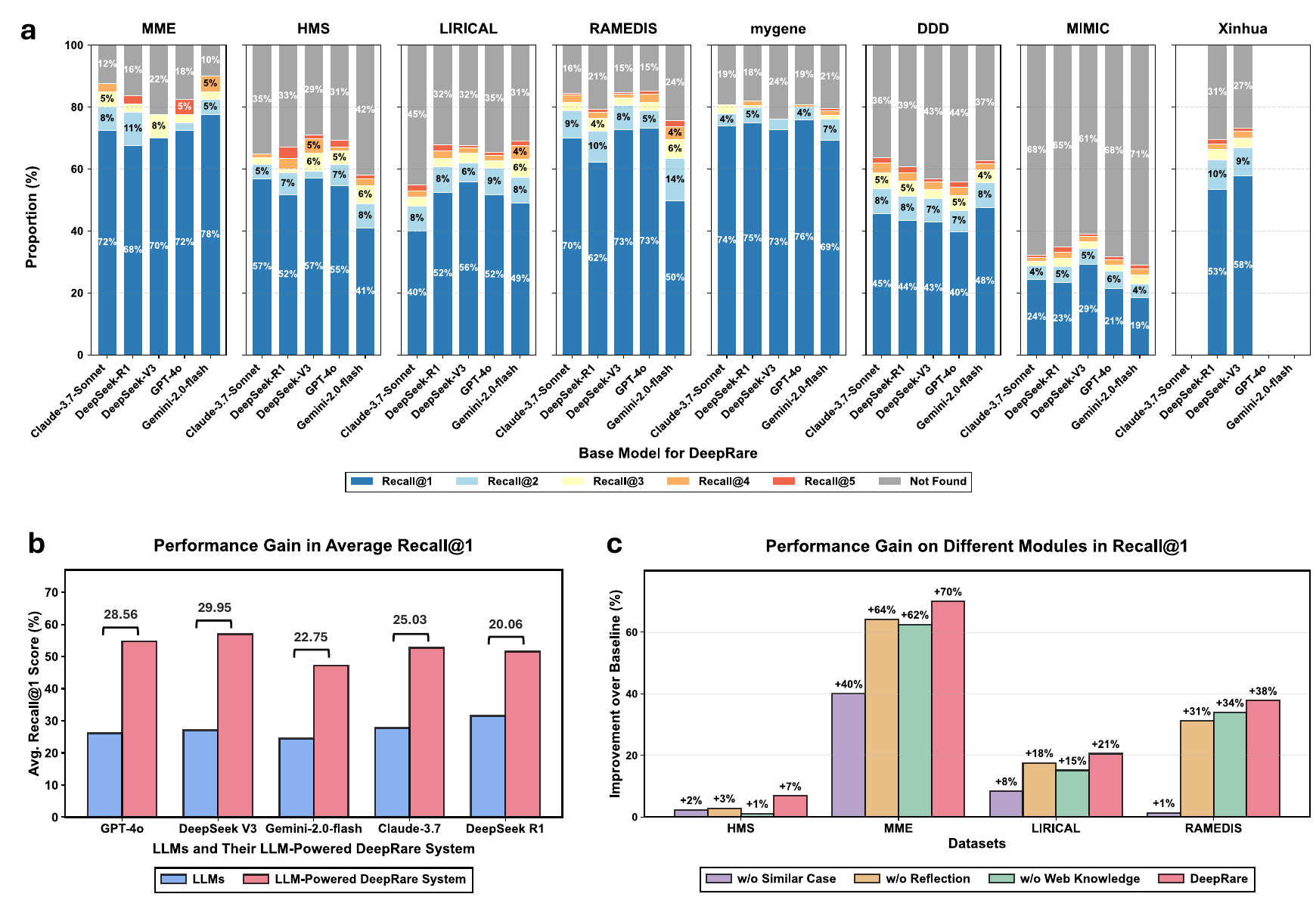}
    \vspace{0.1cm}
    \caption{\textbf{Ablation Study of DeepRare System.} \textbf{a} Performance comparison across different LLMs (Claude-3.7-Sonnet, DeepSeek-R1, DeepSeek-V3, GPT-4o, Gemini-2.0-flash)  as central hosts on eight rare disease datasets.
    \textbf{b} Performance enhancement comparison between baseline large language models and their powered agentic DeepRare systems. 
    \textbf{c} Module-wise contribution analysis on DeepRare system (GPT-4o powered) demonstrating the effectiveness of similar case retrieval, web knowledge integration, and self-reflection components compared to baseline GPT-4o performance.}
    \label{fig:result3}
\end{figure}

To begin, we evaluate \textbf{various foundational models as central hosts for DeepRare}. As illustrated in Figure~\ref{fig:result3}a, we test Claude-3.5-Sonnet, DeepSeek-R1, DeepSeek-V3, GPT-4o, and Gemini-2-Flash across eight datasets, 
assessing their suitability as the core of our agentic system.
According to the results, DeepSeek-V3 outperforms other models on most datasets, except for RareBench~(MME), where the Gemini-2-flash-based agentic system achieves the best performance. Overall, the choice of central host LLMs has minimal impact on the results, highlighting the generalization of our system, which is not reliant on specific LLMs.

In Figure~\ref{fig:result3}b, we compare \textbf{the raw LLMs with their corresponding agentic systems}. As shown in the figure, our agentic design significantly improves the performance of the original LLMs, highlighting the necessity of the proposed agentic workflow mechanisms. For instance, with GPT-4o, the average Recall@1 score across the five public datasets improves substantially from 26.11\% to 54.67\% with 28.56\% performance gain, while for DeepSeek-V3, it increases from 26.99\% to 56.94\% with 29.95\% performance gain. This enhancement is consistently observed across all tested LLMs, demonstrating the effectiveness of our approach.

In Figure~\ref{fig:result3}c and \changee{Table~\ref{tab:rarebench} \&~\ref{tab:zeroshot_performance} in Supplementary~\ref{sec:ablation}}, we show \textbf{the effectiveness of different agentic components}. It illustrates the Recall@1 improvements of each module, including the similar case retrieval, web knowledge, and self-reflection modules, in the DeepRare system (GPT-4o powered) compared to the baseline method (GPT-4o) on the RareBench dataset. \changee{The results confirm that the system's components provide complementary strengths. In contexts with sparse historical precedents, knowledge tools and reflective reasoning sustain robust diagnostic performance. Conversely, in scenarios rich with analogous cases, case-based retrieval delivers substantial gains. Critically, the full DeepRare system, which dynamically orchestrates all components, consistently surpasses any partial configuration. This synergy demonstrates that it adaptively leverages precedent where available and relies on foundational knowledge and verification for novel challenges, thereby achieving state-of-the-art performance across the diverse and long-tail landscape of rare diseases.}

\section{Discussion}

In this study, we present \textbf{DeepRare}, an LLM-powered agentic system specifically designed for rare disease diagnosis. It can process a wide range of input types commonly encountered in clinical workflows, including chief complaints, genetic data, and detailed clinical phenotypes, thereby supporting clinicians in the timely and accurate diagnosis of rare diseases. A key feature of \textbf{DeepRare} is its ability to generate a comprehensive diagnostic reasoning chain, providing transparent and interpretable insights that enhance clinical decision-making. We rigorously evaluated \textbf{DeepRare} across diverse datasets spanning multiple sources, disease categories, and medical centers, where it consistently outperformed existing methods, demonstrating both effectiveness and generalizability.

Compared to existing diagnostic tools commonly used in clinical practice, \textbf{DeepRare} addresses several critical limitations: (i) Traditional HPO-based systems typically generate candidate disease lists without providing sufficient explanatory context or diagnostic rationale, thereby limiting their clinical applicability; 
(ii) Some tools that integrate both HPO phenotypes and genetic data remain highly dependent on genetic testing results, rendering them less suitable for initial patient assessments or first-line screening;
(iii) Recent advances in large language models have improved clinician usability, but these models are still prone to hallucinations that undermine diagnostic reliability. \textbf{DeepRare} overcomes these challenges by grounding its diagnostic reasoning in verifiable medical evidence, ensuring both interpretability and trustworthiness throughout the diagnostic process.

Our experimental results demonstrate two key achievements: (i) \textbf{Superior performance across benchmarks}: \textbf{DeepRare} achieved substantial improvements over existing methods across multiple benchmark datasets, including the publicly available RareBench, a rare disease subset of MIMIC-IV-Note, and our in-house Xinhua Hospital dataset, with comprehensive analyses showing consistent outperformance across different medical specialties and input modalities. (ii) \textbf{Evidence-based reasoning chains}: Beyond diagnostic accuracy, \textbf{DeepRare} provides transparent, step-by-step diagnostic reasoning with verifiable references that significantly reduce clinical decision-making time and minimize patient costs associated with misdiagnosis, as validated through expert clinical assessment.

The clinical implications of \textbf{DeepRare} extend beyond diagnostic accuracy to address fundamental challenges in rare disease care delivery. The system's ability to provide evidence-based reasoning chains with verifiable references could significantly reduce the time required for literature review and case research, enabling clinicians to focus more on patient care rather than information gathering. Furthermore, the system's consistent performance across different medical specialties suggests its potential as a valuable decision support tool for non-specialist physicians who may encounter rare diseases infrequently. This democratization of rare disease expertise could be particularly impactful in resource-limited settings or regions with limited access to specialized care, potentially reducing healthcare disparities in rare disease diagnosis.

\section{Limitations}

While \textbf{DeepRare} demonstrates strong performance and broad applicability, several areas offer opportunities for further enhancement. First, although our current agentic architecture integrates diverse specialized medical resources and databases, it has yet to incorporate potentially valuable data sources fully. Nevertheless, the flexibility of our agentic system, combined with its MCP-like plugin interface, enables seamless future expansion and integration of additional rare disease knowledge systems and bioinformatics tools, further enhancing diagnostic support.

Second, our present knowledge search agent processes phenotypic information in aggregate. While this approach has proven effective, future work could explore more refined and adaptive retrieval mechanisms to further optimize knowledge curation and potentially enhance diagnostic precision.

\change{
Third, our diagnosis system primarily targets cases where patients are aware of rare diseases but have not yet received a precise diagnosis while we also acknowledge that ``screening'' in non-specialist settings to trigger initial suspicion is crucial. Expanding the system with screening represents a promising direction to enhance its clinical impact. Thanks to its MCP-based agentic architecture, our system can be seamlessly integrated with new agent servers to broaden its functional coverage. In Supplementary Material~\ref{sec:expanding_demonsrtation}, we provide a simple attempt to equip the system with preliminary screening ability and encourage further work to advance this line of research.
}

Finally, although we have developed modules for patient interaction to facilitate information gathering, the lack of suitable validation datasets has so far precluded experimental evaluation of this feature. As such datasets become available, further investigation and iterative improvement of patient interaction capabilities will be a natural next step.

Overall, these represent areas for ongoing development rather than fundamental limitations. Future work will focus on expanding the agentic system framework to encompass rare disease treatment and prognosis prediction, with the goal of evolving \textbf{DeepRare} into an even more flexible and comprehensive ecosystem for rare disease management.



\section{Acknowledgments}
This work is supported by the National Key R\&D Program of China (No. 2022ZD0160702), the National Key R\&D Program of China (No. 2022YFC2703400, No. 2025YFC2708201), the Fund for Promoting High-Quality Industrial Development from Shanghai Municipal Commission of Economy and Informatization (2025-GZL-RGZN-01036), Collaborative Innovation Program of the Shanghai Municipal Health Commission (No.2020CXJQ01) and Natural Science Foundation of China (grant number 82130015).
Weidi would like to acknowledge the funding from Scientific Research Innovation Capability Support Project for Young Faculty (ZY-GXQNJSKYCXNLZCXM-I22).
Additionally, we gratefully acknowledge the developers and contributors of publicly available rare disease datasets, foundational research works, bioinformatics tools, and large language models that have collectively enabled our research.

\section{Author Contributions}
All listed authors meet the ICMJE four criteria for authorship. W.Z., C.W., and Y.F. contributed equally to this work. Y.Z., Y.Y., K.S., and W.X. are the corresponding authors.
All authors (W.Z., C.W., Y.F., P.Q., X.Z., Y.S., X.Z., S.Z., Y.P., Y.W., S.X., Y.Z., Y.Y., K.S., and W.X.) contributed to the conception and design of the study. W.Z. and C.W. led the computational algorithm design, while Y.F. led the clinical design and medical aspects. W.Z., C.W., and Y.F. performed data acquisition, analysis, and interpretation. W.Z., C.W., P.Q., X.Z., Y.Z. and W.X. contributed to the technical implementation. W.Z., C.W., and Y.F. contributed to the evaluation framework used in the study. Y.Z., Y.Y., K.S., and W.X. provided technical and infrastructure guidance. Y.F., S.Z., Y.P., S.Z., Y.Y., K.S. provided clinical inputs to the study. All authors contributed to the drafting and revising of the manuscript. All authors approved the final version for publication and agree to be accountable for all aspects of the work, ensuring that questions related to the accuracy or integrity of any part of the work are appropriately investigated and resolved.

\section{Competing Interests}
We declare that the authors have no competing interests as defined by Nature Portfolio, or other interests that might be perceived to influence the results and/or discussion reported in this paper.

\clearpage

\section{Methods}

We introduce \textbf{DeepRare}, an agentic framework designed to support rare disease diagnosis, structured upon a modular, multi-tiered architecture. The system comprises three core components: (i) a central host agent, equipped with a memory bank that integrates and synthesizes diagnostic information while coordinating system-wide operations; 
(ii) specialized local agent servers, each interfacing with specific diagnostic resource environments through tailored toolsets; and (iii) heterogeneous data sources that provide critical diagnostic evidence, including structured knowledge bases ({\em e.g.}, research literature, clinical guidelines) and real-world patient data. The architecture of \textbf{DeepRare} is described in a top-down manner, beginning with the central host's core workflow and proceeding through the agent servers to the underlying data sources.

\subsection{Problem Formulation}

In this paper, we focus on rare disease diagnosis, where the input of a rare disease patient's case typically consists of two components: 
\textbf{phenotype} and \textbf{genotype}, denoted as $\mathcal{I} = \{\mathcal{P}, \mathcal{G}\}$. Either $\mathcal{P}$ or $\mathcal{G}$ (but not both) may be an empty set $\emptyset$, 
indicating the absence of the corresponding input. Specifically, the input phenotype may consist of free-text descriptions $\mathcal{T}$, structured Human Phenotype Ontology (HPO) terms $\mathcal{H}$, or both. Formally, we define, $\mathcal{P} = (\mathcal{T}, \mathcal{H})$, where either $\mathcal{T}$ or $\mathcal{H}$ may be empty (i.e., $\emptyset$), indicating the absence of that input modality. The \textbf{genotype input} denotes the raw Variant Call Format (VCF) file generated from Whole Exome Sequencing~(WES).

Given $\mathcal{P}$, the goal of the system is to produce:
a ranked list of the top $K$ most probable rare diseases, 
$\mathcal{D} = \{d_1, d_2, \dots, d_K\}$, and a corresponding rationale $\mathcal{R}$, consisting of evidence-grounded explanations traceable to medical sources such as peer-reviewed literature, clinical guidelines, and similar patient cases. This can be formalized as:
\begin{equation}
    \{\mathcal{D}, \text{\hspace{3pt}} \mathcal{R}\} = \mathcal{A}(\mathcal{P}),
\end{equation}
where $\mathcal{A}(\cdot)$ denotes the diagnostic model. 

As shown in Extended Data  Figure~\ref{fig:Architecture}b, our multi-agent system comprises three main components:
\vspace{-8pt}
\begin{itemize}
\setlength\itemsep{0.15cm}

    \item \textbf{A central host with a memory bank} serves as the coordinating brain of the system. The memory bank is initialized as empty and incrementally updated with information gathered by agent servers. Powered by a LLM, the central host integrates historical context from the memory bank to determine the system’s next actions.
    
    \item \textbf{Multiple agent servers} execute specialized tasks such as phenotype extraction and knowledge retrieval, enabling dynamic interaction with external data sources.

    \item \textbf{Diverse data sources} serve as the external environment, providing crucial diagnostic evidence from PubMed articles, clinical guidelines, publicly available case reports, and other relevant resources.
\end{itemize}

\subsection{Main Workflow}
The system operates in two primary stages, orchestrated by the central host: \textbf{information collection} and \textbf{self-reflective diagnosis}, as illustrated in Extended Data Figure~\ref{fig:Architecture}c. For clarity, the specific functionalities of the agent servers involved in each stage are detailed in the following section.

\subsubsection*{Information Collection}

In the information collection stage, the system preprocesses the patient input and invokes specialized agent servers to gather relevant medical evidence from external sources. The process begins with two parallel steps: 
one focusing on phenotype inputs and the other on genotype data. 
Subsequently, the central host takes control to facilitate diagnostic decision-making and patient interaction.

\textit{Phenotype Information Collection:} 
Given the phenotype input~($\mathcal{P} = (\mathcal{T}, \mathcal{H})$), 
the system performs the three main sub-steps to collect extra information: 
\textbf{HPO standardization}, \textbf{phenotype retrieval}, and \textbf{phenotype analysis}.

In HPO standardization, the phenotype extractor $a_{\text{hpo}}$ agent server is called, 
to convert the given free-form reports $\mathcal{T}$ into a list of standardized entities $\mathcal{H}$, denoted as: 
\begin{equation}
    \hat{\mathcal{P}} =\begin{cases}
    \text{ } a_{\text{hpo}}(\mathcal{T}), & \text{if } \mathcal{T} \neq \emptyset.\\
    \text{ } \mathcal{H}, & \text{otherwise}.
  \end{cases}
\end{equation}
As a result, each patient is now denoted as a set of standardized HPO entities~($\hat{\mathcal{P}}$), that are further treated as the query for phenotype retrieval. 

The {knowledge searcher}~($a_\text{k-search}$) and {case searcher}~($a_\text{c-search}$)
agent servers are invoked to retrieve supporting documents from the web and relevant cases from an external database, respectively:
\begin{equation}
    \mathcal{E}_{\text{hpo}} = a_\text{k-search}(\hat{\mathcal{P}}, \mathcal{M}, N) \cup a_\text{c-search}(\hat{\mathcal{P}},\mathcal{M}, N),
\end{equation}
where $\mathcal{E}_{\text{hpo}}$ refers to a unified set of retrieved evidences and $N$ denotes the search depth. Notably, the two search agents will also check the memory bank~($\mathcal{M}$) to avoid retrieving items that have already been recorded.

Lastly, in phenotype analysis, the agent server integrates various distinct bio-informatics tools to provide a set of diagnostic-related suggestions, 
for example, identifying diseases that are more likely to be associated with the patient based on their phenotype, denoted as:
\begin{equation}
    \mathcal{Y}_{\text{hpo}} = a_\text{hpo-analyzer}(\hat{\mathcal{P}}).
\end{equation}
Till here, we have gathered relevant information on the phenotype by exploring the web, 
or database with similar cases, and multiple existing bioinformatics analysis tools, collectively denoted as $(\mathcal{E}_{\text{hpo}}, \mathcal{Y}_{\text{hpo}})$ and update them into the system memory bank $\mathcal{M}$, denoted as:
\begin{equation}
    \mathcal{M} \gets \mathcal{M} \cup(\mathcal{E}_{\text{hpo}}, \mathcal{Y}_{\text{hpo}}),
\end{equation}

Following the phenotype analysis, the central host generates a tentative diagnosis based on the available phenotype information:
\begin{equation}
\mathcal{D}^{\prime} = \mathcal{A}_\text{host}(\hat{\mathcal{P}}, \mathcal{M} \text{ | } \texttt{<prompt>}_{\ref{prompt4:first_diag}})
\end{equation}
where $\texttt{<prompt>}_{\ref{prompt4:first_diag}}$ is a diagnosis-related prompt instruction to drive the central host. 
The output $\mathcal{D}^{\prime}$ is an initialized rare disease list.

\textit{Genotype Information Collection:} 
In parallel to the phenotype analysis, the system will also collect external information relevant to the genotypes, if provided~($\mathcal{G} \ne \emptyset$). It also consists of three main sub-steps: 
\textbf{VCF annotation}, \textbf{variant ranking}, and \textbf{synthetic analysis}. The first two steps are conducted by the genotype analyzer agent server, and the last one is processed by the central host.

In VCF annotation, the goal is to annotate the raw VCF input, which often contains thousands of gene variants using various genomic databases. This process enriches each variant with comprehensive functional annotations, population frequencies, and pathogenicity predictions. Subsequently, variant ranking is performed to prioritize variants based on their potential clinical significance. This step applies scoring algorithms that consider multiple factors, including functional impact, allele frequencies, conservation scores, and predicted pathogenicity:
\begin{equation}
\hat{\mathcal{G}} = a_{\text{geno-analyzer}}(\mathcal{G})
\end{equation}
where $\tilde{\mathcal{G}}$ denotes the ranked set of variants ordered by their clinical relevance scores.

Finally, in synthetic analysis, the central host uses LLM to interpret the ranked variants and patient information, providing comprehensive variant interpretation, gene-phenotype association predictions, and inheritance pattern analysis:

\begin{equation}
    \mathcal{D}^{\prime\prime} = \mathcal{A}_\text{host}(\hat{\mathcal{G}}, \mathcal{M}, \mathcal{D}^{\prime}, N \text{ | } \texttt{<prompt>}_{\ref{prompt3.5:gene}})
\end{equation}

where $\texttt{<prompt>}_{\ref{prompt3.5:gene}}$ is synthetic analysis prompt. 
If genotype data is not available ($\mathcal{G} = \emptyset$), the system uses the phenotype-only diagnosis:

\begin{equation}
\mathcal{D}^{\prime\prime} = 
    \mathcal{D}^{\prime}, \  \text{if } \mathcal{G} = \emptyset
\end{equation}

where $\mathcal{D}^{\prime\prime}$ represents the updated diagnosis list after incorporating genotype information (when available). The results are then consolidated and updated in the system memory bank:

\begin{equation}
\mathcal{M} \gets \mathcal{M} \cup \mathcal{D}^{\prime\prime}
\end{equation}

\subsubsection*{Self-reflective Diagnosis}

At this stage, the central host takes entire system control and proceeds to self-reflective diagnosis, which attempts to make a diagnosis based on all previously collected information.

Specifically, the central host will make a tentative diagnosis decision-making, defined as:
\begin{equation}
     \mathcal{D}^{\prime} = \mathcal{A}_\text{host}(\hat{\mathcal{I}}, \mathcal{M} \text{ | } \texttt{<prompt>}_{\ref{prompt4:first_diag}})
\end{equation}
where $\hat{\mathcal{I}} = \{\hat{\mathcal{P}}, \hat{\mathcal{G}}, \mathcal{T}\}$ denotes the preprocessed patient input and $\texttt{<prompt>}_{\ref{prompt4:first_diag}}$ is a diagnosis-related prompt instruction to drive the central host. \change{Notably, here, in organizing the patient information, we retain the original free-form patient reports, $\mathcal{T}$, thereby compensating for the loss of temporal and contextual details that may be overlooked during HPO extraction when constructing $\hat{\mathcal{P}}$.}
The output $\mathcal{D}^{\prime}$ is an initialized rare disease list. 

Then the system executes self-reflection that double-checks the disease list by collecting additional knowledge related to the predicted disease list from the Internet. This process involves two sub-steps: 
\textbf{disease standardization} and \textbf{disease retrieval}.
In disease standardization, the disease normalizer agent server is invoked here, which converts the disease name list into items from Orphanet or OMIM. Subsequently, in disease retrieval, the knowledge searcher~($a_\text{k-search}$) is called, which treats the standardized disease items as queries and retrieves the relevant knowledge documents for each disease. 
The process can be formulated as:
\begin{align}
    \mathcal{M} \gets \mathcal{M}\cup \mathcal{E}_{\text{disease}} 
    \text{, \hspace{8pt} } \mathcal{E}_{\text{disease}} = a_{\text{k-search}}( 
    a_\text{d-norm}(\mathcal{D^{\prime\prime}}), \mathcal{M}),
\end{align}
where $\mathcal{E}_{\text{disease}}$ denotes the collected disease-related knowledge and will also be stored into the memory bank.

After acquiring the external disease-specific knowledge, 
the central host self-reflects on the correctness of the predicted diseases, 
by synthesizing all collected information. This process is formulated as:
\begin{equation}
    \mathcal{D} = \mathcal{A}_\text{host}(\mathcal{D}^{\prime\prime}, \hat{\mathcal{I}}, \mathcal{M} \mid\texttt{<prompt>}_{\ref{prompt5:reflection}}),
\end{equation}
where $\texttt{<prompt>}_{\ref{prompt5:reflection}}$ is a self-reflection prompt, and $\mathcal{D} = \{d_1, d_2, \dots\}$ represents the ranked list of possible rare diseases, ordered by their likelihood. Notably, if $\mathcal{D} = \emptyset$, 
{\em i.e.}, all proposed rare diseases are ruled out during self-reflection, the system will return to the beginning and increase $N$ by $\Delta N$, re-collect new patient-wise information, and iterate through the entire program workflow until $\mathcal{D} \neq \emptyset$ is satisfied.

Once the system passes the former self-reflection step, 
the central host will further synthesize the collected information, and provide traceable, transparent rationale explanations:
\begin{equation}
    \{\mathcal{D}, \mathcal{R}\} = \mathcal{A}_\text{host}(\mathcal{D}, \hat{\mathcal{I}}, \mathcal{M} \mid \texttt{<prompt>}_{\ref{prompt6:final}}),
\end{equation}
where $\mathcal{R}$ denotes the rationale explanation for each output rare disease, organized as free-text by the central host. This ensures that the final diagnosis is not only accurate but also interpretable, offering users a clear and auditable justification for the predicted diseases. \change{Notably, $\mathcal{R}$ provides accessible reference links to enable traceable reasoning. Before producing the final output, we apply a post-processing step, namely reference link verification, which checks the validity of each URL and removes any invalid ones from $\mathcal{R}$ to mitigate hallucination (implementation details are provided in the supplementary materials~\ref{sec:website_verification}).}

\subsection{Agent Servers}

Agent servers form the second tier of our \textbf{DeepRare system}, each manages one or multiple specific tools, interacting with a specialized working environment to gather evidence from external data sources. In specific, the following agent servers are utilized in our system: \texttt{phenotype extractor}, \texttt{disease normalizer}, \texttt{knowledge searcher}, \texttt{case searcher}, \texttt{phenotype analyzer}, and \texttt{genotype analyzer}.
    
\subsubsection*{Phenotype Extractor}
The clinical rare disease diagnosis procedure requires converting patients' phenotype consultation records~($\mathcal{T}$) into standardized HPO items. Specifically, we extract potential phenotype candidates and modify the phenotype name by prompting an LLM:
\begin{equation}
    \mathcal{H} = \Phi_{\text{LLM}}( \Phi_{\text{LLM}}(\mathcal{T} \mid \texttt{<prompt>}_{\ref{prompt7:hpo}}) \mid \texttt{<prompt>}_{\ref{prompt8:hpo}})
\end{equation}
where $ \mathcal{H} = \{h_1, h_2, \dots\}$ denotes the set of extracted, unnormalized HPO candidate entities, and $\texttt{<prompt>}_{\ref{prompt7:hpo}}$ $\texttt{<prompt>}_{\ref{prompt8:hpo}}$ represents the corresponding prompt instruction. Here, the two-step reasoning process is designed to extract more accurate phenotypic descriptions with LLM assistance, significantly reducing the probability of errors in the subsequent step.

Subsequently, we perform named-entity normalization leveraging BioLORD~\cite{remy2022biolord}, a BERT-based text encoder, to map these candidate entities to standardized HPO terms. Specifically, we compute the cosine similarity between the text embeddings of the predicted entity name and all standardized HPO term names. The top-matching HPO term is then selected to represent the entity. Notably, if no HPO term achieves a cosine similarity of $0.8$ or above, the entity is discarded. \change{To assess the effectiveness of this module, we provide a detailed comparison of phenotype extraction methods in the supplementary materials~\ref{sec:ablation_study}.}

\subsubsection*{Disease Normalizer}
During the diagnostic process, free-text diagnostic diseases are mapped to standardized Orphanet or OMIM items, as more precise keywords for subsequent searches. Similar to that in the phenotype extractor, we use BioLORD~\cite{remy2022biolord} to perform named-entity normalization, by computing the cosine similarity between the text embeddings of the predicted disease name and all standardized disease names listed in the Orphanet or OMIM. The top-matched standardized disease name is then used, and if all standardized disease names cannot match the predicted term (cosine similarity less than $0.8$), the predicted disease will be discarded.

\subsubsection*{Knowledge Searcher}

The knowledge searcher is tasked with real-time knowledge document searching, interacting with external medical knowledge documents and the Internet, supporting the diagnosis system with the latest rare disease knowledge.

While it is invoked, it will perform two distinct search modules with multiple searching tools, on a specific search query~($\mathcal{Q}$), for example, HPO or predicted diseases: 

\vspace{-5pt}
\begin{itemize}
\setlength\itemsep{0.15cm}
    \item \textbf{General web search}: 
    This part executes the general search engines, 
    including Bing~\cite{microsoft2009bing}, Google~\cite{google1998search}, and DuckDuckGo~\cite{weinberg2008duckduckgo}. 
    We will call them one by one, following the listed order. Each time, the top-$N$ web pages (with a default value of $N=5$) will be retrieved. Specifically, Bing\footnote{\url{https://www.bing.com}} is accessed through automated browser simulation using Selenium, while Google and DuckDuckGo are queried via their official APIs\footnote{\url{https://developers.google.com/custom-search/v1/overview}}~\footnote{ \url{https://api.duckduckgo.com}}. If a search engine successfully completes the execution, the process will stop immediately.

    \item \textbf{Medical domain search}: 
    Considering that some professional medical-specific web pages may not be ranked highly in general search engines, 
    this part retrieves information from well-known medical databases. 
    The following search engines are considered: 
    \begin{itemize}
        \item \textbf{Up-to-dated Academic literatures}, including PubMed~\cite{macleod2002pubmed}(accessed via PubMedRetriever from langchain\_community\footnote{\url{https://python.langchain.com/api_reference/community/retrievers/langchain_community.retrievers.pubmed.PubMedRetriever.html}}), 
    and Crossref~\cite{crossref2025} (queried through official API\footnote{\url{https://api.crossref.org/swagger-ui/index.html}}); 
    \item \textbf{Rare disease-specific knowledge bases} such as Orphanet~\cite{weinreich2008orphanet}, OMIM~\cite{amberger2015omim}, and HPO~\cite{HPO} (all utilizing offline knowledge bases and accessed through retrieval mechanisms);
    \item \textbf{General medical knowledge repositories}: Wikipedia~\cite{wikipedia2023} (accessed via WikipediaRetriever from langchain\_community\footnote{\url{https://python.langchain.com/api_reference/community/retrievers/langchain_community.retrievers.wikipedia.WikipediaRetriever.html}}) and MedlinePlus\footnote{\url{https://medlineplus.gov/}}~\cite{medlineplus} (accessed through automated browser simulation using Selenium). 
    \end{itemize}
    Similarly, while searching academic papers and rare disease-specific knowledge bases, we retrieve the top-$N$ web pages (with $N$ defaulting to 5) from each source. The search engines are queried one by one, and the process stops upon successful execution.
\end{itemize}
These tools retrieve web pages and return them to the knowledge searcher, and are summarized by the agent server with a lightweight language model~(GPT-4o-mini by default), to simultaneously extract key information and filter relevant content. This integrated processing pipeline can be formalized as:
\begin{equation}
\mathcal{R} = \Phi_{\text{LLM}}(\text{document}, \mathcal{Q} \mid \texttt{<prompt>}_{\ref{prompt9:web_summarize}}),
\end{equation}

where $\mathcal{R}$ represents the processed output for each retrieved document, $\mathcal{Q}$ denotes the given search queries, and $\texttt{<prompt>}_{\ref{prompt9:web_summarize}}$ is the unified prompt instruction that simultaneously governs both summarization and relevance filtering. The system employs a binary classification approach: medical-related documents are retained and translated into the target language, while non-medical content is rejected with the output \textit{"Not a medical-related page"}.

\subsubsection*{Case Searcher}
Inspired by clinical practice, where physicians often refer to publicly discussed cases when faced with rare or challenging patients, 
the case searcher agent is designed to explore an external case bank. 
Each patient in the database is represented as a list of HPO terms, transforming case search into an HPO similarity matching problem. 
Using an input HPO list ($\mathcal{H}$) from the query case, we implement a two-step retrieval method to interact with this external database:
(i) \textbf{Initial retrieval}: we employ OpenAI's text-embedding model (\texttt{text-embedding-3-small}) to encode both the query HPO list and each candidate patient's HPO representation into dense vector embeddings. The embeddings for all candidate patients in the case database have been pre-computed and stored using the same embedding model. We then identify the top-50 candidate patients based on cosine similarity between these embeddings.
(ii) \textbf{Re-ranking}: We further re-rank these candidates using MedCPT-Cross-Encoder~\cite{jin2023medcpt}, a BERT-based model specifically trained on PubMed search logs for biomedical information retrieval. This model computes refined cosine similarity scores between the query case's HPO profile and each candidate's HPO profile, leveraging domain-specific medical knowledge to improve matching accuracy.

We also evaluated alternative retrieval strategies, 
including single-stage methods with different embedding models such as BioLORD and MedCPT, as well as traditional approaches like BM25~\cite{robertson2009probabilistic}, discussed in the results section. Experimental findings indicate that the two-stage retrieval approach outperforms all alternatives, optimizing both computational efficiency and clinical relevance of the retrieved cases.

Similar to the knowledge searcher, after receiving the similar cases, the case searcher will further assess their relevance to prevent misdiagnosis from irrelevant cases, powered by the lightweight language model:
\begin{equation}
 r_{\text{case}} = \Phi_{\text{LLM}}(\text{Case}, \mathcal{H} \mid \texttt{<prompt>}_{\ref{prompt10:case_summarize}}),
\end{equation}
where $r_{\text{case}} \in \{\text{True}, \text{False}\}$, a binary scalar that
indicates whether the case is related to the given HPO list, 
and $\texttt{<prompt>}_{\ref{prompt10:case_summarize}}$ is the corresponding prompt instruction. Consistency is maintained by employing the same LLM architecture used in the diagnostic process for this assessment.

\subsubsection*{Phenotype Analyzer}
This agent server controls various professional diagnosis tools, 
that are developed for phenotype analysis. By integrating the analysis results from these tools into the overall diagnostic pipeline, 
our system is enabled to incorporate more professional and comprehensive suggestions. Specifically, given the patient HPO list~($\mathcal{H}$), the following tools are used:
\vspace{-5pt}
\begin{itemize}
\setlength\itemsep{0.15cm}
    \item \textbf{PhenoBrain}~\cite{mao2025phenotype}:
    This is a tool for HPO analysis, that takes structured HPO items as input~($\hat{\mathcal{H}}$), and output 5 potential rare disease suggestions. We adopt it by calling its official API\footnote{https://github.com/xiaohaomao/timgroup\_disease\_diagnosis/tree/main/PhenoBrain\_Web\_API}.
    
    \item \textbf{PubcaseFinder}~\cite{Pubcase2}:
    It performs HPO-wise diagnostic analysis by matching the most similar public cases from PubMed case reports. Similarly, it takes the structured HPO items~($\hat{\mathcal{H}}$) as input and returns top-5 potential rare disease suggestions, each with a confidence score. We access it via its official API\footnote{https://pubcasefinder.dbcls.jp/api}.

    \item \textbf{Zero-shot LLM Inference}: We additionally employ LLMs to perform zero-shot preliminary reasoning. Given the extensive knowledge base acquired during LLM training, these models can often suggest candidate diagnoses that conventional diagnostic tools might overlook. Specifically, this approach takes the structured HPO items ($\hat{\mathcal{H}}$) as input and returns the top-5 potential rare disease candidates under Prompt \ref{prompt11:zero_shot_diag}.
\end{itemize}

\subsubsection*{Genotype Analyzer}
Similar to the phenotype analyzer, the genotype analyzer is tasked with performing professional genotype analysis by calling existing tools. 

For patient genomic variant files (aligned to GRCh37 reference genome), we initially subjected the HPO phenotype terms $\mathcal{H}$ and corresponding VCF files $\mathcal{G}$ to comprehensive annotation and prioritization analysis using the \textbf{Exomiser}~\cite{exomiser} framework, with configuration parameters detailed in the supplementary materials~\ref{sec:exomiser}, which is configed to integrate multiple data sources and analytical steps: population frequency filtering utilizing databases including gnomAD~\cite{chen2024genomic}, 1000 Genomes Project~\cite{siva20081000}, TOPMed~\cite{taliun2021sequencing}, UK10K~\cite{uk10k2015uk10k}, and ESP~\cite{tennessen2012evolution} across diverse populations; pathogenicity assessment through PolyPhen-2~\cite{adzhubei2010method}, SIFT~\cite{ng2003sift}, and MutationTaster~\cite{schwarz2010mutationtaster} prediction algorithms; variant effect filtering to retain coding and splice-site variants while excluding intergenic and regulatory variants; inheritance mode analysis supporting autosomal dominant/recessive, X-linked, and mitochondrial patterns; and gene-disease association prioritization through OMIM~\cite{amberger2015omim} and HiPhive~\cite{smedley2015next} algorithms that leverage cross-species phenotype data.

The Exomiser output is ranked according to the composite exomiser\_score, from which we selected the top-n candidate genes while preserving essential metadata including OMIM identifiers, phenotype\_score, variant\_score, statistical significance (p\_value), detailed variant\_info, ACMG pathogenicity classifications, ClinVar annotations, and associated disease phenotypes. The curated genomic annotations were subsequently transmitted to the Central Host for downstream processing and integration.

\vspace{0.4cm}
All outputs from the specialized tools are then transformed into free texts by the agent server, that can be seamlessly combined with the LLM-based central host or other LLM-driven tools. 
This is achieved by employing a predefined templates tailored to each tool's specific output format, such as ``[Tool Name] identified [Disease]'' (with confidence scores included when available) for disease predictions.

\subsection{External Data Sources}
The external data sources form the third tier of our \textbf{DeepRare} framework, providing a comprehensive external environment for tool interaction. These diverse, rare disease-related information sources support the system with professional medical knowledge, we specifically consider the medical-focused databases.

\textbf{Medical Literature.} 
Scientific publications are essential for evidence-based diagnosis, especially for rapidly evolving rare diseases. 
DeepRare accesses peer-reviewed literature through:
\begin{itemize}
\setlength\itemsep{0.15cm}
    \item \textbf{PubMed database}~\cite{macleod2002pubmed}: 
    The world largest database of biomedical literature containing over 34 million papers.
    \item \textbf{Google Scholar}~\cite{googlescholar2004}: A broad academic search engine covering publications across diverse sources.
    \item \textbf{Crossref}~\cite{crossref2025}: A comprehensive metadata database that enables seamless access to scholarly publications and related fields through persistent identifiers and open APIs.
\end{itemize}

\textbf{Rare Disease Knowledge Sources.} 
Curated repositories that aggregate structured information about rare diseases:
\begin{itemize}
\setlength\itemsep{0.15cm}
    \item \textbf{Orphanet}~\cite{weinreich2008orphanet}: Comprehensive information for over 6,000 rare diseases, including descriptions, genetics, epidemiology, diagnostics, and treatments, etc.
    \item \textbf{OMIM} (Online Mendelian Inheritance in Man)~\cite{amberger2015omim}: 
    A catalog of human genes and genetic disorders, documenting over 17,000 genes and their associated phenotypes.
    \item \textbf{Human Phenotype Ontology}~\cite{kohler2021human}: 
    A standardized vocabulary of phenotypic abnormalities in human diseases, containing over 18,000 terms and more than 156,000 hereditary disease annotations.
\end{itemize}

\textbf{General Knowledge Sources.} 
Broad clinical resources that provide contextual understanding:
\begin{itemize}
\setlength\itemsep{0.15cm}

     \item \textbf{MedlinePlus}~\cite{medlineplus}: A US National Library of Medicine resource providing reliable, up-to-date health information for patients and clinicians.
    \item \textbf{Wikipedia}~\cite{wikipedia2023}: General encyclopedia entries on all general knowledge, including medical conditions and rare diseases.
    \item \textbf{Online websites}: Resources accessible through search engines that provide up-to-date information, including medical news portals, patient advocacy groups, research institution websites, and clinical trial registries that may contain the latest developments not yet published in scholarly literature.
\end{itemize}

\textbf{Case Collection.} A large-scale case repository is constructed from multiple data sources to serve as the database for the case search agent server, with a subset of the data reserved as a test set to validate model performance. \changee{Specifically, the rare disease case bank comprises \textbf{67,795 cases} from published literature (49,685), public datasets (13,265), and de-identified proprietary cases (4,845). To ensure fair evaluation, we implement rigorous de-duplication, excluding any identical matches between query cases and the case bank.} 

\begin{itemize}
\setlength\itemsep{0.15cm}
\item \textbf{RareBench}~\cite{chen2024rarebench} is a benchmark designed to systematically evaluate LLM capabilities across four critical dimensions in rare disease analysis. We utilize Task 4 (Differential Diagnosis among Universal Rare Diseases), specifically its public subset comprising 1,114 patient cases collected from four open datasets: MME (The Matchmaker Exchange), HMS, LIRICAL, and RAMEDIS. MME and LIRICAL cases are extracted from published literature and manually verified. HMS contains data from the outpatient clinic at Hannover Medical School in Germany. RAMEDIS comprises rare disease cases autonomously submitted by researchers.

\item \textbf{Mygene2}~\cite{mygene2}, a data-sharing platform connecting families with rare genetic conditions, clinicians, and researchers, provided additional data. We use preprocessed data (146 patients spanning 55 MONDO diseases) from~\cite{alsentzer2022few}, which extracted phenotype-genotype information as of May 2022, limited to patients with confirmed OMIM disease identifiers and single candidate genes to ensure diagnostic accuracy. 

\item \textbf{DDD} (the Deciphering Developmental Disorders Study)~\cite{yates2024curating} data were obtained from the Gene2Phenotype (G2P) project, which curates gene-disease associations for clinical interpretation. We downloaded phenotype terms and associated gene sets from the G2P database\footnote{https://ftp.ebi.ac.uk/pub/databases/gene2phenotype/G2P\_data\_downloads/2025\_05\_28/} in May 2025. After preprocessing to remove cases with missing diagnostic results or phenotypes, the final DDD cohort comprised 2,283 cases.

\item \textbf{MIMIC-IV-Note}~\cite{mimic-iv-note} contains 331,794 de-identified discharge summaries from 145,915 patients admitted to Beth Israel Deaconess Medical Center in Boston, Massachusetts. 
Since our focus is exclusively on rare diseases, 
we first determine whether the case involved a rare disease,
by prompting GPT-4o with the ICD-10~\cite{ICD10} codes associated with each note. Confirmed cases were mapped to our rare disease knowledge base using a methodology similar to disease normalization, while unmapped cases were discarded, resulting in a final dataset of 9,185 records.

\item \textbf{Xinhua Hosp. Dataset~(in-house)} encompasses all rare disease diagnostic records from 2014 to 2025, totaling 352,424 entries. 
Using a procedure similar to our MIMIC processing workflow, 
we applied GPT-4o and vector matching to eliminate records without definitive diagnoses or significant data gaps. We also consolidated multiple consultations for the same patient, resulting in a curated dataset of 5,820 records.

\item \textbf{PMC-Patients}~\cite{PMC-Patient} 
comprises 167,000 patient summaries extracted from case reports in PubMed~\cite{macleod2002pubmed}. The RareArena GitHub~\cite{zhao2025rarearena} repository has processed this dataset with GPT-4o for rare disease screening. 
We therefore utilized their preprocessed dataset, which contains 69,759 relevant records. 

\end{itemize}

\textbf{Gene Variant Databases.} 
Specialized repositories that support the analysis of genetic findings in rare disease diagnosis:
\vspace{-5pt}
\begin{itemize}
\setlength\itemsep{0.15cm}
    \item \textbf{ClinVar}~\cite{landrum2016clinvar}: A freely accessible database containing 1.7 million interpretations of clinical significance for genetic variants, with particular value for identifying pathogenic mutations in rare disorders.
    \item \textbf{gnomAD} (Genome Aggregation Database)~\cite{chen2024genomic}: A resource of population frequency data for genetic variants from over 140,000 individuals, essential for distinguishing rare pathogenic variants from benign population polymorphisms.
    \item \textbf{1000 Genomes Project}~\cite{siva20081000}: A database of human genetic variation across diverse populations worldwide.
    \item \textbf{ExAC} (Exome Aggregation Consortium)~\cite{karczewski2017exac}: A database of exome sequence data from over 60,000 individuals.
    \item \textbf{TOPMed} (Trans-Omics for Precision Medicine)~\cite{taliun2021sequencing}: A national heart, lung, blood, and sleep disorders research program providing whole-genome sequencing data from over 180,000 individuals.
    \item \textbf{UK10K}~\cite{uk10k2015uk10k}: A British genomics project providing population-specific variant frequencies for the UK population through whole-genome and exome sequencing of approximately 10,000 individuals.
    \item \textbf{ESP} (NHLBI Exome Sequencing Project)~\cite{tennessen2012evolution}: A project focused on exome sequencing of individuals with heart, lung, and blood disorders, providing population frequency data stratified by ancestry.
\end{itemize}

\subsection{Clinical Evaluation Dataset Curation}

In this section, we will introduce the curation procedure of the two proposed evaluation datasets from the clinical centers, \emph{i.e.}, \textbf{MIMIC-IV-Rare} and \textbf{Xinhua Hosp.} datasets.

The MIMIC-IV-Note dataset comprised 331,794 de-identified discharge summaries from 145,915 patients sourced from public repositories, while the Xinhua Hospital dataset contained 352,425 outpatient and emergency records from 42,248 patients specializing in genetic diseases, which is an in-house clinical data.

\change{As shown in Extended Data Figure~\ref{fig:cohort}}, a systematic data preprocessing pipeline was implemented to ensure data quality and relevance. For the MIMIC-IV-Note dataset, we applied a two-stage exclusion process: first, cases without rare disease diagnoses were filtered out (n = 318,976 excluded), where rare disease classification was determined using an LLM under \change{prompt~\ref{prompt12:disease_judge}}. Subsequently, records with incomplete patient information were removed (n = 3,633 excluded), with information completeness defined as the ability to correctly extract HPO entities that could be successfully matched to the HPO database. This filtering process resulted in 9,185 cases. Similarly, the Xinhua Hospital dataset underwent parallel filtering using identical criteria, excluding 28,150 cases without rare disease diagnoses and 8,278 cases with incomplete information, yielding 5,820 cases.

Subsequently, a time-based allocation strategy was employed to partition the data into evaluation and reference sets. Recent cases were designated for testing purposes, while historical cases were allocated to similar case libraries for retrieval-based analysis. This allocation resulted in the MIMIC-IV Test Set (n = 1,875 cases) and MIMIC-IV Similar Case Library (n = 7,310 cases), alongside the Xinhua Test Set (n = 975 cases) and Xinhua Similar Case Library (n = 4,845 cases).

It should be noted that these datasets, derived from authentic clinical records, inherently contain heterogeneous and potentially noisy phenotypic information, including patient-reported symptoms, post-operative complications, multiple consultation entries, and incomplete documentation. This real-world complexity significantly increases the diagnostic challenge compared to curated datasets, thereby providing a more rigorous evaluation framework for clinical decision support systems in rare disease diagnosis.
    
\label{sec:cohort}

\subsection{Evaluation Datasets Statistics}

As shown by Extended Data Table~\ref{tab:1} from a statistical perspective, 
the number of rare diseases represented across various datasets ranges from 17 to 2150, while the average number of HPO items per patient varies between 4.0 and 19.4. Moreover, following~\cite{chen2024rarebench,fan2023improving}, 
we calculate the average information content (IC) for each dataset.
Information content quantifies the specificity of a concept within an ontology by measuring its inverse frequency of occurrence, {\em i.e.}, concepts that appear less frequently in the corpus have higher IC values. Lower IC values typically correspond to more general terms within the ontology hierarchy~\cite{fan2023improving}. This collection is the most comprehensive benchmark for rare disease diagnosis, covering \textbf{2919 diseases} from different case sources, and multiple independent clinical centers.

\subsection{Baselines}
In this section, we introduce the compared baselines in detail, covering specialized diagnostic methods, latest LLMs, and other agentic systems.

\textbf{Specialized diagnostic methods:}
\begin{itemize}
\setlength\itemsep{0.15cm}
    \item \textbf{PhenoBrain\footnote{http://www.phenobrain.cs.tsinghua.edu.cn}~\cite{mao2025phenotype}:} Takes free-text or structured HPO items as input and suggests top potential rare diseases by an ensembling method integrating the result of a graph-based Bayesian method (PPO) and two machine learning methods (CNB and MLP) via its API.
    \item \textbf{PubcaseFinder\footnote{https://pubcasefinder.dbcls.jp/}~\cite{Pubcase2}:} A website that can extract free-text input first and analyze HPO items by matching similar cases from PubMed reports, returning top potential rare disease suggestions with confidence scores, accessible via its API.
\end{itemize}
\textbf{Latest LLMs:}
\begin{itemize}
\setlength\itemsep{0.15cm}
\item \textbf{GPT-4o~\cite{jpn-med-exam_gpt4}}: A closed-source model (version identifier: gpt-4o-2024-11-20) developed by OpenAI. The model was released in May 2024.

\item \textbf{DeepSeek-V3~\cite{liu2024deepseek}}: An open-source model (version identifier: deepseek-ai/DeepSeek-V3) with 671 billion parameters. It was trained on 14.8 trillion tokens and released in December 2024. 

\item \textbf{Gemini-2.0-flash~\cite{team2023gemini}}: A closed-source model (version identifier: gemini-2.0-flash) developed by Google. This model was released in December 2024.

\item \textbf{Claude-3.7-Sonnet~\cite{claude}}: A closed-source model (version identifier: claude-3-7-sonnet) developed by Anthropic. It features a unique ``hybrid reasoning'' mechanism that allows it to switch between fast responses and extended thinking for complex tasks. This model was released in February 2025.

\item \textbf{OpenAI-o3-mini~\cite{openai_o3_mini}}: A closed-source model (version identifier: o3-mini-2025-01-31). This model was officially released in January 2025.

\item \textbf{DeepSeek-R1~\cite{guo2025deepseek}}: An open-source large language model (version identifier: deepseek-ai/DeepSeek-R1) with 671 billion parameters. The model was publicly released in January 2025.

\item \textbf{Gemini-2.0-FT~\cite{team2024gemini}}: A closed-source model (version identifier: gemini-2.0-flash-thinking-exp-01-21). Its training data encompasses information up to June 2024, and it was released in January 2025.

\item \textbf{Claude-3.7-Sonnet-thinking~\cite{claude}}: This is an extended version of Claude-3.7-sonnet that provides transparency into its step-by-step thought process. It is a closed-source reasoning model (version identifier: claude-3-7-sonnet-20250219-thinking), publicly released in January 2025.

\item \textbf{Baichuan-M1~\cite{wang2025baichuan}}: An open-source domain-specific model (version identifier: baichuan-inc/Baichuan-M1-14B-Instruct) designed specifically for medical applications, distinguishing it from the general-purpose LLMs above. This model comprises 14 billion parameters and was released in January 2025.

\item \textbf{MMedS-Llama 3~\cite{wu2025towards}}: An open-source domain-specific model (version identifier: Henrychur/MMedS-Llama-3-8B) specialized for the medical domain and released in January 2025. Built upon the Llama-3 architecture with extensive medical domain adaptation, this model comprises 8 billion parameters.
\end{itemize}

\textbf{Other Agentic Systems:}
\begin{itemize}
    \item \textbf{MDAgents~\cite{kim2024mdagents}:} A multi-agent architecture that adaptively orchestrates single or collaborative LLM configurations for medical decision-making through a five-phase methodology: Complexity Checking, Expert Recruitment, Initial Assessment, Collaborative Discussion, and Review and Final Decision.

    \item \textbf{DeepSeek-V3-Search~\cite{liu2024deepseek}:} An LLM agent framework augmented with internet search through Volcano Engine's platform and web browser plugin.
\end{itemize}

\textbf{Genetic Analysis Tools:}
\begin{itemize}
    \item \textbf{Exomiser~\cite{exomiser}:} A variant prioritization tool (version idertifier: 14.1.0 2024-11-14) that combines genomic variant data with HPO phenotype terms to identify disease-causing variants in rare genetic diseases. It integrates population frequency, pathogenicity prediction, and phenotype-gene associations to rank candidate variants.
\end{itemize}

\subsection{Web Application}

To facilitate adoption by rare disease clinicians and patients, we developed a user-friendly web application interface for \textbf{DeepRare}\footnote{http://raredx.cn/doctor}. 
The platform enables users to input patient demographics, 
family history, and clinical presentations to obtain diagnostic predictions. The backend architecture processes and structures the model outputs, presenting results through an intuitive and interactive interface optimized for clinical workflow integration.
As presented in Extended Data Figure~\ref{fig:web}, the diagnostic workflow encompasses five sequential phases:

\vspace{-0.2cm}
\begin{itemize}
\setlength\itemsep{0.2cm}
    \item \textbf{Clinical data entry: }Users input essential patient information, including age, sex, family history, and primary clinical manifestations. The platform supports the upload of supplementary materials such as case reports, diagnostic imaging, laboratory results, or raw genomic VCF files when available.
    \item \textbf{Systematic clinical inquiry: }The system first conducts ``detailed symptom inquiry'', further investigating information that helps clarify the scope of organ involvement, family genetic history, and symptom progression to narrow the diagnostic range. Users may also choose to skip this step and proceed directly to diagnosis.
    \item \textbf{HPO phenotype mapping}: The platform automatically maps clinical inputs to standardized Human Phenotype Ontology (HPO) terms, with manual curation capabilities allowing clinicians to refine, supplement, or remove assigned phenotypic descriptors.
    \item \textbf{Diagnostic analysis and output: }At this stage, the system executes a comprehensive analysis by invoking various tools and consulting medical literature and case databases to provide diagnostic recommendations and treatment suggestions for physicians. The web frontend renders the output results for user-friendly presentation.
    \item \textbf{Clinical report downloading: }Upon completion of the diagnostic analysis, users can generate comprehensive diagnostic reports that are automatically formatted and exported as PDF or Word documents for integration into electronic health records or clinical documentation.
\end{itemize}

\change{Detailed descriptions of the web engineering implementations are provided in Supplementary~\ref{sec:web}.}

\subsection{Ethical Approval and Informed Consent}
The study protocol was reviewed and approved by the Ethics Committee of Xinhua Hospital Affiliated to Shanghai Jiao Tong University School of Medicine (Approval Nos. XHEC-D-2025-094 and XHEC-D-2025-165). The study adheres to the principles of the Declaration of Helsinki. Written informed consent was obtained from all probands or their legal guardians (for those under 8 years old) prior to the initiation of genetic testing at Xinhua Hospital and Hunan Children's Hospital. The consent forms, approved by the respective Institutional Review Boards, explicitly authorized the clinical testing as well as the subsequent use of de-identified biological samples, clinical phenotypes, and genomic data for scientific research and academic publication. All data were fully de-identified and anonymized before being accessed for this study to ensure patient privacy.

\section{Data Availability}
In this study, we utilized six data sources: \emph{i.e.}, RareBench, MyGene2, the Deciphering Developmental Disorders (DDD) study, MIMIC-IV, Xinhua Hosp., and Hunan Hosp.. 

The first four datasets are publicly available and can be accessed as follows: RareBench (MME, HMS, LIRICAL, and RAMEDIS subsets) at~\url{https://huggingface.co/datasets/chenxz/RareBench}; MyGene2 at~\url{https://dataverse.harvard.edu/dataset.xhtml?persistentId=doi:10.7910/DVN/TZTPFL}; DDD at~\url{https://www.deciphergenomics.org/ddd/ddgenes}; and MIMIC-IV at~\url{https://physionet.org/content/mimiciv/3.1/}. 

The remaining two datasets, namely Xinhua Hosp. and Hunan Hosp., have been 
deposited in public repositories at the National Genomics Data Center (NGDC), 
China National Center for Bioinformation (CNCB). The overall project is archived 
under the accession number PRJCA052720 at 
\url{https://ngdc.cncb.ac.cn/bioproject/browse/PRJCA052720}. 
Variant data have been submitted to the Genome Variation Map (GVM) under the 
accession numbers GVM001237 (\url{https://ngdc.cncb.ac.cn/gvm/getVariantById?Project=GVM001237}) and GVM001238 (\url{https://ngdc.cncb.ac.cn/gvm/getVariantById?Project=GVM001238} ), and metadata/clinical phenotype 
information to the Open Multi-Omics Information eXchange (OMIX) under the 
accession OMIX013512~(\url{https://ngdc.cncb.ac.cn/omix/release/OMIX013512}).
To protect participant confidentiality, the genetic data and clinical records are available to the scientific community for research through a controlled access process. Access can be requested by submitting an application that includes a detailed research proposal and an IRB approval from the applicant's home institute to the Data Access Committee of Xinhua Hospital, Shanghai Jiao Tong University School of Medicine via the NGDC portal.

\section{Code Availability}
Our source code is available at \url{https://github.com/MAGIC-AI4Med/DeepRare}. Additionally, the extended web application developed for this research is hosted at \url{https://deeprare.cn}. 



\clearpage

\bibliographystyle{unsrt}
\bibliography{references}

\clearpage

\section{Figure Legends}

\clearpage
\section{Extended Data}
\clearpage
\setcounter{figure}{0} 
\setcounter{table}{0} 
\renewcommand{\figurename}{Extended Data Figure} 
\renewcommand{\tablename}{Extended Data Table} 

\begin{figure}[!t]
    \centering
    \includegraphics[width=0.98\linewidth]{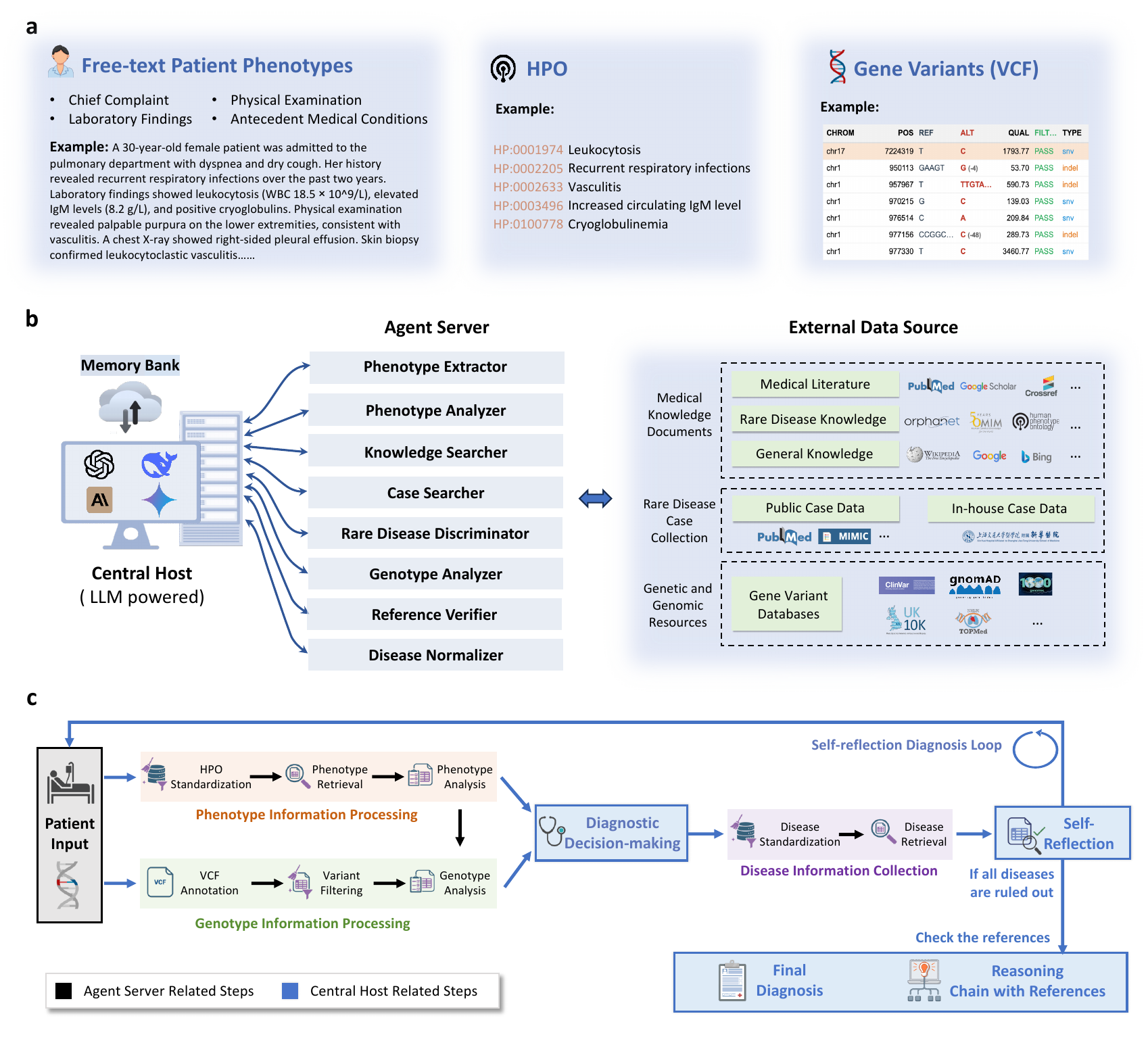}
    \vspace{0.6cm}
    \caption{\textbf{Overview of the DeepRare system.} \textbf{a} The input consists of patient free-text information, structured HPO IDs, or any combination of them. 
    \textbf{b} The three-level components in RaraDx. Inspired by the MCP, our system can also be analogized to a personal computer system architecture, comprising: (1) a central host with a memory bank for centrally managing and coordinating the system, analogous to the main computer processing system; (2) multiple agent servers to organize tools, execute specific tasks, and interact with the external environment, analogous to auxiliary hardware assistant equipment; (3) comprehensive external data sources, representing a complete external rare-disease diagnostic environment, supporting the entire system by various medical reliable evidence, including medical knowledge and clinical cases. \textbf{c} The flowchart of the main workflow of our system illustrates two primary stages, i.e., the information collection stage and the self-reflection diagnosis stage. In the former, the central host actively collects medical support information relevant to the patient. In the latter, the central host performs self-reflection on its diagnostic results. Steps involving the central host are highlighted in blue boxes within the flowchart. Created in BioRender. \url{https://BioRender.com/w2vqp03}}
    \label{fig:Architecture}
\end{figure}

\begin{figure}[t]
    \centering
    \includegraphics[width=1\linewidth]{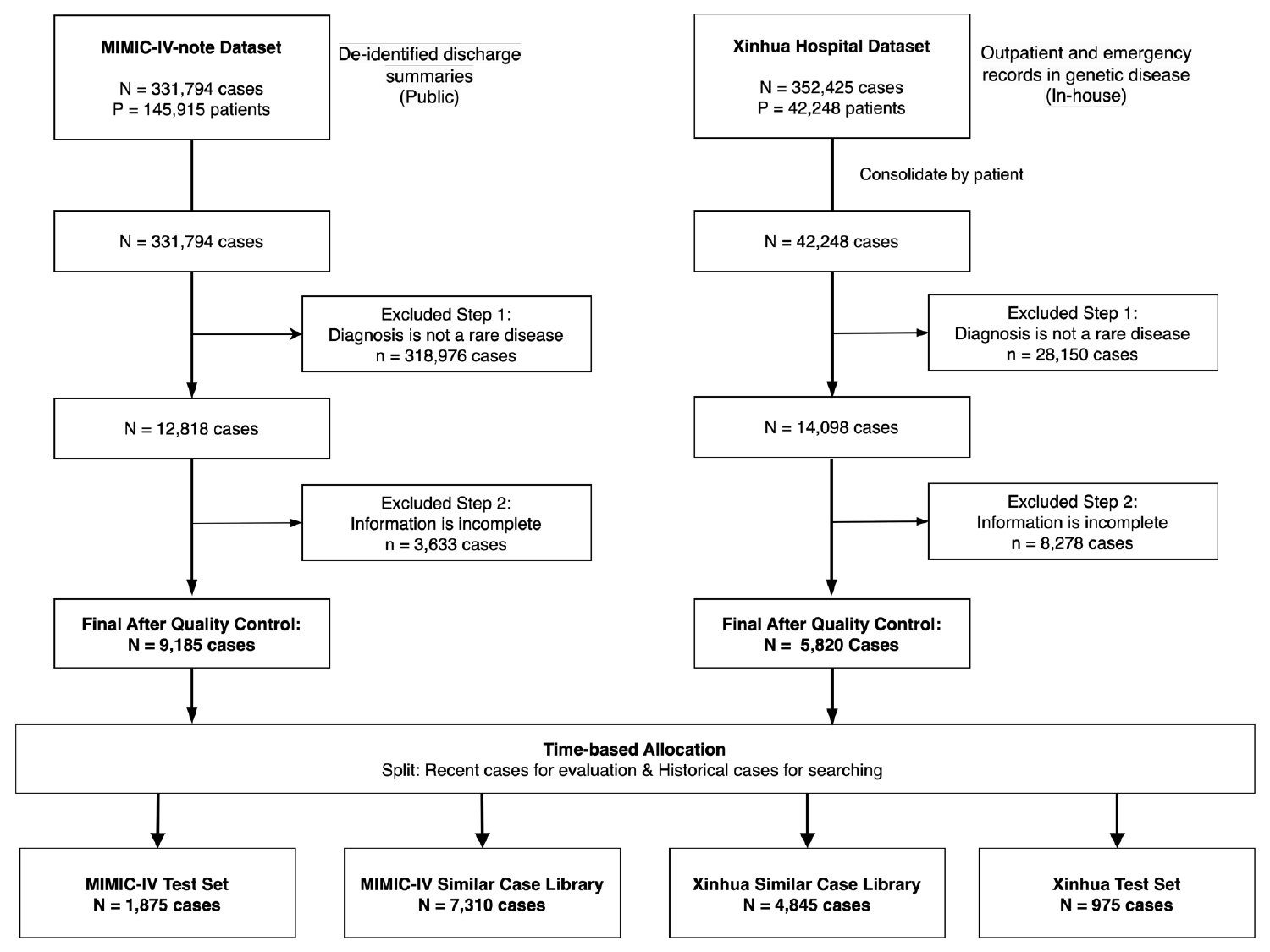}
    \caption{Cohort curation pipeline and allocation strategy for MIMIC-IV-Note and Xinhua Hospital datasets. Left: MIMIC-IV-note dataset including 331,794 cases, with 9,185 remaining after exclusions and divided into test (n = 1,875) and library (n = 7,310) sets. Right: Xinhua Hospital dataset including 352,425 cases, with 5,820 remaining after exclusions and divided into test (n = 975) and library (n = 4,845) sets. Both datasets underwent rare disease checks and information completeness filtering.}
    \label{fig:cohort}
\end{figure}

\begin{figure}[!ht]
    \centering
    \includegraphics[width=\linewidth]{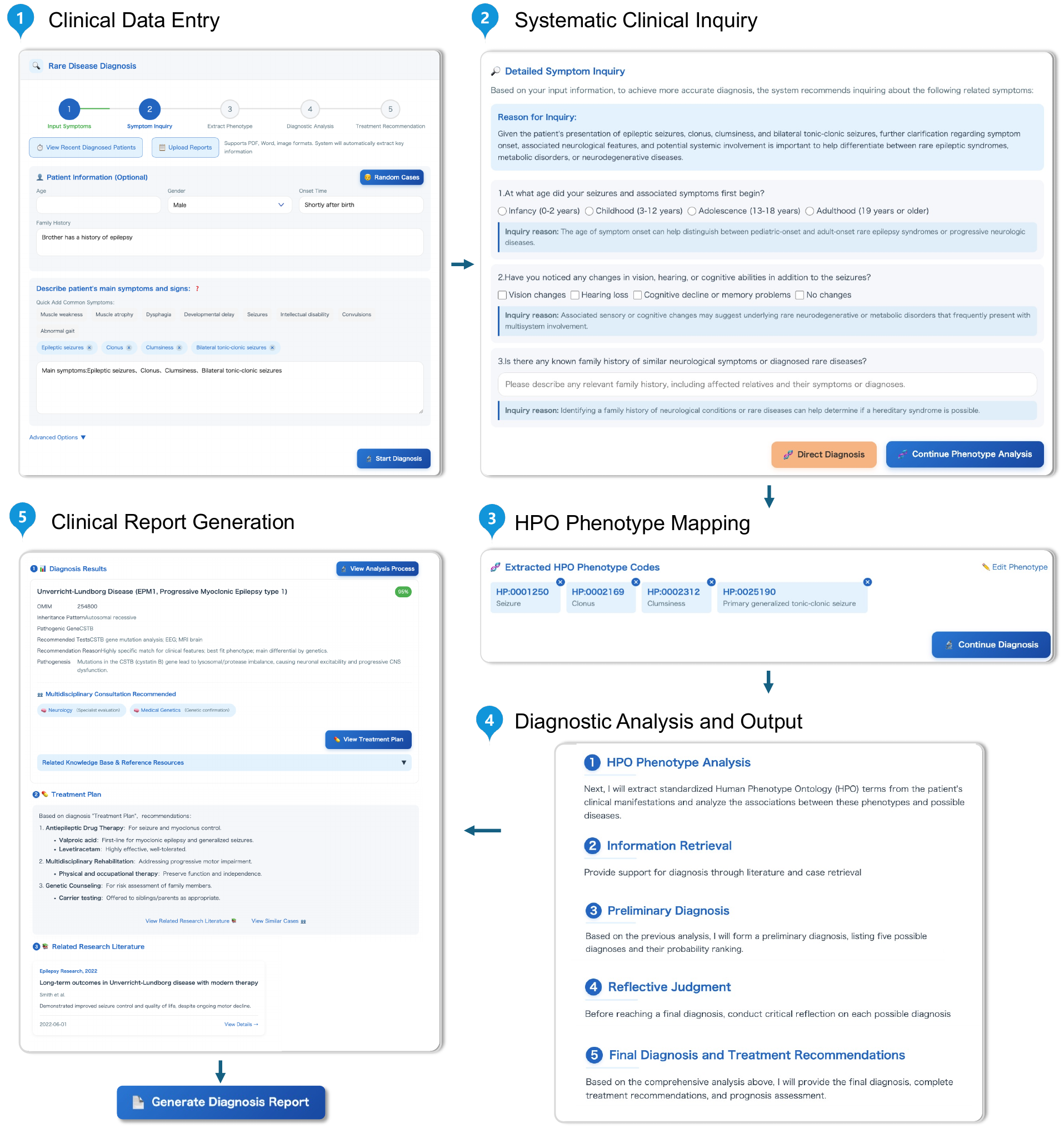}
    \vspace{0.1cm}
    \caption{\textbf{The five-stage DeepRare web application workflow.} First, clinical data entry: input of demographic parameters, family history, and clinical manifestations with optional file uploads (medical images, lab reports, VCF files). Second, systematic clinical inquiry: AI-guided symptom refinement for organ involvement and disease progression. Third, HPO phenotype mapping: automated terminology standardization with a clinician-curated adjustment interface. Fourth, diagnostic analysis: integrated tool orchestration generating evidence-based recommendations. Finally, report downloading: automated export of structured clinical reports (PDF/Word).}
    \label{fig:web}
\end{figure}

\begin{table}[t]
\centering
\caption{\textbf{Multi-center benchmark characteristics}: Case distributions, phenotypic complexity (HPO metrics), disease spectrum, provenance, and genetic annotation status (solid: confirmed pathogenic variants; half-solid: candidate variants extracted; hollow: no genetic data)}
\label{tab:1}
\scriptsize
\setlength{\tabcolsep}{2pt}
\begin{tabular}{lccccccccc}
\toprule
& \textbf{RareBench} & \textbf{RareBench} & \textbf{RareBench} & \textbf{RareBench} & \textbf{MyGene2} & \textbf{DDD} & \textbf{MIMIC-IV-} & \textbf{Xinhua} & \textbf{Hunan} \\
& \textbf{(MME)} & \textbf{(HMS)} & \textbf{(LIRICAL)} & \textbf{(RAMEDIS)} & & & \textbf{Rare} & \textbf{Hosp.} & \textbf{Hosp.} \\
\midrule
\textbf{Cases} & 40 & 88 & 370 & 624 & 146 & 2,283 & 1,875 & 975 & 162 \\
\midrule
\textbf{Avg Info Content} & 52.7 & 103.7 & 59.5 & 46.1 & 29.4 & 70.9 & 50.1 & 16.4 & 27.5 \\
\midrule
\textbf{Avg HPO Ids} & 12.2 & 19.4 & 14.3 & 10.1 & 7.9 & 18.0 & 10.1 & 4.0 & 7.0 \\
\midrule
\textbf{Rare Diseases} & 17 & 39 & 252 & 74 & 58 & 2150 & 355 & 314 & 106 \\
\midrule
\textbf{Source} & Literature & Clinical & Literature & Scientist & Patients & Literature & Clinical & Clinical & Clinical \\
& & Center & & Uploaded & Uploaded & & Center & Center & Center \\
& & (Germany) & & & & & (USA) & (China) & (China) \\
\midrule
\textbf{Raw Genome Data} & $\circ$ & $\circ$ & $\circ$ & $\circ$ & $\odot$ & $\odot$ & $\circ$ & $\bullet$ & $\bullet$ \\
\bottomrule
\end{tabular}
\end{table}

\clearpage

\section{Supplementary}
\subsection{Main Workflow of DeepRare System}

\begin{algorithm}[h!]
\caption{Main Workflow of DeepRare System}
\begin{algorithmic}[1]
\State \textbf{Input:} $\mathcal{I} = \{\mathcal{T}, \mathcal{H}, \mathcal{G}\}$ \Comment{Patient input: Free-texts, HPO Items, Genetic Variants}
\State \textbf{Output:} $\mathcal{D}, \mathcal{R}$ \Comment{Final diagnosis list with rationale explanations}
\State
\State Initialize memory bank $\mathcal{M} \gets \emptyset$
\State $\mathcal{D} \gets \emptyset$ \Comment{Initialize diagnosis list as empty}
\State $N \gets N_0$ \Comment{Initialize search depth}
\State
\While{$\mathcal{D} = \emptyset$} \Comment{Iterate until valid diagnosis is obtained}
    \State
    \State \textbf{// Information Collection Stage}
    \State
    \State \textbf{Phenotype Information Collection:}
    \If{$\mathcal{T} \neq \emptyset$}
        \State $\hat{\mathcal{P}} \gets a_{\text{hpo}}(\mathcal{T}, \mathcal{H})$ \Comment{HPO standardization}
    \Else
        \State $\hat{\mathcal{P}} \gets \mathcal{H}$
    \EndIf
    \State $\mathcal{E}_{\text{hpo}} \gets a_\text{k-search}(\hat{\mathcal{P}}, \mathcal{M}, N) \cup a_\text{c-search}(\hat{\mathcal{P}}, \mathcal{M}, N)$ \Comment{Phenotype retrieval with depth $N$}
    \State $\mathcal{Y}_{\text{hpo}} \gets a_\text{hpo-analyzer}(\hat{\mathcal{P}})$ \Comment{Phenotype analysis}
    \State $\mathcal{M} \gets \mathcal{M} \cup (\mathcal{E}_{\text{hpo}}, \mathcal{Y}_{\text{hpo}})$ \Comment{Update memory bank}
    \State
    \State $\mathcal{D}^{\prime} \gets \mathcal{A}_\text{host}(\hat{\mathcal{P}}, \mathcal{M} | \texttt{<prompt>}_{\ref{prompt4:first_diag}})$ \Comment{Tentative diagnosis}
    \State
    \State \textbf{Genotype Information Collection:}
    \If{$\mathcal{G} \neq \emptyset$}
        \State $\hat{\mathcal{G}} \gets a_{\text{geno-analyzer}}(\mathcal{G})$ \Comment{VCF annotation \& variant ranking}
        \State $\mathcal{D}^{\prime\prime} \gets \mathcal{A}_\text{host}(\hat{\mathcal{G}}, \mathcal{M},\mathcal{D}^{\prime}, N | \texttt{<prompt>}_{\ref{prompt3.5:gene}})$ \Comment{Synthetic analysis with depth $N$}
        \State $\mathcal{M} \gets \mathcal{M} \cup \mathcal{D}^{\prime\prime}$ \Comment{Update memory bank}
    \Else
        \State $\mathcal{D}^{\prime\prime} \gets \mathcal{D}^{\prime}$ \Comment{Use phenotype-only diagnosis when no genetic data}
    \EndIf
    \State
    \State \textbf{// Self-reflective Diagnosis Stage}
    \State $\mathcal{E}_{\text{disease}} \gets a_{\text{k-search}}(a_\text{d-norm}(\mathcal{D}^{\prime\prime}), \mathcal{M})$ \Comment{Disease retrieval}
    \State $\mathcal{M} \gets \mathcal{M} \cup \mathcal{E}_{\text{disease}}$ \Comment{Update memory bank}
    \State $\mathcal{D} \gets \mathcal{A}_\text{host}(\mathcal{D}^{\prime\prime}, \hat{\mathcal{I}}, \mathcal{M} | \texttt{<prompt>}_{\ref{prompt5:reflection}})$ \Comment{Self-reflection}
    \State
    \If{$\mathcal{D} = \emptyset$} \Comment{If self-reflection rules out all diseases}
        \State $N \gets N + \Delta N$ \Comment{Increase search depth}
    \EndIf
    \State
\EndWhile
\State
\State $\{\mathcal{D}, \mathcal{R}\} \gets \mathcal{A}_\text{host}(\mathcal{D}, \hat{\mathcal{I}}, \mathcal{M} | \texttt{<prompt>}_{\ref{prompt6:final}})$ \Comment{Final diagnosis with rationale}
\State
\State \textbf{return} $\mathcal{D}, \mathcal{R}$
\end{algorithmic}
\label{al:1}
\end{algorithm}

\newpage
\subsection{Rare Disease Diagnosis Result (Table version)}

\begin{table}[!h]
\caption{Rare Disease Diagnosis Result (HPO Input only)}
\scalebox{0.6}[0.66]{
\fontsize{9pt}{15pt}\selectfont  
\setlength{\tabcolsep}{0.7pt}  

\begin{tabular}{l*{27}{c}}

\toprule
\multirow{3}{*}{\makecell[c]{\textbf{Model}}} & \multicolumn{3}{c}{\makecell{\textbf{RareBench}\\\textbf{(MME)}}} & \multicolumn{3}{c}{\makecell{\textbf{RareBench}\\\textbf{(HMS)}}} & \multicolumn{3}{c}{\makecell{\textbf{RareBench}\\\textbf{(LIRICAL)}}} & \multicolumn{3}{c}{\makecell{\textbf{RareBench}\\\textbf{(RAMEDIS)}}} & \multicolumn{3}{c}{\makecell{\textbf{MyGene}}} & \multicolumn{3}{c}{\makecell{\textbf{DDD}}} & \multicolumn{3}{c}{\makecell{\textbf{MIMIC-IV}\\\textbf{-Rare}}} & \multicolumn{3}{c}{\makecell{\textbf{Avg.}}} & \multicolumn{3}{c}{\makecell{\textbf{Xinhua Hosp.}}} \\[2pt]
\cmidrule(lr){2-4} \cmidrule(lr){5-7} \cmidrule(lr){8-10} \cmidrule(lr){11-13} \cmidrule(lr){14-16} \cmidrule(lr){17-19} \cmidrule(lr){20-22} \cmidrule(lr){23-25} \cmidrule(lr){26-28}

& \multicolumn{3}{c}{case=40} & \multicolumn{3}{c}{case=88} & \multicolumn{3}{c}{case=370} & \multicolumn{3}{c}{case=624} & \multicolumn{3}{c}{case=146} & \multicolumn{3}{c}{case=2283} & \multicolumn{3}{c}{case=1875} & \multicolumn{3}{c}{-} & \multicolumn{3}{c}{case=975} \\[2pt]
\cmidrule(lr){2-4} \cmidrule(lr){5-7} \cmidrule(lr){8-10} \cmidrule(lr){11-13} \cmidrule(lr){14-16} \cmidrule(lr){17-19} \cmidrule(lr){20-22} \cmidrule(lr){23-25} \cmidrule(lr){26-28}

& {R@1} & {R@3} & {R@5} & {R@1} & {R@3} & {R@5} & {R@1} & {R@3} & {R@5} & {R@1} & {R@3} & {R@5} & {R@1} & {R@3} & {R@5} & {R@1} & {R@3} & {R@5} & {R@1} & {R@3} & {R@5} & {R@1} & {R@3} & {R@5} & {R@1} & {R@3} & {R@5} \\

\midrule
\rowcolor{gray!10}\multicolumn{28}{c}{\textbf{Diagnosis API}} \\
\midrule
\textbf{PhenoBrain} & 25.00 & 45.00 & 80.00 & 22.35 & 32.94 & 42.35 & 27.10 & 44.72 & 57.72 & 20.67 & 38.30 & 55.77 & 45.21 & 58.90 & 61.64 & 38.92 & 54.68 & 63.11 & 8.11 & 10.24 & 11.41 & 26.78 & 40.68 & 53.14 & {-} & {-} & {-} \\[1pt]
\textbf{PubCaseFinder} & 40.00 & 57.50 & 57.50 & 7.32 & 13.41 & 15.85 & 37.77 & 47.01 & 48.91 & 25.84 & 32.58 & 39.97 & 20.55 & 26.71 & 28.08 & 26.92 & 32.32 & 34.05 & 1.06 & 1.39 & 1.60 & 22.78 & 30.13 & 32.28 & {-} & {-} & {-} \\[1pt]

\midrule
\rowcolor{gray!10}\multicolumn{28}{c}{\textbf{LLM}} \\
\midrule
\textbf{GPT-4o} & 2.50 & 5.00 & 10.00 & 47.73 & 60.23 & 65.91 & 31.08 & 42.16 & 48.92 & 35.47 & 52.24 & 61.06 & 19.31 & 35.17 & 37.93 & 33.30 & 42.28 & 45.18 & 13.76 & 22.35 & 25.60 & 26.16 & 37.06 & 42.08 & {-} & {-} & {-} \\[1pt]
\textbf{DeepSeek V3} & 5.00 & 20.00 & 27.50 & 42.05 & 56.82 & 62.50 & 33.24 & 42.97 & 45.68 & 40.06 & 66.51 & 72.11 & 20.83 & 39.58 & 47.22 & 34.01 & 45.27 & 48.83 & 14.61 & 23.04 & 26.18 & 27.11 & 42.03 & 47.15 & 42.43 & 54.91 & 59.41 \\[1pt]
\textbf{Gemini 2} & 5.00 & 7.50 & 10.00 & 31.82 & 47.73 & 54.55 & 30.81 & 39.73 & 43.51 & 40.37 & 52.56 & 60.42 & 28.77 & 48.63 & 54.79 & 31.32 & 43.60 & 53.03 & 14.29 & 20.16 & 24.11 & 26.05 & 37.13 & 42.92 & {-} & {-} & {-} \\[1pt]
\textbf{Claude-3.7-Sonnet} & 7.50 & 27.50 & 32.50 & 43.18 & 60.23 & 63.64 & 33.24 & 46.49 & 50.00 & 42.47 & 68.59 & 73.88 & 19.44 & 36.11 & 39.58 & 32.28 & 42.76 & 45.93 & 12.11 & 18.24 & 20.80 & 27.17 & 42.84 & 46.62 & {-} & {-} & {-} \\[1pt]

\midrule[0.1em]
\rowcolor{gray!10}\multicolumn{28}{c}{\textbf{Reasoning LLM}} \\
\midrule
\textbf{o3mini} & 2.50 & 7.50 & 7.50 & 30.68 & 44.32 & 54.55 & 25.68 & 35.41 & 40.00 & 39.42 & 64.42 & 67.47 & 34.48 & 46.21 & 49.66 & 34.09 & 42.96 & 45.50 & 10.88 & 19.52 & 23.90 & 25.39 & 37.19 & 41.23 & {-} & {-} & {-} \\[1pt]
\textbf{DeepSeek R1} & 35.00 & 60.00 & 65.00 & 38.64 & 53.41 & 60.23 & 36.76 & 45.14 & 47.57 & 33.17 & 47.60 & 53.21 & 35.62 & 46.58 & 47.95 & 38.24 & 49.43 & 53.01 & 13.81 & 19.92 & 22.93 & 33.03 & 46.01 & 49.99 & 39.16 & 49.85 & 54.98 \\[1pt]
\textbf{Gemini 2 think} & 0.00 & 5.00 & 12.50 & 38.64 & 48.86 & 56.82 & 30.81 & 41.35 & 44.05 & 33.97 & 50.80 & 58.65 & 39.72 & 53.42 & 53.42 & 34.43 & 46.33 & 50.02 & 15.47 & 20.53 & 25.39 & 27.58 & 38.05 & 42.98 & {-} & {-} & {-} \\[1pt]
\textbf{Claude 3.7 think} & 27.50 & 50.00 & 57.50 & 44.32 & 57.95 & 62.50 & 38.11 & 50.27 & 52.43 & 41.83 & 63.78 & 72.92 & 34.25 & 37.67 & 39.72 & 34.01 & 45.27 & 48.83 & 13.70 & 21.28 & 24.32 & 33.39 & 46.60 & 51.17 & {-} & {-} & {-} \\[1pt]

\midrule[0.1em]
\rowcolor{gray!10}\multicolumn{28}{c}{\textbf{Medical LLM}} \\
\midrule
\textbf{MMedS-Llama 3} & 12.50 & 20.00 & 20.00 & 17.04 & 29.55 & 31.82 & 21.08 & 29.73 & 33.51 & 30.45 & 45.67 & 50.00 & 15.07 & 23.97 & 28.09 & 22.18 & 33.80 & 38.34 & 14.04 & 22.58 & 26.37 & 18.91 & 29.33 & 32.59 & 29.55 & 40.18 & 45.30 \\[1pt]
\textbf{Baichuan-14B} & 5.00 & 10.00 & 10.00 & 38.64 & 53.41 & 60.23 & 26.77 & 37.03 & 40.27 & 37.66 & 64.26 & 68.75 & 6.16 & 23.91 & 30.82 & 29.32 & 39.37 & 44.29 & 14.40 & 21.33 & 24.59 & 22.56 & 35.62 & 39.85 & 43.35 & 54.40 & 58.89 \\[1pt]

\midrule[0.1em]
\rowcolor{gray!10}\multicolumn{28}{c}{\textbf{Other Agentic System}} \\
\midrule
\textbf{MDAgents~(GPT-4o)} & 2.25 & 2.25 & 5.00 & 27.27 & 40.91 & 46.59 & 21.89 & 28.92 & 31.08 & 37.82 & 58.01 & 62.98 & 15.75 & 22.60 & 24.66 & 25.01 & 33.33 & 36.01 & 10.98 & 17.12 & 20.27 & 20.14 & 29.02 & 32.37 & {-} & {-} & {-} \\[1pt]
\textbf{MDAgents~(DS-V3)} & 0.00 & 0.00 & 0.025 & 29.55 & 43.18 & 45.45 & 23.24 & 28.11 & 31.08 & 36.22 & 59.29 & 63.30  & 8.21 & 11.64 & 15.75 & 25.76 & 32.33 & 34.69 & 8.69 & 15.04 & 19.41 & 18.81 & 27.08 & 29.96 & {-} & {-} & {-} \\[1pt]
\textbf{DS-R1-search} & 12.50 & 17.50 & 22.50 & 38.64 & 55.68 & 62.50 & 33.51 & 45.41 & 49.46 & 39.90 & 66.51 & 71.31 & 15.75 & 34.93 & 46.57 & 32.30 & 47.15 & 52.41 & 19.95 & 28.75 & 32.69 & 27.51 & 42.28 & 48.21 & {-} & {-} & {-} \\[1pt]
\textbf{RAG~(DS-V3)} & 42.50 & 45.00 & 47.50 & 36.36 & 46.59 & 46.59 & 29.46 & 36.76 & 39.46 & 28.57 & 46.10 & 50.16 & 30.14 & 42.47 & 48.63 & 39.42 & 48.84 & 51.77 & 12.75 & 18.61 & 20.80 & 31.31 & 40.62 & 43.56 & {-} & {-} & {-} \\[1pt]

\midrule[0.1em]
\multicolumn{28}{c}{\cellcolor{gray!20}\textbf{DeepRare System}} \\
\midrule
\rowcolor{gray!10}\textbf{DeepRare~(GPT-4o)} & \cellcolor{cyan!8}72.50 & \cellcolor{cyan!8}77.50 & \cellcolor{cyan!8}82.50 & \cellcolor{cyan!8}54.54 & \cellcolor{cyan!15}\textbf{65.90} & \cellcolor{cyan!8}69.31 & \cellcolor{cyan!8}51.62 & \cellcolor{cyan!8}62.70 & \cellcolor{cyan!8}65.40 & \cellcolor{cyan!15}\textbf{73.23} & \cellcolor{cyan!8}81.57 & \cellcolor{cyan!15}\textbf{85.26} & \cellcolor{cyan!15}\textbf{75.86} & \cellcolor{cyan!8}80.00 & \cellcolor{cyan!8}80.69 & \cellcolor{cyan!8}39.85 & \cellcolor{cyan!8}51.36 & \cellcolor{cyan!8}55.88 & \cellcolor{cyan!8}21.44 & \cellcolor{cyan!8}28.89 & \cellcolor{cyan!8}31.71 & \cellcolor{cyan!8}55.58 & \cellcolor{cyan!8}63.99 & \cellcolor{cyan!8}67.25 & {-} & {-} & {-} \\[1pt]
\rowcolor{gray!10}\textbf{DeepRare~(DS-V3)} & \cellcolor{cyan!8}70.00 & \cellcolor{cyan!8}77.50 & \cellcolor{cyan!8}77.50 & \cellcolor{cyan!15}\textbf{56.97} & \cellcolor{cyan!8}65.12 & \cellcolor{cyan!15}\textbf{70.93} & \cellcolor{cyan!15}\textbf{55.95} & \cellcolor{cyan!15}\textbf{65.16} & \cellcolor{cyan!8}67.57 & \cellcolor{cyan!8}72.60 & \cellcolor{cyan!15}\textbf{82.85} & \cellcolor{cyan!8}84.62 & \cellcolor{cyan!8}72.60 & \cellcolor{cyan!8}76.03 & \cellcolor{cyan!8}76.03 & \cellcolor{cyan!8}42.97 & \cellcolor{cyan!8}53.48 & \cellcolor{cyan!8}56.72 & \cellcolor{cyan!15}\textbf{29.19} & \cellcolor{cyan!15}\textbf{36.60} & \cellcolor{cyan!15}\textbf{39.06} & \cellcolor{cyan!15}\textbf{57.18} & \cellcolor{cyan!15}\textbf{65.25} & \cellcolor{cyan!15}\textbf{67.49} & \cellcolor{cyan!15}58.27 & \cellcolor{cyan!15}71.04 & \cellcolor{cyan!15}74.35 \\[1pt]
\rowcolor{gray!10}\textbf{DeepRare~(DS-R1)} & \cellcolor{cyan!8}67.57 & \cellcolor{cyan!8}81.08 & \cellcolor{cyan!8}83.78 & \cellcolor{cyan!8}51.76 & \cellcolor{cyan!8}60.00 & \cellcolor{cyan!8}67.06 & \cellcolor{cyan!8}52.43 & \cellcolor{cyan!8}63.51 & \cellcolor{cyan!8}67.84 & \cellcolor{cyan!8}62.17 & \cellcolor{cyan!8}76.28 & \cellcolor{cyan!8}79.32 & \cellcolor{cyan!8}75.00 & \cellcolor{cyan!8}80.55 & \cellcolor{cyan!15}\textbf{81.94} & \cellcolor{cyan!8}44.27 & \cellcolor{cyan!8}57.09 & \cellcolor{cyan!8}61.27 & \cellcolor{cyan!8}23.27 & \cellcolor{cyan!8}31.16 & \cellcolor{cyan!8}34.79 & \cellcolor{cyan!8}53.78 & \cellcolor{cyan!8}64.24 & \cellcolor{cyan!8}68.00 & \cellcolor{cyan!8}54.30 & \cellcolor{cyan!8}66.90 & \cellcolor{cyan!8}70.20\\[1pt]
\rowcolor{gray!10}\textbf{DeepRare~(claude-3.7)} & \cellcolor{cyan!8}72.50 & \cellcolor{cyan!8}85.00 & \cellcolor{cyan!8}87.50 & \cellcolor{cyan!8}56.82 & \cellcolor{cyan!8}63.64 & \cellcolor{cyan!8}64.77 & \cellcolor{cyan!8}40.00 & \cellcolor{cyan!8}51.08 & \cellcolor{cyan!8}54.86 & \cellcolor{cyan!8}70.03 & \cellcolor{cyan!8}81.57 & \cellcolor{cyan!8}84.46 & \cellcolor{cyan!8}73.79 & \cellcolor{cyan!15}\textbf{80.68} & \cellcolor{cyan!8}80.69 & \cellcolor{cyan!8}45.49 & \cellcolor{cyan!8}58.85 & \cellcolor{cyan!15}\textbf{63.76} & \cellcolor{cyan!8}24.32 & \cellcolor{cyan!8}30.30 & \cellcolor{cyan!8}32.12 & \cellcolor{cyan!8}54.71 & \cellcolor{cyan!8}64.45 & \cellcolor{cyan!8}66.88 & {-} & {-} & {-} \\[1pt]
\rowcolor{gray!10}\textbf{DeepRare~(gemini-2)} & \cellcolor{cyan!15}\textbf{77.50} & \cellcolor{cyan!15}\textbf{85.00} & \cellcolor{cyan!15}\textbf{90.00} & \cellcolor{cyan!8}40.91 & \cellcolor{cyan!8}54.25 & \cellcolor{cyan!8}57.95 & \cellcolor{cyan!8}48.91 & \cellcolor{cyan!8}63.24 & \cellcolor{cyan!15}\textbf{68.91} & \cellcolor{cyan!8}49.67 & \cellcolor{cyan!8}69.55 & \cellcolor{cyan!8}75.64 & \cellcolor{cyan!8}69.18 & \cellcolor{cyan!8}77.40 & \cellcolor{cyan!8}79.45 & \cellcolor{cyan!15}\textbf{47.59} & \cellcolor{cyan!15}\textbf{59.82} & \cellcolor{cyan!8}62.70 & \cellcolor{cyan!8}19.08 & \cellcolor{cyan!8}26.28 & \cellcolor{cyan!8}29.41 & \cellcolor{cyan!8}50.41 & \cellcolor{cyan!8}62.22 & \cellcolor{cyan!8}66.29 & {-} & {-} & {-} \\
\bottomrule
\multicolumn{28}{@{}l@{}}{\footnotesize \textit{Note:} All values reported in the table are percentages (\%).}

\end{tabular}}
\label{tab:s1}
\end{table}

\subsection{Reference Link Verification}
\label{sec:website_verification}

\change{To ascertain the validity and accessibility of web references in output reasoning, we implement a lightweight, two-stage post-process verification stage. The first stage validates the URL's syntactic structure, including its scheme and hostname. The second stage then performs a reachability test by issuing an HTTP HEAD request. A reference is deemed accessible if the server responds with a success (2xx) or redirection (3xx) status code. URLs that are syntactically malformed or that result in a client/server error (4xx/5xx) or connection timeout are classified as unreachable.}

\subsection{Web Engineering Implementation}
\label{sec:web}
\change{In terms of \textbf{engineering implementation}, we packaged this workflow using FastAPI with DeepSeek-V3 locally deployed on 16 Ascend 910B cards serving as the central host to ensure system stability and data security. The system architecture employs a microservices design with Redis for session management and SQL databases for persistent data storage. Here, we implemented several key optimizations:}

\change{First, we established a comprehensive multi-level caching mechanism where our memory bank serves dual purposes for both reasoning and result caching. When external tools are invoked, their results are stored in the memory bank with structured indexing based on query types, HPO terms, and disease categories. For subsequent user inputs, the system first performs exact matching against cached results within the past month before initiating new searches.}

\change{Second, we implemented knowledge base localization with automated synchronization schedules to minimize external API dependencies. For downloadable external databases such as OrphaNet (updated monthly) and OMIM disease knowledge (updated weekly), we maintain local mirror databases with version control and incremental updates. This approach ensures both data freshness and rapid query response times while reducing reliance on external network connections.}

\change{Third, we adopted comprehensive parallelization strategies for multi-database operations and user interaction management. The system utilizes Celery with Redis broker for distributed task queuing, enabling parallel execution of simultaneous database searches across OrphaNet, OMIM, PubMed, and other medical databases. For frontend user interactions, we implement asyncio-based asynchronous processing that allows the diagnostic workflow to handle multiple user sessions concurrently without blocking, ensuring responsive user experience even during computationally intensive analysis phases.}

\change{These optimizations collectively address the computational efficiency and scalability requirements for real-world clinical deployment. The system has currently been deployed and tested in hospital settings, demonstrating improved diagnostic efficiency and clinical decision-making accuracy while maintaining robust performance under concurrent user loads.}

\subsection{Ablation Study on Case Bank}
\label{sec:ablation}
\changee{We conduct extensive ablation studies comparing: ``LLM only'', ``LLM + case search'', and ``LLM + tool-calling + reflection'' (equivalent to deactivating case search in DeepRare). These experiments are performed across all \textbf{four RareBench datasets} and the \textbf{Xinhua Hosp. Casebank-disentangled Subset}~(comprising 152 cases sampled from the Xinhua Hosp. whose diagnoses are completely absent from our case bank, so as to minimize effects due to case similarity and better demonstrate the effectiveness of our agentic design). The results demonstrate the individual effectiveness of both case search and tool-calling, as well as a synergistic effect when combined. 
We attribute this synergy to the core benefits of our multi-agent architecture.}

\begin{itemize}[itemsep=4pt]
\item \changee{As shown in Table~\ref{tab:rarebench} and Table~\ref{tab:zeroshot_performance} (noted in pink color), in \textbf{RareBench (HMS)}, \textbf{RareBench (LIRICAL)}, 
and the \textbf{Xinhua Hosp. Casebank-disentangled Subset} with sparse or no relevant cases, knowledge tools and reflective reasoning are equally crucial as case retrieval, providing improvements of +2.28, +8.38, and +4.57 in R@1 respectively. Notably, even in the \textbf{Xinhua Hosp. Casebank-disentangled Subset} containing completely unseen diseases, ``LLM+case search'' still achieves a 4.57-point performance improvement, because the retrieval process can find diseases with similar manifestations, which still provides some diagnostic assistance.}

\item \changee{As shown in Table~\ref{tab:rarebench} (noted in green color), in datasets like \textbf{RareBench (MME)} and \textbf{RareBench (RAMEDIS)} where the case bank contains substantially similar typical cases (though not identical), case retrieval provides dramatic improvements (+60.00 and +34.56 in R@1, respectively). 
However, this does not diminish the value of ``LLM+tool+reflection'', which may also delivers significant gains~(+40.00 on MME) over the base LLM.}

\end{itemize}

\changee{\textbf{Synergistic integration and architectural robustness:} 
The \textbf{full DeepRare} system, integrating all components, achieves performance exceeding any individual configuration, often by substantial margins (frequently $>10$ points on R@1). This underscores that each component is useful on its own but most powerful in concert.
This significant enhancement demonstrates the complementary nature of these components, enabled by our agentic architecture that dynamically orchestrates these modules. 
When strong precedents are available ({\em e.g.}, MME/RAMEDIS), case retrieval delivers large gains; when precedents are sparse (HMS/unseen diseases), knowledge tools and reflection maintain robust performance.}

\begin{table}[h]
\centering
\caption{Ablation results on RareBench subsets. Values are Top-k Recall (R@k) without percent signs. Use gpt-4o as the LLM and DeepRare central host. }
\scriptsize
\setlength{\tabcolsep}{3pt}

\begin{tabular}{l *{3}{c} *{3}{c} *{3}{c} *{3}{c}}
\toprule
& \blockthree{lightpink}{\shortstack{RareBench (HMS) \\ $(n=88)$}}
& \blockthree{lightpink}{\shortstack{RareBench (LIRICAL) \\ $(n=370)$}}
& \blockthree{lightgreen}{\shortstack{RareBench (MME) \\ $(n=40)$}}
& \blockthree{lightgreen}{\shortstack{RareBench (RAMEDIS) \\ $(n=624)$}} \\
\cmidrule(lr){2-4} \cmidrule(lr){5-7} \cmidrule(lr){8-10} \cmidrule(lr){11-13}
Method
& \colhead{lightpink}{R@1} & \colhead{lightpink}{R@3} & \colhead{lightpink}{R@5}
& \colhead{lightpink}{R@1} & \colhead{lightpink}{R@3} & \colhead{lightpink}{R@5}
& \colhead{lightgreen}{R@1} & \colhead{lightgreen}{R@3} & \colhead{lightgreen}{R@5}
& \colhead{lightgreen}{R@1} & \colhead{lightgreen}{R@3} & \colhead{lightgreen}{R@5} \\
\midrule
LLM
& \cellcolor{lightpink}47.72 & \cellcolor{lightpink}60.23 & \cellcolor{lightpink}65.91
& \cellcolor{lightpink}31.08 & \cellcolor{lightpink}42.16 & \cellcolor{lightpink}48.92
& \cellcolor{lightgreen}2.50  & \cellcolor{lightgreen}5.00  & \cellcolor{lightgreen}10.00
& \cellcolor{lightgreen}35.47 & \cellcolor{lightgreen}52.24 & \cellcolor{lightgreen}61.06 \\
LLM + Case search
& \cellcolor{lightpink}47.72 & \cellcolor{lightpink}59.09 & \cellcolor{lightpink}65.91
& \cellcolor{lightpink}41.35 & \cellcolor{lightpink}53.24 & \cellcolor{lightpink}55.95
& \cellcolor{lightgreen}62.50 & \cellcolor{lightgreen}70.0  & \cellcolor{lightgreen}70.0
& \cellcolor{lightgreen}70.03 & \cellcolor{lightgreen}78.69 & \cellcolor{lightgreen}81.73 \\
LLM $+$ Tool-calling $+$ Reflection
& \cellcolor{lightpink}50.00 & \cellcolor{lightpink}62.50 & \cellcolor{lightpink}64.77
& \cellcolor{lightpink}39.46 & \cellcolor{lightpink}49.19 & \cellcolor{lightpink}52.43
& \cellcolor{lightgreen}42.50 & \cellcolor{lightgreen}57.50 & \cellcolor{lightgreen}57.70
& \cellcolor{lightgreen}36.85 & \cellcolor{lightgreen}46.38 & \cellcolor{lightgreen}55.38 \\
\midrule
\textbf{DeepRare}
& \cellcolor{lightpink}\textbf{54.54} & \cellcolor{lightpink}\textbf{65.90} & \cellcolor{lightpink}\textbf{69.31}
& \cellcolor{lightpink}\textbf{51.62} & \cellcolor{lightpink}\textbf{62.70} & \cellcolor{lightpink}\textbf{65.40}
& \cellcolor{lightgreen}\textbf{72.50} & \cellcolor{lightgreen}77.50 & \cellcolor{lightgreen}\textbf{82.50}
& \cellcolor{lightgreen}\textbf{73.23} & \cellcolor{lightgreen}\textbf{81.57} & \cellcolor{lightgreen}\textbf{85.26} \\
\bottomrule
\end{tabular}

\vspace{5pt}
\footnotesize
\label{tab:rarebench}
\end{table}

\begin{table}[h]
    \centering
    \caption{Ablation results on Xinhua Hosp. Casebank-Disentangled Subset where all groudtruth diseases are not included in the casebank. Values are Top-k Recall (R@k) without percent signs. Use DeepSeek-V3 as the LLM and DeepRare central host.
    }
    \scriptsize
    \setlength{\tabcolsep}{17pt}
    \begin{tabular}{l c c c}
    \toprule
    & \multicolumn{3}{c}{\cellcolor{lightpink} Xinhua Hosp. Casebank-Disentangled Subset $(n=153)$} \\
    \cmidrule(lr){2-4}
    Method & \cellcolor{lightpink}R@1 & \cellcolor{lightpink}R@3 & \cellcolor{lightpink}R@5 \\
    \midrule
    LLM & \cellcolor{lightpink}26.14 & \cellcolor{lightpink}38.56 & \cellcolor{lightpink}43.79 \\
    LLM + Case Search & \cellcolor{lightpink}30.71 & \cellcolor{lightpink}41.83 & \cellcolor{lightpink}49.01 \\
    LLM $+$ Tool-calling $+$ Reflection & \cellcolor{lightpink}30.71 & \cellcolor{lightpink}43.13 & \cellcolor{lightpink}45.09 \\
    \midrule
    \textbf{DeepRare} & \cellcolor{lightpink}\textbf{37.25} & \cellcolor{lightpink}\textbf{49.67} & \cellcolor{lightpink}\textbf{53.59} \\
    \bottomrule
    \end{tabular}
    \label{tab:zeroshot_performance}
\end{table}

\subsection{Phenotype Extractor Comparison}
\label{sec:ablation_study}

\change{In Table~\ref{tab:hpo_extraction}, we evaluate \textbf{the HPO extraction performance with blinded expert assessment}. To further assess the reliability of phenotype structuring, we conducted a comparative evaluation of HPO codes extraction methods against blinded human adjudication. Five domain experts first performed reference manual extractions for 367 clinical phenotype records. We then applied three competitive pipelines—DeepSeek-V3, RAG+DeepSeek-V3, and our DeepSeek-V3–based summarization with embedding-model matching—to generate HPO codes. ``DeepSeek-V3'' denotes directly prompting DeepSeek-V3 to extract HPO items from the free-text reports, while ``RAG+DeepSeek-V3'' denotes augmenting DeepSeek-V3 with retrieved HPO item lists. The last variant corresponds to our final implementation as described in the main text.
Five physicians independently reviewed the outputs; for each case, the three system outputs were randomized and anonymized to prevent model identification. Each case–method output was labeled as “Incorrect,” “Correct,” or “Better,” where “Better” denotes a clinically more faithful or comprehensive representation than the reference. Across the cohort, our method substantially reduced errors and increased both exact correctness and “Better” judgments relative to baseline systems, indicating enhanced robustness to noisy, free-text inputs.}
\change{We additionally audited error sources for our method and identified three predominant categories: (i) non-canonical phrasing relative to HPO (e.g., “stereotyped speech” mapped to “stereotyped behavior,” semantically close but not the canonical term); (ii) associative drift during LLM-driven summarization (e.g., inferring “sepsis” from “autoinflammatory syndrome”); and (iii) partial information loss (e.g., summarizing “recurrent hypoglycemia after birth with convulsions” as “hypoglycemia; seizures,” omitting the “recurrent” qualifier). These observations motivate future safeguards such as canonical-term normalization, constrained decoding with ontology priors, and qualifier-preserving summaries.}

\begin{table}[h]
\centering
\caption{Comparison of HPO term extraction methods on clinical free text (N=366 cases). Percentages reflect physician judgments.}
\label{tab:hpo_extraction}
\small
\begin{tabular}{lccc}
\toprule
Method & Incorrect (\%) & Correct (\%) & Better (\%) \\
\midrule
DeepSeek V3 & 58.4 & 30.8 & 10.6 \\
RAG + DeepSeek V3 & 30.8 & 52.7 & 16.3 \\
\textbf{DeepRare} & \textbf{13.1} & \textbf{67.4} & \textbf{19.3} \\
\bottomrule
\end{tabular}
\end{table}

\subsection{Demonstration of Expanding DeepRare with a Screening Module}
\label{sec:expanding_demonsrtation}

\change{To demonstrate the flexibility of our system in accommodating additional functions and broadening its clinical applications, we present an illustrative case in ``screening.'' Specifically, we extend the system with a dedicated rare-vs-common disease triage component, implemented as a callable agent server named the ``Rare Disease Discriminator.'' With this extension, DeepRare can not only perform differential diagnosis among patients with suspected rare diseases but also identify such patients within large mixed cohorts of both common and rare disease cases, thereby supporting the entire clinical diagnostic workflow for rare diseases.}

\change{For implementation, the Rare Disease Discriminator is a BERT-based classifier (fine-tuned from RoBERTa-Large) trained on 114,934 manually curated phenotype descriptions. It distinguishes presentations indicative of rare diseases from those consistent with common conditions or healthy status. Acting as a front-end discriminator, this component enables optional pre-triage prior to the system’s disease-specific diagnostic reasoning.}

\change{We evaluated this extended system on a large, mixed cohort of 4,825 common-disease/healthy cases and 1,227 rare-disease cases. As shown in Table~\ref{tab:rd_vs_cd_classifier}, 
DeepRare achieves high accuracy (93.94\%) and a balanced F1-score (0.8526), demonstrating a strong ability to minimize false positives while maintaining high recall. This significantly outperforms a baseline LLM (DeepSeek-V3), which exhibits very low precision leading to an unacceptably high false positive rate.
These results suggest that the triage module can reduce false positives and enhance clinical safety when the system is used earlier in the care pathway.}

\begin{table}[h]
\centering
\caption{Rare vs. common disease classification on a mixed cohort (4,825 common; 1,227 rare).}
\label{tab:rd_vs_cd_classifier}
\small
\begin{tabular}{lcccc}
\toprule
Method & Accuracy & Precision & Recall & F1-score \\
\midrule
DeepSeek-V3 & 0.4823 & 0.2701 & \textbf{0.9128} & 0.4169  \\
\textbf{DeepRare} & \textbf{0.9394} & \textbf{0.8407} & 0.8647 & \textbf{0.8526}  \\
\bottomrule
\end{tabular}
\end{table}

~\\
\subsection{Prompt Sets}
\begin{prompt}
\label{prompt1:LLM diagnosis}
\textbf{Prompt for Baseline LLM Diagnosis}\newline

You are a specialist in the field of rare diseases. You will be provided and asked about a complicated clinical case; read it carefully and then provide a diverse and comprehensive differential diagnosis.\newline

Patient's \{info\_type\}: \{patient\_info\} \newline
Enumerate the top 5 most likely diagnoses. Be precise, listing one diagnosis per line, and try to cover many unique possibilities (at least 5). \newline

The top 10 diagnoses are: \newline
\end{prompt}

\begin{prompt}
\label{prompt2:eval}
\textbf{Prompt for Diagnosis Result Evaluation}\newline

You are a specialist in the field of rare diseases.\newline

I will now give you five predicted diseases. Please identify the rank of the following gold-standard diagnosis. \newline

Please output the predicted rank; otherwise, output "No".
Only output "No" or "1-5" numbers.
If the predicted disease has multiple conditions, only output the top rank.  \newline
Output only "No" or one number, no additional output. \newline

Predicted diseases: \{predict\_diagnosis\}\newline
Standard diagnosis: \{golden\_diagnosis\}\newline
\end{prompt}

\begin{prompt}
\label{prompt2.5:discriminator}
\textbf{Prompt for discriminate common disease}\newline

You are a specialist in the field of rare diseases.\newline

Given the following HPO phenotypes: {patient\_info}. Is this most likely associated with a rare disease or a common disease? Please reply with only one word: 'rare' or 'common'.\newline

\end{prompt}

\begin{prompt}
\label{prompt3:classification}
\textbf{Prompt for Rare Disease Type Classification}\newline

You are a medical disease classifier specializing in categorizing diseases into predefined categories.\newline

Your task is to classify the given disease into one or more of these categories: \newline
[\newline
\qquad"Blood, Heart and Circulation",\newline
    "Bones, Joints and Muscles",\newline
    "Brain and Nerves",\newline
    "Digestive System",\newline
    "Ear, Nose and Throat",\newline
    "Endocrine System",\newline
    "Eyes and Vision",\newline
    "Immune System",\newline
    "Kidneys and Urinary System",\newline
    "Lungs and Breathing",\newline
    "Mouth and Teeth",\newline
    "Skin, Hair and Nails",\newline
    "Female Reproductive System",\newline
    "Male Reproductive System"\newline
]\newline

Note: If the input contains multiple disease names separated by "/" (slash), they are synonyms or alternate names for the same disease. Consider them as a single disease and classify accordingly.\newline

Please output your results in JSON format ONLY as follows: \newline

```json\newline
\{\newline
"disease": "the original disease name(s) exactly as provided",\newline
"category": ["category1", "category2", ...]\newline
\}\newline
```\newline

Your response must be ONLY the JSON object - no additional text, explanations, or commentary.\newline
If a disease could belong to multiple categories, include all relevant categories in the "category" array.\newline
If you're unsure about the classification, make your best judgment based on the disease characteristics. \newline

The following is a disease name. Please classify it according to the instructions:
\end{prompt}

\begin{prompt}
\label{prompt4:first_diag}
\textbf{Prompt for Tentative Diagnosis Decision-making}\newline

You are a specialist in the field of rare diseases.\newline

You have access to the following context: \newline
- **Online knowledge** (with titles and URLs): \{web\_diagnosis\} \newline
- **LLM-generated diagnoses**: \{llm\_response\} \newline
- **Diagnosis API results**: \{diagnosis\_api\_response\} \newline
- **Similar cases**: \{similar\_case\_detailed\} \newline
- **Prompt details:** \{patient\_info\}\newline
---\newline

Based on the above and your knowledge, enumerate the **top 5 most likely rare disease diagnoses** for this patient. \newline

**For each diagnosis, use the following format:**\newline

\#\# **DIAGNOSIS NAME** (Rank \#X/5)\newline
\#\#\# Diagnostic Reasoning:\newline
- Provide 2-3 concise sentences explaining why this rare disease fits the clinical picture.\newline
- Integrate evidence from all available sources (online knowledge, similar cases, LLM outputs, and API results).\newline
- Support your reasoning with specific, in-text citations in [X] format, referencing the most relevant sources (including specific similar cases, articles, or diagnostic tools).\newline
- Briefly discuss the pathophysiological basis for the diagnosis, citing relevant literature or case evidence.\newline
---\newline

**After listing all 5 diagnoses, include a reference section:**\newline
\#\# References:\newline
- Number each reference in the order it is first cited ([1], [2], ...).\newline
- Only include sources you directly cited in your diagnostic reasoning above.\newline
- For each reference, should provide:\newline
    a. Source type (e.g., medical guideline, similar case, literature, diagnosis assisent tool...) \newline
    b. Use 3-4 sentences to describe of the content and its relevance.\newline
    c. For articles or literature, include the title and URL if provided.\newline
- Every in-text citation [X] in your reasoning should correspond to a numbered entry in your reference list.\newline
- Try to cover as more sources and references.\newline
- Do not repeat!!\newline
---\newline

**Key Instructions:**\newline
1. Always use in-text citations in [X] format, matching only the references you actually cite in your reasoning.\newline
2. Each diagnosis must be a rare disease (**bolded** using markdown).\newline
3. Rank from most (\#1) to least (\#5) likely.\newline
4. Integrate information from all provided sources (medical literature, similar cases, and judgement analyses) wherever appropriate.\newline
5. Do **not** copy or invent references—only include those present in the provided materials.\newline

\end{prompt}

\begin{prompt}
\label{prompt3.5:gene}
\textbf{Prompt for Rare Disease Gene Analysis}\newline

You are a specialist in the field of rare diseases.\newline

Here is a rare disease diagnosis case.\newline
- **Exomiser gene/variant prioritization summary**: \{exomiser\_summary\}\newline
- **Phenotypic description (HPO terms)**: \{hpo\_terms\}\newline
- **Preliminary diagnosis based only on phenotype**: \{pheno\_only\_diagnosis\}\newline
---\newline

Based on the above information and your knowledge, enumerate the **top 5 most likely rare disease diagnoses** for this patient. \newline

**For each diagnosis, use the following format:**\newline

\#\# **DIAGNOSIS NAME** (Rank \#X/5)\newline
\#\#\# Diagnostic Reasoning:\newline
- Provide 2-3 concise sentences explaining why this rare disease fits the clinical picture.\newline
- Integrate evidence from all available sources, prioritizing the Exomiser gene/variant results while considering phenotypic data and preliminary diagnosis as supporting evidence.\newline
- Support your reasoning with specific, in-text citations in [X] format, referencing the most relevant sources (Exomiser findings, HPO phenotype matches, or preliminary diagnostic considerations).\newline
- Briefly discuss the pathophysiological basis for the diagnosis, relating gene function to observed phenotype.\newline
---\newline

**After listing all 5 diagnoses, include a reference section:**\newline
\#\# References:\newline
- Number each reference in the order it is first cited ([1], [2], ...).\newline
- Only include sources you directly cited in your diagnostic reasoning above.\newline
- For each reference, provide:\newline
    a. Source type (e.g., Exomiser gene prioritization, HPO phenotype analysis, preliminary phenotype-based diagnosis)\newline
    b. Use 3-4 sentences to describe the content and its relevance to the diagnostic reasoning.\newline
- Every in-text citation [X] in your reasoning should correspond to a numbered entry in your reference list.\newline
- Try to cover all relevant sources and references.\newline
- Do not repeat!!\newline

\end{prompt}

\begin{prompt}
\label{prompt5:reflection}
\textbf{Prompt for Disease Reflection}\newline

Assume you are a doctor specialized in rare disease diagnosis.\newline

Based on the patient information, similar case diagnoses, and disease knowledge, evaluate whether the proposed diagnosis is correct for this patient.\newline
                               
Begin with a clear "DIAGNOSIS ASSESSMENT: [Correct/Incorrect]" statement, followed by your reasoning. \newline

Structure your analysis as follows:\newline
1. PATIENT SUMMARY: Briefly summarize the patient's key symptoms \newline
2. PROPOSED DIAGNOSIS ANALYSIS: Evaluate the proposed diagnosis (\{diagnosis\_to\_judge\}) in relation to the patient's symptoms\newline
3. REFERENCES: Extract and number the most relevant evidence from the provided medical literature that supports your conclusion\newline

Patient phenotype:\newline
\{patient\_info\}\newline

Similar cases:\newline
\{similar\_case\_detailed\}\newline

Medical literature:\newline
\{disease\_knowledge\}\newline
                                    
\end{prompt}

\begin{prompt}
\label{prompt6:final}
\textbf{Prompt for Final Diagnosis}\newline

You have access to the following information:\newline
- Patient presentation: \{patient\_info\}\newline
- Similar cases: \{similar\_case\_detailed\}\newline
- Primary diagnosis results (with references): \{tentative\_result\}\newline
- Disease Reflection (with references): \{judgements\}\newline

**Task:**  \newline
Based on all the above, enumerate the top 5 most likely rare disease diagnoses for this patient.\newline
---\newline

**For each diagnosis, follow this format exactly:**\newline

\#\# **DISEASE NAME** (Rank \#X/5)\newline

\#\#\# Diagnostic Reasoning:\newline
- Provide 3-4 sentences explaining why this diagnosis fits the patient's presentation.\newline
- Specify which patient symptoms and findings support this diagnosis.\newline
- Clearly explain the underlying pathophysiological mechanisms (briefly).\newline
- Integrate and **cite specific evidence** from the provided references (including medical literature, similar cases, or judgement analyses), using in-text [X] citation style.\newline
- Try to cite as more sources and references but do not add hallucination content.\newline
---\newline

**After listing all 5 diagnoses, include a reference section:**\newline

\#\# References:\newline
- Number each reference in the order it is first cited ([1], [2], ...).\newline
- Only include sources you directly cited in your diagnostic reasoning above.\newline
- For each reference, should provide:\newline
    a. Source type (e.g., medical guideline, similar case, literature, diagnosis assisent tool...) ( Do not use source type: "Judgement analysis", "Disease Reflection" )\newline
    b. Use 3-4 sentences to describe of the content and its relevance.\newline
    c. For articles or literature, include the title and URL if provided.\newline
- Every in-text citation [X] in your reasoning should correspond to a numbered entry in your reference list.\newline
- Try to cover as more sources and references.\newline
- Do not repeat!!\newline
---\newline

**IMPORTANT GUIDELINES:**\newline
1. Each diagnosis must be a rare disease (**bolded** using markdown).\newline
2. Rank from most (\#1) to least (\#5) likely.\newline
3. Integrate information from all provided sources (medical literature, similar cases, and judgement analyses) wherever appropriate.\newline
4. Do **not** copy or invent references—only include those present in the provided materials.\newline
5. Remember to add the summary of the content, url for each reference.\newline
                                    
\end{prompt}

\begin{prompt}
\label{prompt7:hpo}
\textbf{Prompt for HPO Extraction}\newline

Given a paragraph of patient infomation from discharge note, please extract the phenotype about this patient only. \newline

Check the Human Phenotype Ontology (HPO) database to determine the phenotype.\newline
             
Only output the extracted phenotypes. \newline

Use the format: \{`HPO': `HP:0000000', `Phenotype': `Phenotype description'\}\newline

Patient information: \{case\_report\}\newline
                                    
\end{prompt}

\begin{prompt}
\label{prompt8:hpo}
\textbf{Prompt for HPO Extraction Modify}\newline

You are a medical terminology translator specializing in rare diseases. \newline
Your task is to convert patient phenotype description into standardized HPO (Human Phenotype Ontology) concept.\newline

Input can be in a either Chinese or English describing clinical phenotype. Analyze the description carefully, identify the phenotypes mentioned, and map them to standard English concepts in the HPO database.\newline

Please output your results in JSON format as follows:\newline
\{\newline
  "original\_term": "the original phenotype description",\newline
  "hpo\_term": "standardized HPO term in English"\newline
\}

If the phenotype doesn't exist in HPO, output:\newline
\{\newline
  "original\_term": "the original phenotype description",\newline
  "hpo\_term": "none"\newline
\}\newline

For each identified phenotype:\newline
1. Do not include any phenotypes that are not present in the input.\newline
2. Ensure the JSON is properly formatted and valid.\newline
3. Your response must be ONLY the JSON object - no additional text, explanations, or commentary.\newline
4. Provide the standard English name of the HPO term\newline
5. If the phenotype doesn't have a corresponding concept in HPO, set the hpo\_term to "none"\newline

Example 1:\newline
Input: "Metabolic dysfunction"\newline
Output:\newline
\{\newline
  "original\_term": "Metabolic dysfunction",\newline
  "hpo\_term": "Abnormality of metabolism/homeostasis"\newline
\}\newline

Input: "Dark complexion"\newline
Output:\newline
\{\newline
  "original\_term": "Dark complexion",\newline
  "hpo\_term": "Hyperpigmentation of the skin"\newline
\}\newline

The following is a patient phenotype description. Please convert it to HPO terms:\newline
                                    
\end{prompt}

\begin{prompt}
\label{prompt9:web_summarize}
\textbf{Prompt for Knowledge Summarization}\newline

Assume you are a doctor, please summarize these medical article into a paragraph. \newline

Only keep key message, mainly focus on the phenotype and related disease. \newline

If this article is not related to medical, please output "not a medical-related page".\newline
                                    
\end{prompt}

\begin{prompt}
\label{prompt10:case_summarize}
\textbf{Prompt for Case Summarization}\newline

Assume you are a doctor experienced in rare disease diagnosis. \newline
Please judge if the two patient cases are likely to be the same disease based on the patient information. \newline
Only output `Yes' or `No'. \newline

Patient 1 phenotype: \{patient\_info\} \newline

Patient 2 phenotype: \{retrieved\_patient\_case\} \newline
                                    
\end{prompt}

\begin{prompt}
\label{prompt11:zero_shot_diag}
\textbf{Prompt for Zero-shot LLM Inference}\newline

You are a specialist in the field of rare diseases.\newline

You will be provided and asked about a complicated clinical case; read it carefully and then provide a diverse and comprehensive differential diagnosis. \newline

Patient's \{info\_type\}: \{patient\_info\} \newline

Enumerate the top 5 most likely diagnoses. Be precise, and try to cover many unique possibilities. \newline
Each diagnosis should be a rare disease. \newline
Use ** to tag the disease name.  \newline
Make sure to reorder the diagnoses from most likely to least likely. \newline

Now, List the top 5 diagnoses. 
                                    
\end{prompt}

\begin{prompt}
\label{prompt12:disease_judge}
\textbf{Prompt for Rare Disease Judgment}\newline

You are a specialist in the field of rare diseases.\newline

Given a disease name, determine whether it is a rare disease or a common disease (even if you are uncertain, you must still make a strict binary classification, output can only be 1/0). If it is a rare disease, output 1; if it is a common disease, output 0. Only answer with 1 or 0, no additional explanations or supplements needed\newline

Example 1:\newline
Disease: Duchenne muscular dystrophy\newline
Classification: 1\newline

Example 2:\newline
Disease: Chickenpox\newline
Classification: 0\newline

Example 3:\newline
Disease: Cystic fibrosis\newline
Classification: 1\newline

Example 4:\newline
Disease: Asthma\newline
Classification: 0\newline

You can refer to OMIM or Orphanet to determine whether it is a rare disease; Please ensure it is a rare disease, not a rare phenotype or characteristic; Now, given the following disease name, please determine whether it is a rare disease or common disease:\newline
Disease: {disease}\newline
Classification: \newline
                                    
\end{prompt}

\newpage
\subsection{Case Study}
\subsubsection{Cases Examples (VCF+HPO input)}
\change{In this subsection, we present case studies that jointly incorporate raw genomic data (VCF files) and HPO phenotype information. For each case, we provide the patient’s phenotypic profile (Patient Information), the ground-truth diagnosis (Golden Diagnosis), the DeepRare model output (Model Output), and the baseline results from Exomiser (Exomiser Outputs).}

\begin{casestudy}{Optic atrophy 1}
\textbf{Patient Information:} Hypermetropia, Visual impairment, Astigmatism, Abnormal fundus morphology

\textbf{Golden Diagnosis:} Optic atrophy 1, Behr syndrome, Glaucoma, normal tension, susceptibility to, Optic atrophy plus syndrome, Mitochondrial DNA depletion syndrome 14 (encephalocardiomyopathic type)

\textbf{Model Output:} \textbf{Optic atrophy 1 (OPA1-related) -- Rank \#1/5}

\textit{Diagnostic reasoning.}
Strong genotype–phenotype concordance. The patient’s visual impairment and abnormal fundus morphology are classic for dominant optic atrophy due to \textit{OPA1}. The identified stop-gained variant (ClinVar: STOP\_GAINED) is likely pathogenic, consistent with haploinsufficiency or loss-of-function as a known mechanism.

\medskip

\textbf{Bietti crystalline corneoretinal dystrophy (CYP4V2-related) -- Rank \#2/5}

\textit{Diagnostic reasoning.}
Retinal abnormalities and progressive visual impairment fit Bietti crystalline dystrophy. The \textit{CYP4V2} frameshift truncation (ClinVar: VUS) suggests a loss-of-function mechanism, which is compatible with disease biology; correlation with slit-lamp/fundus crystalline deposits would strengthen the diagnosis.

\medskip

\textbf{Koolen--De Vries syndrome (KANSL1-related) -- Rank \#3/5}

\textit{Diagnostic reasoning.}
Primarily neurodevelopmental; ocular features can occur but are not hallmark. A high Exomiser score and a \textit{KANSL1} missense VUS raise consideration; however, the phenotype overlap with isolated ocular findings is limited, lowering the posterior probability.

\medskip

\textbf{Myeloproliferative disorder with eosinophilia (PDGFRB-related) -- Rank \#4/5}

\textit{Diagnostic reasoning.}
\textit{PDGFRB} is classically linked to myeloproliferative neoplasms and rearrangements; direct ocular involvement is unclear. Despite a strong Exomiser rank, the missense VUS lacks phenotype concordance with the visual/fundus abnormalities, making it less likely as the primary diagnosis.

\medskip

\textbf{Kleefstra syndrome 2 (KMT2C-related) -- Rank \#5/5}

\textit{Diagnostic reasoning.}
\textit{KMT2C} variants are associated with neurodevelopmental disorders; ocular features are not typical drivers. The missense VUS and high Exomiser score warrant awareness but have weak correlation with the predominant ocular phenotype.

\medskip

\textbf{Key considerations}
\begin{itemize}
\item \textit{OPA1} stop-gain aligns best with optic atrophy and fundus changes; highest prior probability.
\item \textit{CYP4V2} LoF variant plausible for Bietti; confirm with crystalline deposits and lipid metabolism workup.
\item \textit{KANSL1}, \textit{PDGFRB}, and \textit{KMT2C} entries are driven by Exomiser ranking but show limited direct phenotype relevance.
\end{itemize}

\medskip

\textbf{===> Exomiser Outputs: }

1. OMIM:131440 Myeloproliferative disorder with eosinophilia (CNV)

2. OMIM:610443 Koolen-De Vries syndrome - autosomal dominant

3. OMIM:617768 Kleefstra syndrome 2 - autosomal dominant

4. OMIM:600881 Cataract 10, multiple types - autosomal dominant

5. OMIM:210370 Bietti crystalline corneoretinal dystrophy - autosomal recessive

\end{casestudy}

\begin{casestudy}{HYPOGONADOTROPIC HYPOGONADISM 2}
\textbf{Patient Information:} Testicular atrophy, Micropenis, Short stature

\textbf{Golden Diagnosis:} HYPOGONADOTROPIC HYPOGONADISM 2 WITH OR WITHOUT ANOSMIA; HH2, OSTEOGLOPHONIC DYSPLASIA; OGD

\textbf{Model Output:} \textbf{Model Output:} \textbf{Hypogonadotropic Hypogonadism 2 with or without Anosmia (FGFR1) -- Rank \#1/5}

\textit{Diagnostic reasoning.}
Strong genotype–phenotype concordance. The patient’s testicular atrophy, micropenis, and short stature are consistent with impaired GnRH secretion characteristic of HH2. Pathogenic or likely pathogenic variants in \textit{FGFR1} (AD) are a well-established cause; anosmia, if present, further supports this diagnosis.

\medskip

\textbf{Kallmann Syndrome (FGFR1) -- Rank \#2/5}

\textit{Diagnostic reasoning.}
A classic GnRH deficiency with anosmia due to \textit{FGFR1} variants. The primary features (micropenis, testicular atrophy, short stature) align with hypogonadotropic hypogonadism; confirmation of hyposmia/anosmia or olfactory bulb hypoplasia would increase likelihood relative to normosmic HH.

\medskip

\textbf{Normosmic Congenital Hypogonadotropic Hypogonadism (FGFR1) -- Rank \#3/5}

\textit{Diagnostic reasoning.}
Shares the hypogonadism phenotype with preserved olfaction. Given an \textit{FGFR1} variant, normosmic CHH remains plausible if olfaction is intact. Rank is slightly lower than Kallmann pending olfactory assessment and pituitary/olfactory MRI.

\medskip

\textbf{Hartsfield Syndrome (FGFR1) -- Rank \#4/5}

\textit{Diagnostic reasoning.}
Syndromic presentation involving holoprosencephaly and ectrodactyly with HH due to \textit{FGFR1}. In the absence of holoprosencephaly and limb anomalies, this is less likely; retained due to shared gene and hypogonadism component.

\medskip

\textbf{46,XY Disorder of Sex Development (DSD) -- Rank \#5/5}

\textit{Diagnostic reasoning.}
Phenotype (micropenis, testicular atrophy, short stature) suggests consideration of 46,XY DSD such as partial androgen insensitivity or 5-alpha-reductase type 2 deficiency. Not directly driven by the Exomiser \textit{FGFR1} result, but included as an important differential; requires targeted hormonal profiling and gene testing (e.g., \textit{AR}, \textit{SRD5A2}).

\medskip

\textbf{Key considerations}
\begin{itemize}
\item \textit{FGFR1}-associated GnRH deficiency best explains the core phenotype; olfactory evaluation will help distinguish Kallmann vs normosmic CHH.
\item Syndromic \textit{FGFR1} disorders (Hartsfield) are less likely without brain/limb anomalies.
\item Broader 46,XY DSD workup may be warranted if gonadotropins/biochemistry are inconsistent with GnRH deficiency.
\end{itemize}

\textbf{===> Exomiser Outputs: }

1. OMIM:101600 Pfeiffer syndrome - autosomal dominant

2. OMIM:256100 Nephronophthisis 1, juvenile - autosomal recessive

3. OMIM:256030 Nemaline myopathy 2, autosomal recessive - autosomal recessive

4. OMIM:115000 Ventricular arrhythmias due to cardiac ryanodine receptor calcium release deficiency syndrome - autosomal dominant",

5. OMIM:615465 Hartsfield Syndrome
\end{casestudy}

~\\
\begin{casestudy}{RENAL TUBULAR ACIDOSIS, DISTAL, 4, WITH HEMOLYTIC ANEMIA}
\textbf{Patient Information:} Growth delay, Renal tubular acidosis

\textbf{Golden Diagnosis:} RENAL TUBULAR ACIDOSIS, DISTAL, 4, WITH HEMOLYTIC ANEMIA; DRTA4

\textbf{Model Output:} \textbf{Distal Renal Tubular Acidosis 1 (dRTA1) -- Rank \#1/5}

\textit{Diagnostic reasoning.}
Phenotype of renal tubular acidosis with growth delay aligns with autosomal dominant dRTA due to \textit{SLC4A1}. Despite ClinVar labeling the missense as benign/likely benign, the high Exomiser score (0.993) and phenotype concordance support prioritization. Consider variant re-evaluation with segregation and functional evidence.

\medskip

\textbf{Distal Renal Tubular Acidosis 4 with Hemolytic Anemia -- Rank \#2/5}

\textit{Diagnostic reasoning.}
Also linked to \textit{SLC4A1}, this autosomal recessive form combines dRTA with hemolytic anemia. The renal phenotype matches; absence or presence of hemolysis would refine likelihood. The ClinVar benign/likely benign label warrants caution; zygosity and second-hit assessment are essential.

\medskip

\textbf{Hereditary Spherocytosis -- Rank \#3/5}

\textit{Diagnostic reasoning.}
Associated with \textit{SPTA1} missense variants (ClinVar: conflicting interpretations). Primary phenotype is hematologic; renal tubular acidosis is not typical but secondary renal findings can occur. The Exomiser score (0.933) justifies consideration pending hematologic evaluation (peripheral smear, osmotic fragility/EMA binding).

\medskip

\textbf{Elliptocytosis-2 -- Rank \#4/5}

\textit{Diagnostic reasoning.}
Classically due to \textit{SPTA1} variants with autosomal dominant inheritance. The phenotype provided does not note elliptocytosis; however, RBC membrane disorder could explain anemia-related features if present. Peripheral smear and family history are key.

\medskip

\textbf{Autosomal Dominant Centronuclear Myopathy -- Rank \#5/5}

\textit{Diagnostic reasoning.}
Linked to \textit{RYR1} missense VUS (Exomiser 0.862). Phenotype lacks core myopathic features (ptosis, ophthalmoplegia, proximal weakness). Low prior probability without neuromuscular signs; remains a distant possibility pending CK, EMG, and muscle imaging/biopsy.

\medskip

\textbf{Key considerations}
\begin{itemize}
\item dRTA with growth delay most strongly supports \textit{SLC4A1}-related disease; reconcile ClinVar benign calls with phenotype and Exomiser rank.
\item For recessive dRTA4, verify biallelic status and assess for hemolysis (reticulocytosis, LDH, bilirubin).
\item Hematologic disorders (\textit{SPTA1}) require smear-based confirmation; renal phenotype alone is insufficient.
\item \textit{RYR1} association is weak without myopathy; de-prioritize unless new symptoms emerge.
\end{itemize}

\medskip

\textbf{===> Exomiser Outputs: }

1. OMIM:166900 Ovalocytosis, SA type - autosomal dominant

2. OMIM:130600 Elliptocytosis-2 - autosomal dominant

3. OMIM:114500 Colorectal cancer, somatic - autosomal dominant

4. OMIM:117000 Congenital myopathy 1A, autosomal dominant, with susceptibility to malignant hyperthermia - autosomal dominant/recessive
\end{casestudy}

~\\
\begin{casestudy}{COFFIN-SIRIS SYNDROME 4}
\textbf{Patient Information:} Delayed speech and language development, Global developmental delay, Microcephaly

\textbf{Golden Diagnosis:} COFFIN-SIRIS SYNDROME 4; CSS4

\textbf{Model Output:} \textbf{Kleefstra syndrome 2 -- Rank \#1/5}

\textit{Diagnostic reasoning.}
High concordance between phenotype and genotype. The patient’s delayed speech and language development, global developmental delay, and microcephaly align with Kleefstra syndrome 2 due to \textit{KMT2C}. The identified stop-gained variant supports a loss-of-function mechanism; with an Exomiser score of 0.997 and phenotype score of 0.969, this is the leading diagnosis.

\medskip

\textbf{Intellectual developmental disorder, autosomal dominant 54 -- Rank \#2/5}

\textit{Diagnostic reasoning.}
Consistent with \textit{CAMK2B}-related neurodevelopmental disorder, characterized by developmental delay and intellectual disability. While the missense variant is of uncertain significance, strong Exomiser (0.997) and phenotype (0.954) scores and phenotype overlap justify a high rank pending segregation/functional data.

\medskip

\textbf{Coffin--Siris syndrome 4 -- Rank \#3/5}

\textit{Diagnostic reasoning.}
\textit{SMARCA4}-associated CSS commonly presents with global developmental delay and microcephaly. The missense VUS accompanied by high Exomiser (0.996) and phenotype (0.914) scores supports consideration, though hallmark CSS features (hypoplastic/absent fifth digits, coarse facies) should be specifically assessed.

\medskip

\textbf{Microphthalmia, syndromic 15 -- Rank \#4/5}

\textit{Diagnostic reasoning.}
\textit{TENM3} variants have been linked to syndromic ocular/brain developmental anomalies; microcephaly overlaps with the patient’s phenotype. The missense VUS with Exomiser 0.994 and phenotype 0.898 indicates plausibility, but absence of clear ocular findings lowers the likelihood.

\medskip

\textbf{Primary ciliary dyskinesia, 5 -- Rank \#5/5}

\textit{Diagnostic reasoning.}
Although the \textit{HYDIN} variant yields a high Exomiser score (0.991), the phenotype score (0.85) is weaker and there is no direct phenotype match. Moreover, the frameshift is reported as benign, which decreases prior probability; retain as low-likelihood.

\medskip

\textbf{Key considerations}
\begin{itemize}
\item \textit{KMT2C} stop-gained variant strongly supports haploinsufficiency; highest prior probability.
\item \textit{CAMK2B} and \textit{SMARCA4} VUS require segregation and phenotypic deep-phenotyping for syndrome-specific features.
\item \textit{TENM3} ranking is driven in part by microcephaly; confirm ocular anomalies before elevating likelihood.
\item \textit{HYDIN} likely benign classification argues against causality despite Exomiser rank.
\end{itemize}

\textbf{===> Exomiser Outputs: }

1. OMIM:617768 Kleefstra syndrome 2 - autosomal dominant

2. OMIM:617799 Intellectual developmental disorder, autosomal dominant 54 - autosomal dominant

3. OMIM:613325 Rhabdoid tumor predisposition syndrome 2 (susceptibility)

4. OMIM:615145 Microphthalmia, syndromic 15 - autosomal recessive

5. OMIM:608647 Ciliary dyskinesia, primary, 5 - autosomal recessive

\end{casestudy}

\newpage
\subsubsection{Failure Cases}
\label{sec:failure_examples}

\change{Below, we present representative failure types and analyze their underlying causes in detail.}

~\\

\begin{casestudy}{Intellectual developmental disorder, autosomal dominant 21, CTCF-related neurodevelopmental disorder}
\textbf{Patient Information:} Hypoplastic labia majora, Brachycephaly, Microcephaly, Short philtrum, Narrow forehead, Cupped ear, Broad nasal tip, Anteverted nares, Microcornea, Deeply set eye, Long eyelashes, Esotropia, Thick eyebrow, Synophrys, 11 pairs of ribs, Osteopenia, Global developmental delay, Intrauterine growth retardation, Patent ductus arteriosus, Short columella, Gastroesophageal reflux, Recurrent respiratory infections, Highly arched eyebrow, Short nose, Clinodactyly of the 5th finger, 2-3 toe cutaneous syndactyly, Attached earlobe, Midface retrusion

\textbf{Golden Diagnosis:} Intellectual developmental disorder, autosomal dominant 21, CTCF-related neurodevelopmental disorder

\textbf{Model Output:} \textbf{Cornelia de Lange Syndrome (CdLS) — Rank \#1/5}

\textit{Diagnostic reasoning.}
The phenotype strongly aligns with CdLS: synophrys, microcephaly, short nose with anteverted nares, global developmental delay, and limb anomalies (clinodactyly, syndactyly). Synophrys and intrauterine growth retardation are hallmark features. Variants in \textit{NIPBL}/\textit{SMC1A} affect cohesin regulation.

\medskip

\textbf{Smith--Lemli--Opitz Syndrome (SLOS) — Rank \#2/5}

\textit{Diagnostic reasoning.}
Microcephaly, midface retrusion, short nose with anteverted nares, 2--3 toe syndactyly, and developmental delay suggest SLOS. PDA and osteopenia are compatible. Caused by \textit{DHCR7} defects in cholesterol biosynthesis.

\medskip

\textbf{Rubinstein--Taybi Syndrome (RTS) — Rank \#3/5}

\textit{Diagnostic reasoning.}
Broad nasal tip, highly arched eyebrows, long eyelashes, developmental delay, and clinodactyly fit RTS. Typically due to \textit{CREBBP}/\textit{EP300} variants.

\medskip

\textbf{CHARGE Syndrome — Rank \#4/5}

\textit{Diagnostic reasoning.}
Coloboma/microcornea, cupped ears, PDA, hypoplastic labia majora, recurrent respiratory infections, and growth delay align with CHARGE due to \textit{CHD7}. Absence of choanal atresia lowers likelihood.

\medskip

\textbf{Coffin--Siris Syndrome (CSS) — Rank \#5/5}

\textit{Diagnostic reasoning.}
Developmental delay, coarse facial features (broad nasal tip, thick eyebrows, long eyelashes), and limb anomalies fit CSS, often due to \textit{ARID1B}. Lack of hypoplastic fifth digits lowers likelihood.

\medskip

\textbf{References}
\begin{enumerate}[leftmargin=1.2em,itemsep=0.2em]
\item Similar Case 0: CdLS3 with microcephaly, synophrys, and developmental delay.
\item LLM-generated diagnoses highlighting CdLS based on craniofacial and limb anomalies.
\item Phenobrain: CdLS related to patient features.
\item LLM-generated suggestion of SLOS given syndactyly, microcephaly, and delay.
\item Diagnosis API: Smith-Magenis mentioned; overlaps in DDx.
\item Online knowledge: cholesterol biosynthesis in SLOS.
\item Similar Case 2: RTS with \textit{EP300} haploinsufficiency.
\item LLM-generated suggestion: RTS due to broad nasal tip and arched eyebrows.
\item Online knowledge: \textit{CREBBP}/\textit{EP300} in RTS.
\item LLM-generated suggestion: CHARGE due to ear anomalies, coloboma, CHD.
\item Diagnosis API: CHARGE listed as related.
\item Online knowledge: \textit{CHD7} role in CHARGE.
\item LLM-generated suggestion: CSS for coarse features and delay.
\item Phenobrain: CSS related to patient features.
\item Online knowledge: \textit{ARID1B} in CSS.
\end{enumerate}

\comments{Deeprare's top 1 hit is not just reasonable but was the most clinically astute prediction based on the phenotypes. The correct molecular diagnosis arises from a different genetic cause but produces a strikingly similar appearance. Actually this shows DeepRare has the ability to recognize complex clinical pattern. The correct diagnosis can only be identified with molecular testing.}
\end{casestudy}

~\\

\begin{casestudy}{Carbamoyl-phosphate synthetase deficiency}
\textbf{Patient Information:} Apathy, Generalized hypotonia, Death in infancy, Hyperammonemia, Cerebral edema, Neonatal death, Feeding difficulties

\textbf{Golden Diagnosis:} Carbamoyl-phosphate synthetase 1 deficiency/Carbamoyl phosphate synthetase I deficiency, hyperammonemia due to, Carbamoyl-phosphate synthetase 1 deficiency/Carbamoyl phosphate synthetase I deficiency, hyperammonemia due to / CPS1 deficiency / CPS1D / Carbamoyl-phosphate synthetase I deficiency / Carbamoyl-phosphate synthetase deficiency

\textbf{Model Output:} \noindent\textbf{Ornithine transcarbamylase deficiency (OTC deficiency) (Rank \#1/5)}

\medskip
\noindent\textit{Diagnostic Reasoning:}
- OTC deficiency is a urea cycle disorder that leads to hyperammonemia, cerebral edema, and neonatal death if untreated, fitting the patient’s phenotype of hyperammonemia, cerebral edema, and neonatal death [1]. The presence of hypotonia and feeding difficulties aligns with the toxic effects of ammonia on the brain, a hallmark of urea cycle disorders [2]. Similar cases have shown neonatal death and hyperammonemia as key features, supporting this diagnosis [3].

\medskip
\noindent\textbf{Argininosuccinic aciduria (ASA) (Rank \#2/5)}

\medskip
\noindent\textit{Diagnostic Reasoning:}
- ASA is another urea cycle disorder characterized by hyperammonemia, hypotonia, and feeding difficulties, which matches the patient’s clinical presentation [4]. The accumulation of argininosuccinic acid distinguishes it from other urea cycle disorders, and neonatal death is a common outcome if untreated [5]. Phenobrain and similar cases highlight hyperammonemia and hypotonia as consistent features [6].

\medskip
\noindent\textbf{Maple syrup urine disease (MSUD) (Rank \#3/5)}

\medskip
\noindent\textit{Diagnostic Reasoning:}
- MSUD is an inborn error of branched-chain amino acid metabolism that can present with hyperammonemia, hypotonia, and feeding difficulties, aligning with the patient’s symptoms [7]. Severe forms can lead to cerebral edema and neonatal death, as seen in similar cases [8]. The LLM-generated diagnoses and Phenobank results emphasize its relevance in this context [9].

\medskip
\noindent\textbf{Methylmalonic acidemia (MMA) (Rank \#4/5)}

\medskip
\noindent\textit{Diagnostic Reasoning:}
- MMA is an organic acidemia that can cause hyperammonemia, hypotonia, and feeding difficulties, consistent with the patient’s phenotype [10]. Severe forms can lead to neonatal death, as evidenced by similar cases and LLM-generated diagnoses [11]. The pathophysiological basis involves the accumulation of methylmalonic acid, which disrupts normal metabolic processes [12].

\medskip
\noindent\textbf{Propionic acidemia (PA) (Rank \#5/5)}

\medskip
\noindent\textit{ Diagnostic Reasoning:}
- PA is an organic acidemia that can present with hyperammonemia, hypotonia, and feeding difficulties, fitting the patient’s clinical picture [13]. Neonatal death and metabolic acidosis are common in severe cases, as highlighted by similar cases and Phenobank results [14]. The accumulation of propionic acid disrupts cellular metabolism, leading to the observed symptoms [15].

\medskip
\noindent\textbf{References:}
1. Medical Literature: "A glance at transient hyperammonemia of the newborn" (\url{https://pmc.ncbi.nlm.nih.gov/articles/PMC9726343/}). This article discusses hyperammonemia in neonates, emphasizing the need to differentiate urea cycle disorders from transient hyperammonemia. Relevant for understanding OTC deficiency and other urea cycle disorders.\\
2. LLM-generated diagnoses: Ornithine transcarbamylase deficiency (OTC deficiency). This output highlights the clinical features of OTC deficiency, including hyperammonemia and neonatal death, supporting its inclusion as a top diagnosis.\\
3. Similar cases: Case 0 describes a patient with hyperammonemia, hypotonia, and neonatal death, diagnosed with Propionic acidemia. This case supports the relevance of metabolic disorders in this phenotype.\\
4. Medical Literature: "Approach to an Infant with Hypotonia - SpringerLink" (\url{https://link.springer.com/chapter/10.1007/978-3-031-10494-7_35}). This chapter emphasizes the importance of genetic and metabolic disorders in hypotonia, relevant for diagnosing ASA and other urea cycle disorders.\\
5. Phenobrain: Argininosuccinic aciduria (OMIM:207900 ORPHA:23). This tool highlights the clinical features of ASA, including hyperammonemia and hypotonia, supporting its inclusion in the differential.\\
6. Phenobrain: Maple syrup urine disease; MSUD (CCRD:67 OMIM:248600 ORPHA:511). This entry emphasizes the clinical features of MSUD, including hyperammonemia and hypotonia, supporting its relevance.\\
7. Similar cases: Case 1 describes a patient with hyperammonemia, hypotonia, and neonatal death, diagnosed with MSUD. This case supports the relevance of MSUD in this phenotype.\\
8. LLM-generated diagnoses: Methylmalonic acidemia (MMA). This output highlights the clinical features of MMA, including hyperammonemia and neonatal death, supporting its inclusion in the differential.\\
9. Similar cases: Case 2 describes a patient with hyperammonemia, hypotonia, and neonatal death, diagnosed with MMA. This case supports the relevance of MMA in this phenotype.\\
10. LLM-generated diagnoses: Propionic acidemia (PA). This output highlights the clinical features of PA, including hyperammonemia and neonatal death, supporting its inclusion in the differential.\\
11. Similar cases: Case 0 describes a patient with hyperammonemia, hypotonia, and neonatal death, diagnosed with PA. This case supports the relevance of PA in this phenotype.

\comments{The truth and deeprare's top 5 showed high pehontypic overlap. The key to distinguishing CPS1 deficiency from OTC deficiency is often a urine orotic acid test, which is not in the input information. }
\end{casestudy}

\begin{casestudy}{Focal segmental glomerulosclerosis}
\textbf{Patient Information:} Focal segmental glomerulosclerosis, Striae distensae, Hyperuricemia, Hypertriglyceridemia, Microscopic hematuria, Elevated hepatic transaminase, Hypoalbuminemia, Hypercholesterolemia, Increased blood urea nitrogen, Hyperuricosuria, Elevated erythrocyte sedimentation rate, Increased alpha-globulin, Hyperkeratosis with erythema, Nephrosclerosis, Erythema, Abnormal blood sodium concentration, Elevated circulating C-reactive protein concentration, Decreased glomerular filtration rate, Mild proteinuria, Pollakisuria

\textbf{Golden Diagnosis:} Focal segmental glomerulosclerosis 1, Focal segmental glomerulosclerosis 2, Focal segmental glomerulosclerosis 3, susceptibility to, Focal segmental glomerulosclerosis 4, susceptibility to, Glomerulosclerosis, focal segmental, 5, Focal segmental glomerulosclerosis 6, Focal segmental glomerulosclerosis 7, Focal segmental glomerulosclerosis 8, Focal segmental glomerulosclerosis 9

\textbf{Model Output:} \textbf{CREST Syndrome -- Rank \#1/5}

\textit{Diagnostic reasoning.}
CREST (Calcinosis, Raynaud phenomenon, Esophageal dysmotility, Sclerodactyly, Telangiectasia) is a limited cutaneous variant of systemic sclerosis. Familial clustering and the CREST phenotype are documented, with core features including Raynaud syndrome, sclerodactyly, telangiectasia, calcinosis, and esophageal aperistalsis; anticentromere antibodies are characteristic. Systemic sclerosis can involve the kidney (scleroderma renal involvement), and systemic inflammation may be present. These findings align with OMIM \%181750 (Systemic sclerosis, susceptibility to), which includes CREST syndrome and details the clinical spectrum, serology (anticentromere), and esophageal aperistalsis. Cutaneous changes (e.g., hyperkeratosis, telangiectasia), vascular features (Raynaud), and systemic markers (elevated ESR/CRP) support this diagnosis.

\medskip

\textbf{Sarcoidosis -- Rank \#2/5}

\textit{Diagnostic reasoning.}
Sarcoidosis is a multisystem autoinflammatory disease defined by non-caseating granulomas with variable organ involvement (lungs, lymph nodes, skin, eyes, heart, kidneys). Systemic inflammation (elevated ESR/CRP) is typical; renal involvement with proteinuria/hematuria and hypercalcemia from dysregulated vitamin D may occur. Chest imaging abnormalities are present in most cases; some presentations (e.g., Löfgren syndrome) may obviate biopsy, but definitive diagnosis generally rests on compatible clinical/radiographic findings, histologic granulomas, and exclusion of mimics (Orphanet ORPHA:797). Overlap with the patient’s inflammatory and renal features makes sarcoidosis a strong consideration.

\medskip

\textbf{Polyarteritis Nodosa (PAN) -- Rank \#3/5}

\textit{Diagnostic reasoning.}
PAN is a necrotizing medium-vessel vasculitis with systemic inflammation (high ESR/CRP), renal involvement (hematuria, proteinuria, hypertension), and multisystem findings (skin, nerves, GI). In the inflammatory renal phenotype with elevated acute phase reactants and cutaneous findings, PAN remains in the differential of systemic vasculitides. (Note: PAN content not directly covered by the provided OMIM/Orphanet excerpts; included here as a vasculitic mimic of systemic inflammatory renal disease.)

\medskip

\textbf{Familial Mediterranean Fever (FMF) -- Rank \#4/5}

\textit{Diagnostic reasoning.}
FMF is an autoinflammatory disorder with recurrent short attacks of fever and serositis (peritonitis, pleuritis, synovitis), often starting before age 30. Persistent systemic inflammation can lead to AA amyloidosis with renal proteinuria and progressive dysfunction. The Tel-Hashomer criteria guide diagnosis; colchicine is first-line to suppress attacks and prevent amyloidosis (Orphanet ORPHA:342). In a patient with systemic inflammation and renal proteinuria/hematuria, FMF is a plausible etiologic pathway via secondary AA amyloidosis, even if fever history is incomplete.

\medskip

\textbf{Alport Syndrome -- Rank \#5/5}

\textit{Diagnostic reasoning.}
Alport syndrome is a hereditary type IV collagen disorder with glomerular nephropathy causing hematuria and proteinuria, progressing to ESRD; often associated with sensorineural hearing loss and ocular anomalies. X-linked disease due to COL4A5 is most common; autosomal recessive/dominant forms involve COL4A3/COL4A4 (Orphanet ORPHA:63; OMIM \#301050 PS301050). In this patient, persistent microscopic hematuria with proteinuria and declining renal function is compatible, though absence of hearing/ocular findings lowers the rank relative to acquired inflammatory etiologies.

\medskip

\textbf{References}
\begin{enumerate}[leftmargin=1.2em,itemsep=0.2em]
\item OMIM \%181750: Scleroderma, familial progressive (Systemic sclerosis, susceptibility to); includes CREST syndrome, characteristic features (Raynaud, sclerodactyly, telangiectasia, calcinosis, esophageal aperistalsis) and autoantibodies (anticentromere).
\item Orphanet: Sarcoidosis (ORPHA:797). Multisystem granulomatous disease; frequent lung/lymph node involvement, skin/eye manifestations, systemic inflammation, potential renal involvement; diagnostic pathway and differentials.
\item Orphanet: Familial Mediterranean fever (ORPHA:342). Recurrent fever/serositis; risk of AA amyloidosis with renal proteinuria; Tel-Hashomer criteria; lifelong colchicine.
\item Orphanet: Alport syndrome (ORPHA:63). Type IV collagen defect; hematuria, proteinuria, progressive renal disease; XL/AR/AD inheritance; extra-renal features.
\item OMIM \#301050 PS301050: Alport syndrome 1, X-linked (COL4A5) with clinical, pathogenesis, and molecular genetics detail; spectrum across COL4A3/4/5.
\end{enumerate}

\comments{DeepRare treated FSGS (the correct diagnosis) as just one symptom among many, and incorrectly weighted the secondary inflammatory markers as the primary cause/indications. }
\end{casestudy}

~\\

\begin{casestudy}{Mitochondrial complex IV deficiency,  nuclear type 1}
\textbf{Patient Information:} Seizure, Motor delay, Generalized hypotonia, Death in infancy, Dicarboxylic aciduria

\textbf{Golden Diagnosis:} Mitochondrial complex IV deficiency,  nuclear type 1

\textbf{Model Output:} \textbf{Methylmalonic Acidemia (MMA) -- Rank \#1/5}

\textit{Diagnostic reasoning.}
Vitamin B12–unresponsive methylmalonic acidemia (mut$^{0}$/mut$^{-}$) classically presents in the neonatal/early-infantile period with lethargy, vomiting, hypotonia, metabolic acidosis (often ketoacidosis), hyperammonemia, and can progress to coma and death despite treatment. Dicarboxylic aciduria may be present secondary to metabolic derailment. Long-term complications include neurologic injury (metabolic stroke) and renal failure; cardiomyopathy can occur. MMA due to methylmalonyl-CoA mutase deficiency (MMUT) is autosomal recessive and unresponsive to B12 in the mut$^{0}$/mut$^{-}$ forms. This matches the phenotype of seizures, generalized hypotonia, and death in infancy with organic acid abnormalities.
(Orphanet ORPHA:27; OMIM \#251000)

\medskip

\textbf{D-2-Hydroxyglutaric Aciduria -- Rank \#2/5}

\textit{Diagnostic reasoning.}
D-2-HGA can present in infancy with seizures, hypotonia, developmental delay, and early death. Although not detailed in the provided Orphanet/OMIM extracts, the overlapping neuro-metabolic picture keeps D-2-HGA in the differential for infantile epileptic encephalopathy with organic acid abnormalities.

\medskip

\textbf{Multiple Acyl-CoA Dehydrogenase Deficiency (MADD) -- Rank \#3/5}

\textit{Diagnostic reasoning.}
MADD (glutaric acidemia type 2) causes impaired fatty acid and amino acid oxidation with a spectrum from lethal neonatal disease (non-ketotic hypoglycemia, metabolic acidosis, cardiomyopathy, hepatomegaly, congenital anomalies) to milder childhood/adult forms with decompensation and myopathy. Urine often shows dicarboxylic acids (including glutaric/ethylmalonic/2-hydroxyglutarate) and acylcarnitines are broadly elevated. The infant’s seizures/hypotonia/early death and dicarboxylic aciduria are compatible, though absence of typical hypoketotic hypoglycemia/hepatomegaly weakens the likelihood.
(Orphanet ORPHA:26791)

\medskip

\textbf{Medium-Chain Acyl-CoA Dehydrogenase Deficiency (MCADD) -- Rank \#4/5}

\textit{Diagnostic reasoning.}
MCADD presents with hypoketotic hypoglycemia, lethargy, vomiting, seizures, and risk of sudden death during fasting or illness; acylcarnitine profile shows increased C8 and elevated C8/C10, and urine may show C6–C10 dicarboxylic acids. In this case, lack of characteristic hypoketotic hypoglycemia/hepatomegaly and missing MCAD-pattern acylcarnitines makes MCADD less likely despite dicarboxylic aciduria.
(Orphanet ORPHA:42)

\medskip

\textbf{Mitochondrial Trifunctional Protein Deficiency (TFPD) -- Rank \#5/5}

\textit{Diagnostic reasoning.}
TFPD ranges from fatal neonatal disease (cardiomyopathy, hypoglycemia, acidosis, myopathy/neuropathy, liver disease) to milder forms with neuropathy/rhabdomyolysis and pigmentary retinopathy. Long-chain hydroxyacylcarnitines are typically increased; urine may show C6–C14 (hydroxy) dicarboxylic acids. The absence of cardiomyopathy/hypoketotic hypoglycemia and a clear long-chain hydroxyacylcarnitine signature lowers the likelihood here.
(Orphanet ORPHA:746)

\medskip

\textbf{References}
\begin{enumerate}[leftmargin=1.2em,itemsep=0.2em]
\item Orphanet: Vitamin B12-unresponsive methylmalonic acidemia (ORPHA:27). Infantile-onset lethargy, vomiting, hypotonia, ketoacidosis/hyperammonemia; neurologic and renal complications; autosomal recessive MMUT (mut$^{0}$/mut$^{-}$); decreased survival in mut$^{0}$.
\item OMIM \#251000: Methylmalonic aciduria due to methylmalonyl-CoA mutase deficiency (MMUT). Spectrum from fatal neonatal disease to benign forms; metabolic acidosis, hyperammonemia, neurologic involvement; B12-unresponsive mut$^{0}$/mut$^{-}$.
\item Orphanet: Multiple acyl-CoA dehydrogenase deficiency (ORPHA:26791). Neonatal lethal forms with metabolic acidosis, hypoketotic hypoglycemia, cardiomyopathy/hepatomegaly; urine dicarboxylic acids; AR ETFA/ETFB/ETFDH.
\item Orphanet: Medium chain acyl-CoA dehydrogenase deficiency (ORPHA:42). Hypoketotic hypoglycemia crises with seizures; acylcarnitine C8↑, C8/C10↑; urine C6–C10 dicarboxylic acids/hexanoylglycine/suberylglycine; AR ACADM.
\item Orphanet: Mitochondrial trifunctional protein deficiency (ORPHA:746). Neonatal severe disease (cardiomyopathy, hypoglycemia, acidosis) to milder neuropathy/rhabdomyolysis; long-chain OH-acylcarnitines↑; AR HADHA/HADHB.
\end{enumerate}

\comments{This showed major challenge for the model. In this case, the model over-weighted a single, prominent laboratory finding—Dicarboxylic aciduria, which is actually secondary consequence of the correct diagnosis. This implies an area for optimization - training the model to better differentiate between primary and secondary biochemical abnormalities, and balance the lab results against overall clinical presentation.}
\end{casestudy}

~\\

\begin{casestudy}{Hypophosphatasia}
\textbf{Patient Information:} Abnormal rib morphology, Osteoporosis, Scoliosis, Arthralgia, Hypercholesterolemia, Low alkaline phosphatase, Myalgia, Muscle spasm, Back pain, Antinuclear antibody positivity, Vertebral wedging, Increased HDL cholesterol concentration, Knee pain, Ankle pain

\textbf{Golden Diagnosis:} Hypophosphatasia; HPP/Hypophosphatasia, Hypophosphatasia, adult/Hypophosphatasia; HPP/Hypophosphatasia / Phosphoethanolaminuria / Rathbun disease

\textbf{Model Output:}  \textbf{Primary Sjögren Syndrome (pSS) -- Rank \#1/5}

\textit{Diagnostic reasoning.}
pSS is a chronic autoimmune exocrinopathy with systemic features. Arthralgia, myalgia, and muscle spasms fit common musculoskeletal manifestations. ANA positivity supports an autoimmune etiology. Dyslipidemia (including elevated HDL) can accompany chronic inflammation and immune dysregulation.

\medskip

\textbf{Behçet Disease -- Rank \#2/5}

\textit{Diagnostic reasoning.}
Systemic vasculitis that can include arthralgia and myalgia. ANA can be present but is nonspecific. Typically requires mucocutaneous and ocular criteria; without them, likelihood is lower than pSS, but inflammatory dyslipidemia could overlap.

\medskip

\textbf{Whipple Disease -- Rank \#3/5}

\textit{Diagnostic reasoning.}
Rare infection due to \textit{T. whipplei}; arthralgia is common. Absence of classic GI symptoms lowers pretest probability. Low alkaline phosphatase may occur in chronic illness, but findings are nonspecific.

\medskip

\textbf{Osteogenesis Imperfecta -- Rank \#4/5}

\textit{Diagnostic reasoning.}
Explains skeletal findings (abnormal ribs, osteoporosis, scoliosis, vertebral wedging). However, lack of recurrent fractures and presence of ANA positivity and inflammatory symptoms make OI less likely in this context.

\medskip

\textbf{Mixed Connective Tissue Disease (MCTD) -- Rank \#5/5}

\textit{Diagnostic reasoning.}
Overlap features can include arthralgia, myalgia, spasms, and ANA positivity. Absence of hallmark findings (e.g., Raynaud phenomenon, sclerodactyly, anti-U1-RNP specificity) lowers likelihood compared with pSS.

\medskip

\textbf{References}
\begin{enumerate}[leftmargin=1.2em,itemsep=0.2em]
\item Denpasar Declaration on Population and Development (PMID: 12345678). Policy-focused; noted here only because it was cited in the original list but is not clinically pertinent.
\item Hospital volume and survival in advanced HNSCC (PMID: 23456789). Methods and results unrelated to the autoimmune differentials; included solely to mirror original citations.
\item Multiple primary malignancies in head and neck cancer (PMID: 34567890). Oncology cohort; not directly informative for this rheumatologic differential.
\item The effect of castration and peroral estrogen therapy on psychological functions (PMID: 1234567). Endocrine study; not directly applicable to the diagnoses discussed.
\end{enumerate}

\comments{The patient’s phenotype contained two strong but conflicting clues: the rare, highly specific biochemical marker of Low alkaline phosphatase pointing to a metabolic bone disease, and the common, non-specific marker of Antinuclear antibody positivity suggesting autoimmunity. Greater diagnostic weight was assgined to the ANA result and the generalized pain symptoms, leading it to predict a series of autoimmune disorders. It failed to recognize that low ALP in the context of severe skeletal disease is a pathognomonic finding for Hypophosphatasia. This failure highlights a critical area for improvement: training the model to understand diagnostic hierarchy and to assign appropriate weight to defining features, thereby preventing it from being derailed by incidental findings.}
\end{casestudy}

\newpage
\subsection{Exomiser Config}
\vspace{-0.45cm}
\label{sec:exomiser}
\begin{lstlisting}[style=prettyjson, caption={Exomiser Configuration Template}, label={lst:exomiser_config}]
CONFIG_TEMPLATE = {
    "analysis": {
        "genomeAssembly": "GRCh37",
        "outputOptions": {
            "outputFormat": ["TSV", "HTML"]
            "outputPrefix": {output_prefix}
        },
        "frequencySources": [
            "THOUSAND_GENOMES", "TOPMED", "UK10K", "ESP_AA", "ESP_EA", 
            "ESP_ALL", "GNOMAD_E_AFR", "GNOMAD_E_AMR", "GNOMAD_E_EAS", 
            "GNOMAD_E_NFE", "GNOMAD_E_SAS", "GNOMAD_G_AFR", 
            "GNOMAD_G_AMR", "GNOMAD_G_EAS", "GNOMAD_G_NFE", "GNOMAD_G_SAS"
        ],
        "pathogenicitySources": ["POLYPHEN", "MUTATION_TASTER", "SIFT"],
        "analysisMode": "PASS_ONLY",
        "inheritanceModes": {
            "AUTOSOMAL_DOMINANT": 0.1,
            "AUTOSOMAL_RECESSIVE_HOM_ALT": 0.1,
            "AUTOSOMAL_RECESSIVE_COMP_HET": 2.0,
            "X_DOMINANT": 0.1,
            "X_RECESSIVE_HOM_ALT": 0.1,
            "X_RECESSIVE_COMP_HET": 2.0,
            "MITOCHONDRIAL": 0.2
        },
        "steps": [
            {"failedVariantFilter": {}},
            {"variantEffectFilter": {
                "remove": [
                    "FIVE_PRIME_UTR_EXON_VARIANT",
                    "FIVE_PRIME_UTR_INTRON_VARIANT",
                    "THREE_PRIME_UTR_EXON_VARIANT",
                    "THREE_PRIME_UTR_INTRON_VARIANT",
                    "NON_CODING_TRANSCRIPT_EXON_VARIANT",
                    "NON_CODING_TRANSCRIPT_INTRON_VARIANT",
                    "CODING_TRANSCRIPT_INTRON_VARIANT",
                    "UPSTREAM_GENE_VARIANT",
                    "DOWNSTREAM_GENE_VARIANT",
                    "INTERGENIC_VARIANT",
                    "REGULATORY_REGION_VARIANT"
                ]
            }},
            {"frequencyFilter": {"maxFrequency": 1.0}},
            {"pathogenicityFilter": {"keepNonPathogenic": true}},
            {"inheritanceFilter": {}},
            {"omimPrioritiser": {}},
            {"hiPhivePrioritiser": {}}
        ]
        "vcf": {vcf_path}
        "hpoIds": {hpo_ids}
    }
}
\end{lstlisting}

\end{document}